\documentclass[10pt]{article}

\usepackage[preprint]{tmlr}
\newcommand{\arxivversion}{} %

\usepackage[utf8]{inputenc} %
\usepackage[T1]{fontenc}    %
\usepackage{hyperref}       %
\usepackage{url}            %
\usepackage{booktabs}       %
\usepackage{amsfonts}       %
\usepackage{amsmath}        %
\usepackage{nicefrac}       %
\usepackage{microtype}      %
\usepackage{xcolor}         %
\usepackage{graphicx}       %
\usepackage{natbib}         %

\newcounter{finding}
\renewcommand{\thefinding}{\#\arabic{finding}}  %
\newcommand{\findingpar}[2]{%
  \stepcounter{finding}%
  \paragraph{Finding \thefinding{}#2}%
  \addtocounter{finding}{-1}\refstepcounter{finding}\label{#1}}

\newcommand{\OursConceptPoolSize}{750}

\newcommand{\OursConceptNZeroAssertion}{1}

\newcommand{\OursNKeptConcepts}{14}

\newcommand{\OursNDistinctConcepts}{37}
\newcommand{\OursConceptMarks}{1{,}229}

\newcommand{\OursTopConcept}{river}

\newcommand{\OursTopConceptShareAll}{80.5\%}

\newcommand{\OursSecondConceptShareAll}{18.4\%}
\newcommand{\OursWeaverShareAll}{13.6\%}

\newcommand{\OursEffNConcepts}{6.23}

\newcommand{\OursTopConceptShareAllRecovered}{80.7\%}
\newcommand{\OursEffNConceptsRecovered}{6.23}
\newcommand{\OursNMenuResponses}{132}
\newcommand{\OursNCommittedResponses}{617}
\newcommand{\OursNMenuCueAmongMenu}{131}
\newcommand{\OursNMenuCueAmongCommitted}{0}
\newcommand{\OursNModelsWithMenuResponses}{5}
\newcommand{\OursTopMenuModelRate}{86.0\%}
\newcommand{\OursNConceptsOnlyInMenuResponses}{26}
\newcommand{\OursCommittedTopConceptShare}{85.1\%}
\newcommand{\OursCommittedEffNConcepts}{1.86}

\newcommand{\JiangConceptPoolSize}{2{,}150}
\newcommand{\JiangNZeroImage}{14}
\newcommand{\JiangNDistinctImages}{144}
\newcommand{\JiangNKeptImages}{29}
\newcommand{\JiangImageMarks}{3{,}593}

\newcommand{\JiangTopImageShareResponses}{90.2\%}
\newcommand{\JiangTopImageShareMarks}{54.0\%}
\newcommand{\JiangSecondImage}{thief}
\newcommand{\JiangSecondImageShareResponses}{4.5\%}

\newcommand{\JiangWeavingFamilyShareResponses}{3.3\%}

\newcommand{\JiangNQwQResponses}{50}
\newcommand{\JiangQwQShareMarks}{36.8\%}

\newcommand{\JiangNDistinctImagesNoQwQ}{78}

\newcommand{\JiangWeavingFamilyShareNoQwQ}{1.5\%}

\newcommand{\JiangNZeroPairWithVehicle}{12}
\newcommand{\BlindRelabelNPerPool}{100}
\newcommand{\BlindRelabelOursPrimaryInSet}{100}
\newcommand{\BlindRelabelJiangPrimaryInSet}{96}
\newcommand{\BlindRelabelOursPrimaryInSetRaw}{89}
\newcommand{\BlindRelabelJiangPrimaryInSetRaw}{95}
\newcommand{\BlindRelabelOursRiverAgree}{100}
\newcommand{\BlindRelabelJiangRiverAgree}{98}
\newcommand{\JiangNClaims}{247}
\newcommand{\JiangMeanClaimsPerResponse}{4.9}
\newcommand{\JiangShareTwoPlusClaims}{89.4\%}
\newcommand{\JiangRiverNClaims}{139}
\newcommand{\JiangRiverEffNClaims}{33}
\newcommand{\OursTopImageShareMarks}{49.1\%}

\newcommand{\OursNClaims}{87}
\newcommand{\OursMeanClaimsPerResponse}{4.0}
\newcommand{\OursShareTwoPlusClaims}{89.6\%}
\newcommand{\OursRiverNClaims}{49}
\newcommand{\OursRiverEffNClaims}{15}

\providecommand{\name}{}
\providecommand{\addr}{}

\title{
How Strong Is the Evidence for the Artificial Hivemind?\\Reevaluating Evidence for the Open-Ended Homogeneity of Language Models
}

\author{%
  \name \href{https://openreview.net/profile?id=~Rylan_Schaeffer2}{Rylan Schaeffer}\textsuperscript{1}\quad
  \href{https://openreview.net/profile?id=~Brando_Miranda1}{Brando Miranda}\textsuperscript{1}\quad
  \href{https://openreview.net/profile?id=~Joshua_Kazdan1}{Joshua Kazdan}\textsuperscript{2}\\
  \name \href{https://openreview.net/profile?id=~Jessica_Chudnovsky1}{Jessica Chudnovsky}\textsuperscript{1}\quad
  \href{https://openreview.net/profile?id=~Sanmi_Koyejo1}{Sanmi Koyejo}\textsuperscript{1}\\[0.5em]
  \addr \textsuperscript{1}Stanford Computer Science\quad
  \textsuperscript{2}Stanford Statistics%
}
\hypersetup{
  hidelinks,
  pdfauthor={Rylan Schaeffer, Brando Miranda, Joshua Kazdan, Jessica Chudnovsky, Sanmi Koyejo}
}

\begin{document}

\maketitle
\begin{abstract}
Recent research argues that language models exhibit pronounced homogeneity in open-ended generation, framing such behavior as an Artificial Hivemind that poses a long-term threat to human creativity.
We examine three of its central results.
First, the flagship example is that model responses to ``Write a metaphor involving time'' collapse into two clusters.
Visualization, spectral analysis, clustering, and language model labels all contradict this description.
The labels record each response's vehicle, what it compares time to.
Our responses and the original authors' own show one dominant vehicle plus a heavy tail of distinct minority vehicles.
``Time'' is one of our least diverse topics, so the example is a favorable case, not a representative one.
Second, the paper measures homogeneity against an undemanding null: responses to unrelated prompts.
Under a more demanding null (same-prompt responses expressing genuinely different ideas), 20\%--32\% of such pairs already exceed the paper's 0.8 convergence threshold.
A residual effect survives this null.
The paper's same-prompt pairs exceed 0.8 roughly two to three times as often as our different-idea pairs.
Much of what the paper calls homogeneity is the shared geometry of answering the same prompt.
The remaining measurements lack any null: no human baseline is collected, and the model-indistinguishability statistic has no null.
Third, the paper concludes that inference-time interventions are inadequate for combating the Artificial Hivemind, writing that ``more generalizable solutions are needed at the model training level.''
We show that this conclusion is unsupported in three ways, and that an inference-time intervention (prompting) reliably raises measured response diversity.
We do not resolve whether the Artificial Hivemind is real. We show that the published evidence does not establish it.
\end{abstract}

\section{Introduction}
\label{sec:introduction}

Do large language models produce homogeneous outputs? If billions of people use language models for creative work, and those models converge on a narrow range of responses, collective human expression could diminish over time \citep{bommasani2022picking}. Several studies argue this concern is real, e.g., \citep{doshi2024generative, anderson2024homogenization, ashkinaze2025how, kumar2025human, kosmyna2025your}.

\citet{jiang2025artificial} recently argued that language models are converging to an ``Artificial Hivemind'', both in \emph{intra-model repetition} (similar outputs from one model)
and \emph{inter-model homogeneity} (similar outputs across models).
The paper presents the Hivemind as ``long-term AI safety risks'' to human creativity and pluralistic alignment.
The paper asks an important question. We examine its evidence as a measurement problem.
Homogeneity is a comparative property, so a similarity threshold without a null distribution is not a measurement.
Section~\ref{sec:wrong-null} develops that point.
Sections~\ref{sec:artificial_hivemind_figure1} and~\ref{sec:inference-time} ask whether the flagship example and the training-level conclusion survive first-line checks.
We organize our manuscript around three claims.

\paragraph{First, the Artificial Hivemind's flagship example is undercut by five lines of evidence.}
We use the same embedding model, sampling parameters, and a closely matching prompt, but a more recent set of 15 models.
(1) \textbf{Visually}: When we reproduce the 2D PCA visualization \citet{jiang2025artificial} used to display the claim, every topic shows more than two groupings.
(2) \textbf{Spectrally}: the 2D projection captures only 25--41\% of the variance. The eigenspectrum shows no concentration of the kind a tight two-cluster structure would produce.
(3) \textbf{Clustering}: Three clustering algorithms of three types (probabilistic, density, spectral) on the full-dimensional embeddings select many clusters rather than two.
(4) \textbf{Language Model Labeling}: A metaphor has a tenor (its subject, such as time), a vehicle (what the subject is compared to, such as a river), and a ground (what the comparison asserts) \citep{richards1936philosophy}. The original paper uses one tenor, time; we test twelve, but label vehicles only for time. A language model (Claude Fable 5) labels, from the text alone, the vehicle of each of our own \OursConceptPoolSize{} time-metaphor responses. The labels show one dominant vehicle on \OursTopConceptShareAll{} of responses plus a heavy tail of distinct minority vehicles, not the two clusters the original paper describes.
(5) \textbf{Case Selection}:
We test eleven other topics, fixed ex ante. ``Time'' is one of the least diverse of the twelve: it ranks first of twelve (least diverse) on both within-model statistics and third of twelve on pooled statistics (Appendix~\ref{app:per-topic-diversity-ranking}, Table~\ref{tab:per-topic-diversity-ranking}), meaning the flagship example is the most favorable case for the claim rather than a representative one.

\begin{figure}[t!]
\centering
\includegraphics[width=\linewidth]{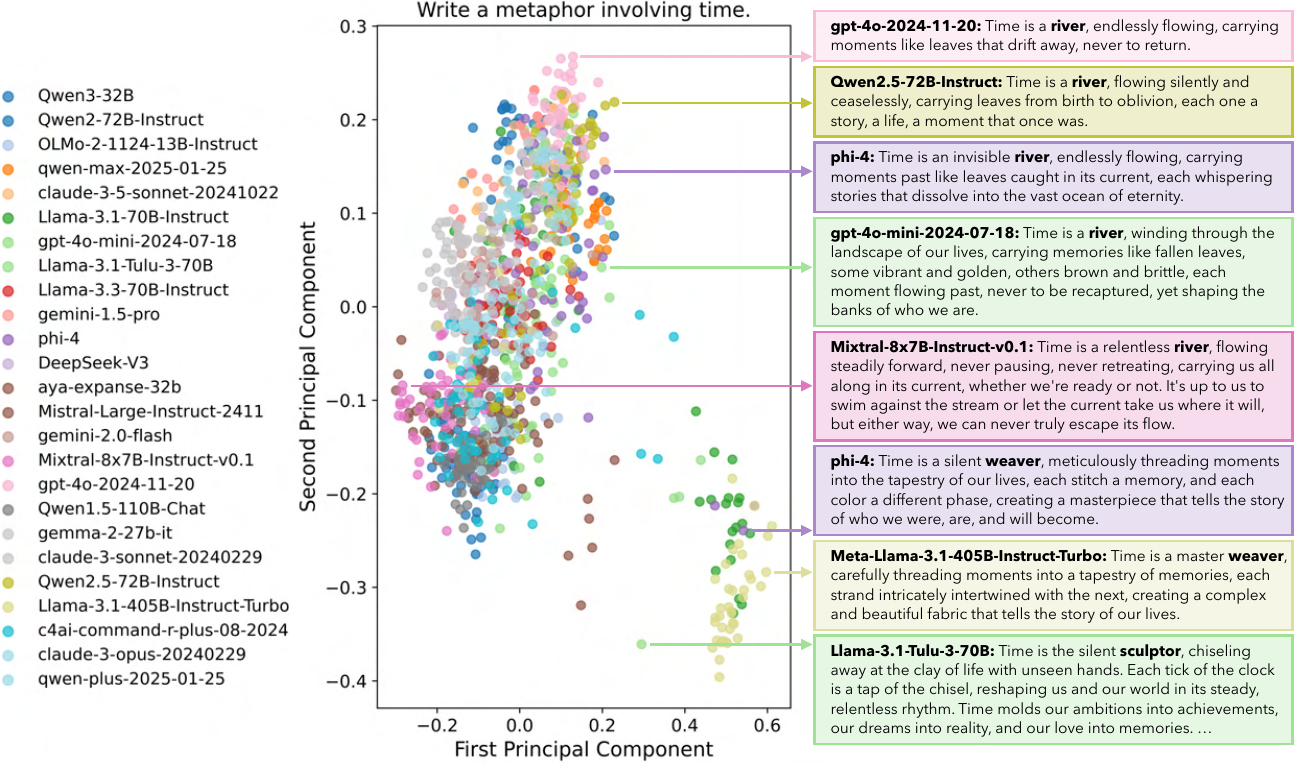}
\caption{\textbf{The Artificial Hivemind's Flagship Example Is Undercut by Five Lines of Evidence.} Figure 1 from \citet{jiang2025artificial}. When 25 language models are each prompted 50 times: ``Write a metaphor involving time,'' the authors found: ``Despite the diversity of model families and sizes, the responses form just two primary clusters''.  
Our reproduction and re-analysis of the original data undercuts this description (Section~\ref{sec:artificial_hivemind_figure1}).}
\label{fig:jiang-figure1}
\end{figure}

\begin{figure}[t!]
\centering
\includegraphics[width=\linewidth,trim={0 0 0 2cm}, clip]{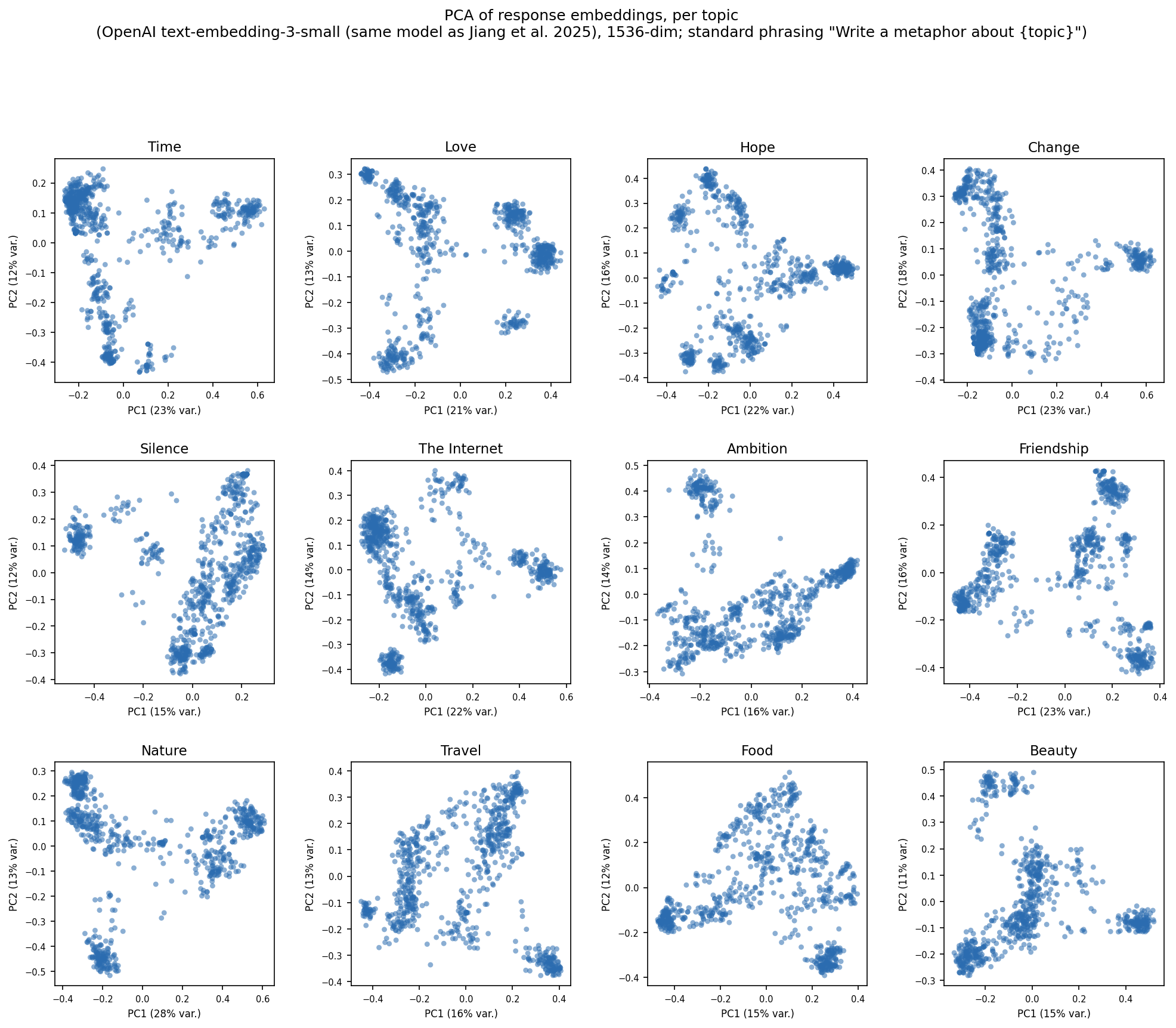}
\caption{\textbf{Across 12 Topics in Our Reproduction, Every Topic Shows More Than Two Clusters.} PCA projections of embedded responses for each of the 12 baseline topics ``Write a metaphor about \{topic\}'', reproducing the original flagship figure visualization. 750 responses per panel: 15 models $\times$ 50 samples. We selected 12 topics: ``time'' to match the original paper, plus 11 others to test the existence of the Hivemind and how representative ``time'' is.
Every panel shows more than two clusters.
}
\label{fig:our-pca-replication}
\end{figure}

\paragraph{Second, the Artificial Hivemind's homogeneity is compared against an undemanding null.}
The null \citet{jiang2025artificial} supplies does not support its conclusion.
The paper compares same-prompt response similarity against pairs of responses to \emph{different} prompts. Such pairs score near zero (median cosine ${\sim}0.11$) because they share no topic, vocabulary, or format.
A more reasonable null is the similarity of same-prompt responses expressing \emph{genuinely different ideas}. We measure it two ways (geometric clustering, LLM labels) across up to 50 prompts from the original paper's \textsc{Infinity-Chat} benchmark. The null's median lies at 0.70--0.72, just above the paper's 0.7 threshold. At the primary 0.8 threshold, 20\%--32\% of such pairs exceed it.
These thresholds cannot distinguish conceptual collapse from prompt conditioning plus a shared house style. Much of the measured convergence is therefore the shared geometry of answering the same prompt. Whether it is most is unknown without a matched human baseline.
A residual effect remains. The paper's same-prompt pairs exceed the 0.8 threshold roughly two to three times as often as our different-idea pairs.
The paper's other homogeneity measurements lack any null: no human-response baseline is measured, and the model-indistinguishability statistic has no null.

\paragraph{Third, the conclusion that training-level solutions are needed rests on invalid logic and is also refuted by a simple inference-time intervention (prompting).}
\citet{jiang2025artificial} ask whether inference-time methods can counter the Hivemind or whether training-time intervention is required. The stakes are practical: Inference-time fixes cost little, work immediately, and are open to anyone; training-time fixes cost much, take time, and are open only to model developers. The paper concludes that inference-time interventions are inadequate, writing: ``more generalizable solutions are needed at the model training level to robustly preserve output diversity without requiring user intervention.''
The paper's evidence is that min-$p$ sampling \citep{minh2025turning}, an algorithm claimed to enhance diversity, slightly reduced extreme repetition while leaving homogeneity largely intact.
We show that this conclusion is unsupported in three ways. The inference is a logical jump. The paper provides no evidence for training-level solutions. And the evidence, that min-$p$ sampling failed to increase diversity, is better explained by min-$p$ not increasing diversity at all.
We then show that prompting, a simple inference-time intervention, reliably increases embedding-space diversity.

\paragraph{A Note on Scope.}
Our scope is narrow. We apply standard first-line checks to the three claims that support the Artificial Hivemind. That its central evidence does not survive is our result.
Findings~\ref{finding:visualization}--\ref{finding:case-selection} address the first claim (Section~\ref{sec:artificial_hivemind_figure1}). Findings~\ref{finding:null-undemanding}--\ref{finding:indistinguishability} address the second (Section~\ref{sec:wrong-null}). Findings~\ref{finding:training-level}--\ref{finding:rewordings} address the third (Section~\ref{sec:inference-time}).
More demanding audits remain open (a matched human-response pool, a proper null for the model-indistinguishability statistic, tail-sensitive summaries in place of scalar means) and any could move the Hivemind case in either direction.
Section~\ref{sec:discussion-metric-validity} states what a validated diversity instrument would require.

\section{The Flagship Example Is Undercut by Five Lines of Evidence}
\label{sec:artificial_hivemind_figure1}

Figure 1 of \citet{jiang2025artificial} is the paper's flagship example of the Artificial Hivemind: when 25 models are each prompted 50 times ``Write a metaphor involving time,''\footnote{The main text and Figure 1 caption of \citet{jiang2025artificial} state the prompt as ``Write a metaphor about time,'' but the actual prompt and Figure 1's data originate from ``Write a metaphor involving time.'' We study both.} the paper reports that the responses ``form just two primary clusters'': ``time is a river'' and ``time is a weaver'' (Fig.~\ref{fig:jiang-figure1}).

The five findings differ in what they assume.
Finding~\ref{finding:visualization} accepts the original paper's 2D view and still finds more than two groups.
Finding~\ref{finding:spectral} shows that the view is not representative.
Finding~\ref{finding:clustering} clusters in the full embedding space.
Finding~\ref{finding:vehicle-labels} uses no embedding.
Finding~\ref{finding:case-selection} asks whether the topic is representative.

\begin{figure}[t!]
\centering
\includegraphics[width=\linewidth]{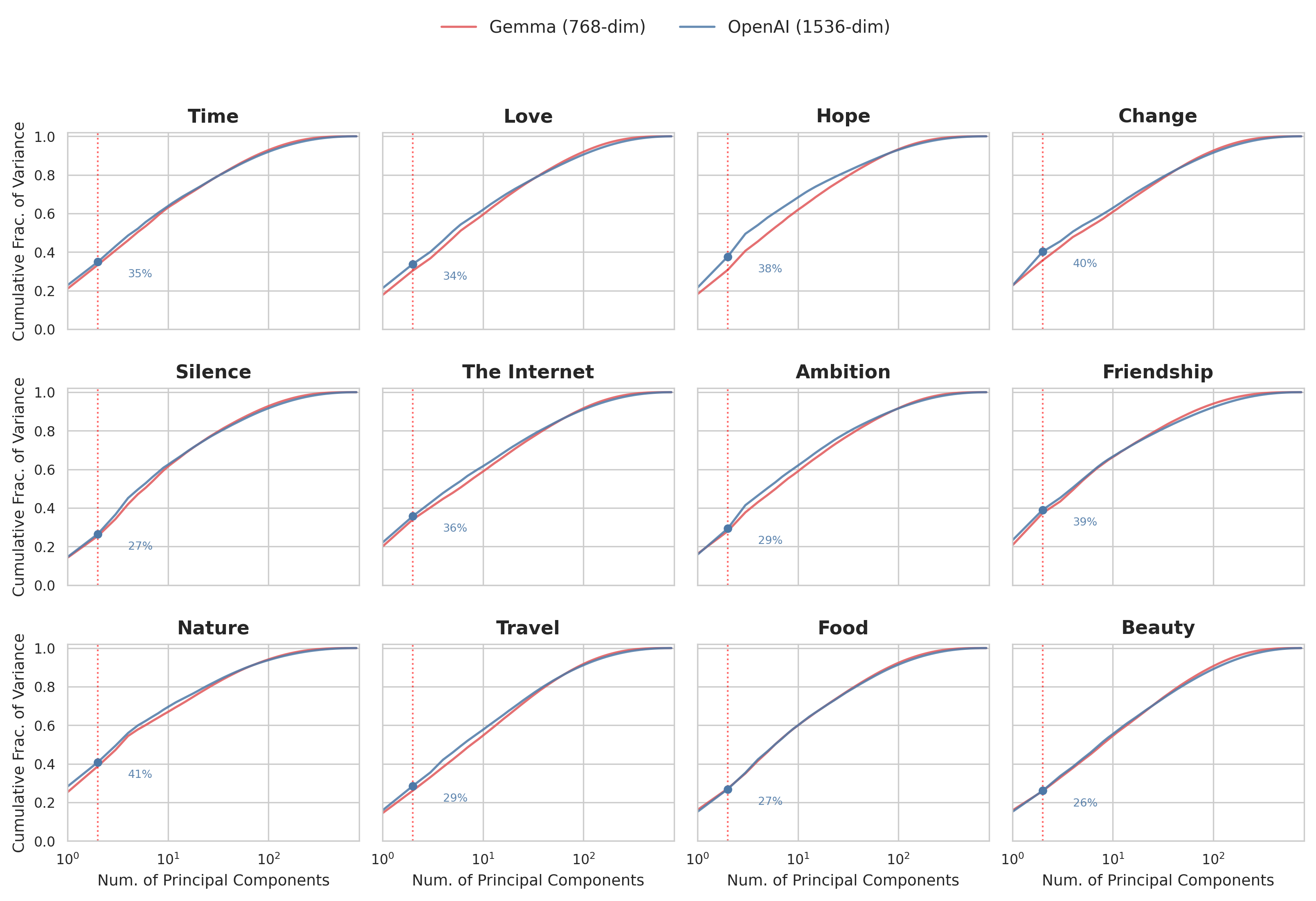}
\caption{\textbf{The Eigenspectra Show No Sharp Elbow, and Two Principal Components Fail to Capture the Majority of the Variance.} Cumulative variance explained versus number of principal components, for all 12 topics under the prompt ``Write a metaphor about \{topic\}''.
These eigenspectra show that (1)~no sharp elbow exists, contrary to what the tight-two-cluster reading of the flagship figure predicts, and (2)~the first two dimensions are not representative.
2 components capture only 25--41\% of variance, depending on topic and embedding model; reaching 90\% of variance requires 60--110 components. 
The 2D PCA visualizations are not a summary; they are a lossy compression that discards most of the structure.
}
\label{fig:pca-spectrum}
\end{figure}

\paragraph{Experimental Methods.}

We selected 12 topics: ``time'' to match the original paper, plus 11 others to test the presence of the Hivemind and how representative ``time'' is (love, hope, change, silence, the Internet, ambition, friendship, nature, travel, food, and beauty).
All twelve were fixed before any data collection.
Models were prompted with ``Write a metaphor about \{topic\}''.
We collected responses from 15 models spanning three providers (four Google, six Anthropic, five OpenAI); Appendix~\ref{app:model-roster} lists the full roster.
We collected 50 responses per model per topic at temperature $= 1.0$ and top-$p = 0.9$ (provider-specific exception in Appendix~\ref{app:model-roster}), matching the original paper, yielding 9{,}000 responses for this section.

\begin{figure}[t!]
\centering
\includegraphics[width=0.62\linewidth]{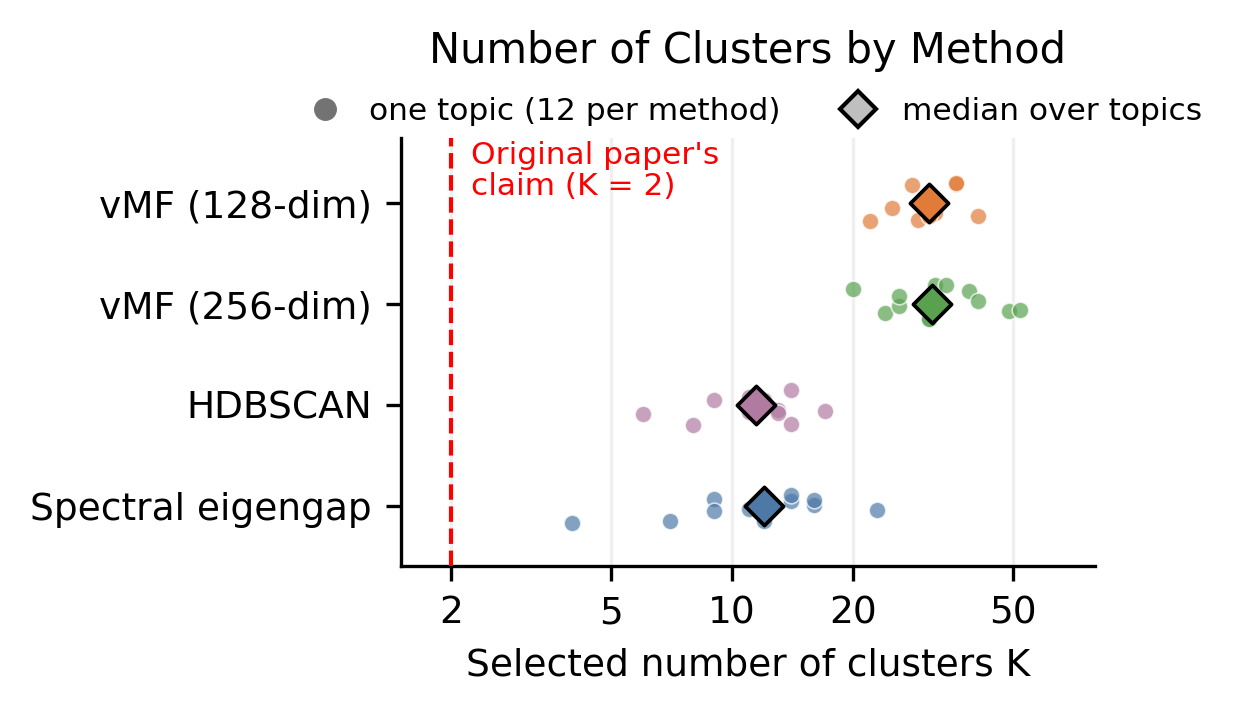}
\caption{\textbf{Every Clustering Method Selects More Than Two Clusters.} Each method's selected number of clusters $K$ over the 12 baseline topics. One small marker per topic; the diamond is the median over topics. The dashed line marks $K{=}2$, the original paper's claim. The methods span probabilistic parametric, nonparametric density-based, and graph-based approaches. Per-method ranges (Appendix~\ref{app:clustering-method-details}): vMF 128-dim $22$--$41$, vMF 256-dim $20$--$52$, HDBSCAN $6$--$17$, spectral eigengap $4$--$23$. Fig.~\ref{fig:clustering-model-selection-per-topic} (appendix) gives the per-topic values.}
\label{fig:clustering-model-selection}
\end{figure}

We embedded all responses with the original paper's embedding model, OpenAI's \texttt{text-embedding-3-small} \citep{neelakantan2022text}.
To test whether the choice of embedding model matters, we additionally used Google DeepMind's \texttt{embeddinggemma-300m} \citep{vera2025embeddinggemmapowerfullightweighttext}; results are consistent across both.

\begin{figure}[t!]
\centering
\includegraphics[width=\linewidth]{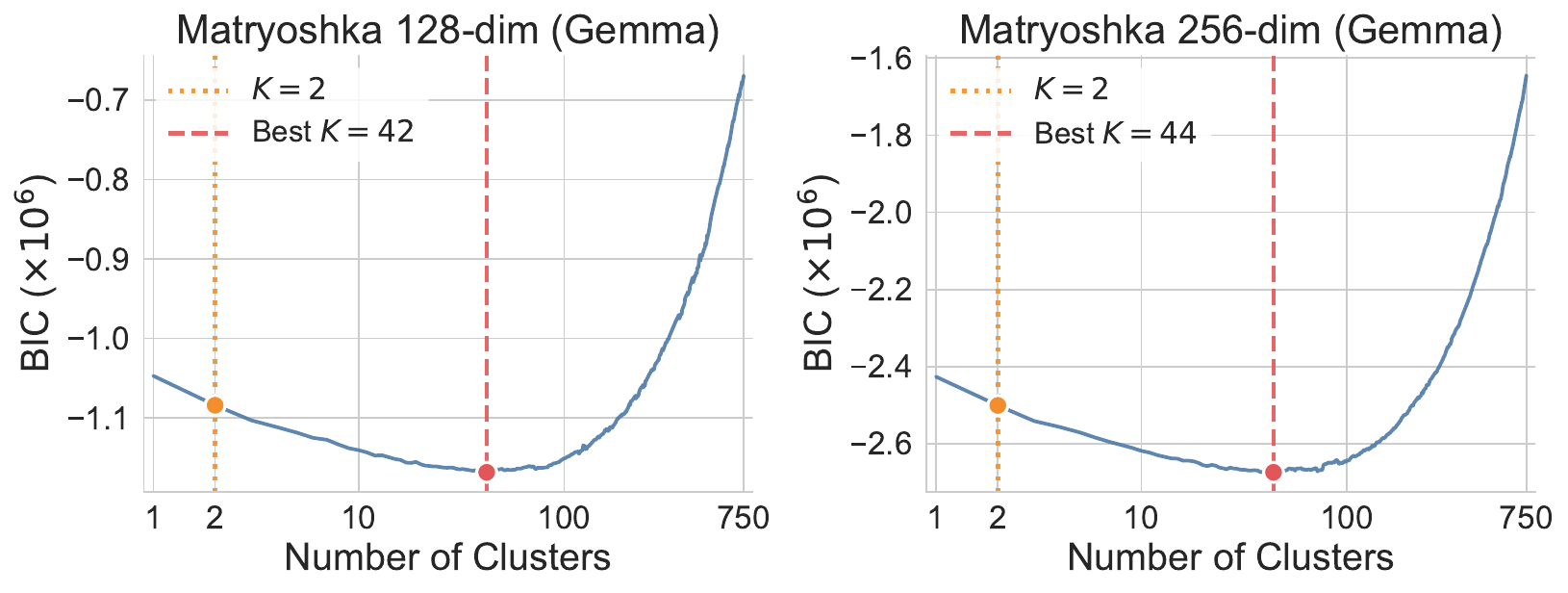}
\caption{\textbf{On the Original Data, BIC Selects More Than Two Clusters.}
vMF-mixture BIC versus $K$ on Gemma Matryoshka slices of \citet{jiang2025artificial}'s model responses (privately shared with us) to ``Write a metaphor involving time.'' ($n=2{,}150$: $43$ models $\times$ $50$ samples), same procedure as Fig.~\ref{fig:clustering-model-selection}: BIC selects $K^{*}{=}42$ (128-dim) and $K^{*}{=}44$ (256-dim) and never selects $K{=}2$ or anything near it.
HDBSCAN (\texttt{min\_cluster\_size}~$=15$, \texttt{min\_samples}~$=1$) finds $K{=}5$ (10.7\% noise) and the spectral eigengap ($k$-NN, $k{=}20$) suggests $K{=}4$: coarser than on our data, but still not two.
See Appendix~\ref{app:jiang-43model-retest-details} for more information.
}
\label{fig:jiang-43model-time-reclustering}
\end{figure}

\paragraph{Experimental Differences from the Original.}
Our model pool differs from the original paper's; we therefore re-test the claim on the original authors' own 43-model responses (Fig.~\ref{fig:jiang-43model-time-reclustering}; Appendix~\ref{app:jiang-43model-retest-details} discusses the model-vintage confound and the other experimental differences).

\findingpar{finding:visualization}{ (Visualization): In the Original Paper's Own 2D View, Every Topic Produces More Than a Handful of Groupings.}
We reproduce the original paper's visualization, a 2D PCA projection, on our data (Fig.~\ref{fig:our-pca-replication}).
Every topic shows visibly more than two clusters in each topic's 2D PCA plane.
``Time,'' the original paper's flagship topic, is visually among the most concentrated of the 12 baseline conditions and still shows roughly six visible clusters.

\findingpar{finding:spectral}{ (Spectral): Two Principal Components Are Not Representative, and the Spectrum Shows No Two-Cluster Concentration.}
In our data, PCs 1 \& 2 capture 25--41\% of total variance, depending on topic and embedding model.
The original paper did not report variance explained.
Reaching 90\% cumulative variance requires 60--110 principal components (Appendix~\ref{app:pca-variance-loglog} reports the 95\% threshold).
Both embedding models show nearly identical curves, so this is not an artifact of the embedding (Fig.~\ref{fig:pca-spectrum}).
A 2D PCA view is not a summary.
It is a lossy compression that can make the data look simpler than they are.
A tight two-cluster structure would put most of the variance in the top few eigenvalues.
We see smooth decay instead, with roughly 100--200 directions of meaningful variation per topic (Fig.~\ref{fig:pca-spectrum}).
Smooth decay is generic for sentence-embedding clouds, so this rules out only the variance concentration that a two-cluster reading predicts.
Appendix~\ref{app:pca-variance-loglog} shows the same smooth decay on a per-component, log-log scale.

\findingpar{finding:clustering}{ (Clustering): Every Algorithm Selects More Than a Couple of Clusters.}
The eigenspectrum rules out collapse into a few tight clusters, but not looser structure. Our clustering analyses, run on the full-dimensional embeddings rather than on the 2D view of Findings~\ref{finding:visualization}--\ref{finding:spectral}, independently find $K \gg 2$.
On the high-dimensional embeddings, we applied three clustering methods spanning three paradigms: von Mises--Fisher (vMF) mixture models \citep{banerjee2005clustering} selected by BIC (probabilistic parametric), HDBSCAN \citep{campello2013density} (nonparametric density-based), and spectral eigengap analysis \citep{vonluxburg2007tutorial} (graph-based).
All three find more than a few clusters (Fig.~\ref{fig:clustering-model-selection}).
The vMF mixture models found $K=22$--$41$ clusters per topic at 128 dimensions; Appendix~\ref{app:clustering-method-details} reports the remaining per-method ranges.
The precise cluster count can be sensitive to each method's hyperparameters (Appendices~\ref{app:clustering-method-details} and~\ref{app:hdbscan-sensitivity}), but the key result is the direction on which every method agrees. BIC never selects $K{\approx}2$, for any topic, at either dimensionality.
HDBSCAN and the spectral eigengap are not formal tests, but they independently select $K \gg 2$.
Our data do not support the characterization of responses as collapsing into a couple of clusters.

\begin{table}[t]
\centering
\caption{\textbf{Language Model Labels Suggest One Dominant Vehicle Plus a Heavy Tail, Not the Two Clusters.} Distribution of our $750$ responses to ``Write a metaphor about time'' ($15$ models $\times$ $50$ samples) over vehicles a language model assigned from the response text alone, with no embedding (procedure in Appendix~\ref{app:concept-label-details}). Vehicles carried by fewer than $7$ responses (0.93\% of the pool) are pooled. A response counts once per vehicle it carries, so shares do not sum to $1$. The effective number of vehicles is computed over the $1{,}229$ vehicle--response marks, not over the response column. Those $1{,}229$ are the $1{,}173$ in the $14$ kept rows plus $56$ in the tail. A response that offers several comparisons can carry more than one tail vehicle. $11$ of the $45$ tail responses carry two, so the tail holds $56$ vehicle--response marks over $45$ responses. Tables~\ref{tab:ours-menu-decomposition} and~\ref{tab:ours-committed-only-concept-distribution} in Appendix~\ref{app:concept-label-details} give the menu-answering caveat and the committed-only restriction.}
\label{tab:ours-concept-distribution}
\begin{tabular}{lrr}
\toprule
Vehicle & $n$ responses & Share of all $750$ \\
\midrule
\texttt{river} & 604 & 80.5\% \\
\texttt{sculptor} & 138 & 18.4\% \\
\texttt{pickpocket} & 123 & 16.4\% \\
\texttt{weaver} & 102 & 13.6\% \\
\texttt{fire} & 45 & 6.0\% \\
\texttt{currency} & 31 & 4.1\% \\
\texttt{train} & 31 & 4.1\% \\
\texttt{glacier} & 22 & 2.9\% \\
\texttt{calligrapher} & 20 & 2.7\% \\
\texttt{tide} & 19 & 2.5\% \\
\texttt{candle} & 17 & 2.3\% \\
\texttt{book} & 7 & 0.9\% \\
\texttt{ocean} & 7 & 0.9\% \\
\texttt{shadow} & 7 & 0.9\% \\
\midrule
\emph{other vehicles} ($n<7$; 23 of them) & 45 & 6.0\% \\
\emph{no vehicle} (asserted nothing) & 1 & 0.1\% \\
\midrule
\multicolumn{3}{l}{\emph{All vehicles}: $750$ responses, $1{,}229$ vehicle--response marks} \\
\multicolumn{3}{l}{\quad $1{,}229 = 1{,}173 + 56$: the $14$ kept rows plus the vehicle--response marks on the $45$ tail responses} \\
\multicolumn{3}{l}{Effective number of vehicles: $6.23$ (incl.\ remainder), $5.60$ (kept only)} \\
\bottomrule
\end{tabular}
\end{table}

\findingpar{finding:vehicle-labels}{ (Language Model Labels): One Dominant Vehicle Plus a Heavy Tail, Not Two Clusters.}
Findings~\ref{finding:visualization}--\ref{finding:clustering} are all computed from embeddings, so the visualization, spectral, and clustering lines share one instrument.
The vehicle labels use no embedding.
A metaphor has a tenor (its subject, such as time), a vehicle (what the subject is compared to, such as a river), and a ground (what the comparison asserts) \citep{richards1936philosophy}.
We labeled the vehicle of each of our $\OursConceptPoolSize$ time-metaphor responses, that is, what each compares time to, from the response text alone (Appendix~\ref{app:concept-label-details}).
A claim about ``primary clusters'' of ideas is a claim about content.
Its direct test is therefore the distribution of responses over vehicles, not the number of density modes.
Our statistic is the effective number of vehicles: $\exp$ of the Shannon entropy of the vehicle shares.
An effective number of 3 behaves like 3 equally common vehicles.
We read it together with the top-vehicle share.
An effective number near two has two possible sources.
It can come from two vehicles of similar share.
It can also come from one dominant vehicle plus a heavy tail of distinct minority vehicles.
Only the shares distinguish the two cases.
Vehicle groupings are groupings by content.
The geometric clusters of Finding~\ref{finding:clustering} are density modes, and the two need not coincide.
The distribution (Table~\ref{tab:ours-concept-distribution}) is one dominant vehicle plus a heavy tail of distinct minority vehicles, not the two clusters the original paper describes.
\texttt{\OursTopConcept} is carried by \OursTopConceptShareAll{} of the $\OursConceptPoolSize$ responses.
The next-largest vehicle is carried by \OursSecondConceptShareAll{}.
\texttt{weaver}, the second cluster the original paper names, is carried by \OursWeaverShareAll{}.
$\OursNKeptConcepts$ vehicles clear the retention bar.
The effective number of vehicles is $\OursEffNConcepts$ over the $\OursConceptMarks$ vehicle--response marks.
Table~\ref{tab:ours-concept-distribution-with-recovered} repeats this with the $\OursConceptNZeroAssertion$ zero-assertion response's vehicle recovered; the top share is then \OursTopConceptShareAllRecovered{} and the effective number $\OursEffNConceptsRecovered$.
The count of tail vehicles depends on menu responses; the one-dominant-vehicle shape does not.
$\OursNMenuResponses$ responses answer with a menu of several metaphors rather than one, and $\OursNMenuCueAmongMenu$ of them say so explicitly.
The behavior is concentrated in $\OursNModelsWithMenuResponses$ of the $15$ models.
$\OursNConceptsOnlyInMenuResponses$ of the $\OursNDistinctConcepts$ vehicles occur in no single-metaphor response (Table~\ref{tab:ours-menu-decomposition}).
Restricting to the $\OursNCommittedResponses$ responses that committed to one metaphor collapses the tail.
The effective number falls to $\OursCommittedEffNConcepts$.
This restriction makes the conclusion stronger, not weaker.
\texttt{\OursTopConcept} then holds \OursCommittedTopConceptShare{}, and no second dominant vehicle appears under either reading.
The same reading, applied to the original authors' own responses, is reported in the re-test paragraph below.

\paragraph{Why the Shape Matters.}
We do not dispute concentration.
\texttt{river} holds \OursTopConceptShareAll{} of our responses and \JiangTopImageShareResponses{} of the original authors'.
The effective number of vehicles is $\OursEffNConcepts$ over all our vehicle--response marks.
Among our committed responses it is $\OursCommittedEffNConcepts$, below the 2 that two equal clusters would give.

What we dispute is the description.
The original paper names two clusters.
On its own responses, the second named cluster, weaving, holds \JiangWeavingFamilyShareResponses{} and is not the second-largest vehicle.
The rest of the pool is not a second cluster but many small vehicles.

The description bears on the remedy.
If the pool had two clusters and no tail, only a change to the models could add a third vehicle.
Our pools instead show one dominant vehicle plus a heavy tail of distinct minority vehicles.
The alternatives already appear under the default prompt, so an inference-time intervention can act on them.
Section~\ref{sec:inference-time} tests one such intervention, prompting, and finds that it moves the measured diversity.

Whether every model carries the same tail is a per-model question we do not test.
Appendix~\ref{app:concept-label-details} shows the tail comes largely from $\OursNModelsWithMenuResponses$ models that answer with menus.

\findingpar{finding:case-selection}{ (Case Selection): ``Time'' Is One of the Least Diverse Topics We Tested, Not a Representative One.}
The results above argue that the two-cluster description is inaccurate.
This result instead says that even if the description held, ``time'' is not a representative topic on which to establish the Artificial Hivemind.
We fixed the 12 topics before any data collection.
Among them, ``time'' ranks first of twelve (least diverse) on both within-model statistics: Rao's quadratic entropy at the within-model scale and within-model mean pairwise cosine.
It ranks third of twelve on pooled Rao's Q (Appendix~\ref{app:per-topic-diversity-ranking}, Table~\ref{tab:per-topic-diversity-ranking}).
The ranking uses the original paper's own statistic, within-model mean pairwise cosine, and its Rao's Q equivalent (Appendix~\ref{app:alpha-beta-gamma-primer}).
We adopt it deliberately, as Section~\ref{sec:inference-time} does when it reproduces the original paper's intra-model statistic.
Rao's Q under cosine dissimilarity is the total variance.
It measures tightness, not the number of directions the spread uses.
Effective rank and participation ratio measure the number of directions.
On them ``time'' ranks 11th and 12th of twelve (Appendix~\ref{app:per-topic-diversity-ranking}).
Its small residual scatter around \texttt{river} points in many directions.
Concentration is a question of tightness, so we do not rank topics on those two statistics.
Section~\ref{sec:inference-time}'s gains do not depend on this choice.
They hold on Rao's Q at the $\alpha$ and $\gamma$ scales and on effective rank and participation ratio.
A content instrument agrees on time's rank.
The idea labels of Appendix~\ref{app:conceptual-vs-stylistic} cover all twelve topics under the baseline prompt in two sampling seeds.
There ``time'' ranks second of twelve on both modal-idea share and effective number of ideas (Table~\ref{tab:per-topic-idea-label-ranking}).

\paragraph{Re-Testing on the Original Paper's Own 43-Model Responses.}
The 43-model pool the original authors privately shared with us is not the 25-model pool behind the flagship figure.
All 43 are open-weight models from the original paper's set of more than 70.
None is in our 15-model pool (Appendix~\ref{app:jiang-43model-retest-details}).
We repeated this section's analyses on their $2{,}150$ responses ($43$ models $\times$ $50$ samples) to ``Write a metaphor involving time.''
That is the wording recorded in the original paper's released prompt file.
On their pool, as on ours, all three clustering methods recover more than two clusters at essentially every setting. vMF--BIC never selects anything near $K{=}2$; the exceptions are deliberately coarse density- and graph-based settings that also degenerate on our pool (Fig.~\ref{fig:jiang-43model-time-reclustering}; Appendix~\ref{app:jiang-43model-retest-details}).
Their pool is more geometrically concentrated than ours, but matched-$n$ subsamples still show no collapse to two ideas (Appendix~\ref{app:jiang-43model-retest-details}).

The same reader and protocol also labeled the vehicle of each of their $\JiangConceptPoolSize$ responses (Appendix~\ref{app:concept-label-details}; Table~\ref{tab:jiang-image-distribution}).
\texttt{river} is carried by \JiangTopImageShareResponses{} of their responses and accounts for \JiangTopImageShareMarks{} of the $\JiangImageMarks$ vehicle--response marks. The second vehicle, \texttt{\JiangSecondImage}, is carried by \JiangSecondImageShareResponses{}. The weaving family, the second cluster the original paper names, is a small minority (\JiangWeavingFamilyShareResponses{} of responses; \JiangWeavingFamilyShareNoQwQ{} without the brainstorm model).
The reading induces $\JiangNDistinctImages$ distinct vehicles. Without the one model whose $\JiangNQwQResponses$ brainstorm-list responses supply \JiangQwQShareMarks{} of the vehicle--response marks, it induces $\JiangNDistinctImagesNoQwQ$. Under either count the shape matches our pool: one dominant vehicle plus a heavy tail of distinct minority vehicles.
Two differences between the pools do not affect this test, which is internal to their pool: the prompt wording (``involving'' versus ``about,'' Appendix~\ref{app:prompt-wording-check}) and the label seeding, which changed three top label names but no count (their \texttt{\JiangSecondImage} and \texttt{tapestry} are our \texttt{pickpocket} and \texttt{weaver}; Appendix~\ref{app:concept-label-details}).
The two-cluster characterization thus fails on the original models under both instruments, mixture modeling and text-only vehicle labels. We do not claim model responses are diverse; we test only that characterization.

\section{Undemanding and Missing Nulls}
\label{sec:wrong-null}

Homogeneity is a comparative property: responses can only be homogeneous relative to some reference distribution. \citet{jiang2025artificial} quantifies the Artificial Hivemind through cosine similarities between embedded responses (their Figs.~4--6) and the number of unique models contributing to each query's most-similar responses (their Fig.~8). For the former, the paper supplies a single reference point: pairs of responses answering \emph{different} prompts. For the latter, it supplies none.

\paragraph{Experimental Setup.}
This section's analyses use \citet{jiang2025artificial}'s own model responses, privately shared by the authors (the 43 of their 70+ models whose generations were available), rather than the 15-model pool we generated for Section~\ref{sec:artificial_hivemind_figure1}. The prompt set is \textsc{Infinity-Chat-50}: the 50-prompt annotation subset of the original paper's \textsc{Infinity-Chat-100} benchmark. \citet{jiang2025artificial} mined the \textsc{Infinity-Chat} prompts from WildChat \citep{zhao2024wildchat} and contributed the filtering, taxonomy, and human annotations. Appendix~\ref{app:null-analysis-details} gives this section's methods. Appendix~\ref{app:null-pair-sampling} states which statistics use sampled pairs and which use all pairs.

\findingpar{finding:null-undemanding}{: The Null Distribution Is Undemanding; Any Reasonable Conditional Generator Beats It.}
We replicated the paper's null. We embedded cross-prompt response pairs with the paper's embedding model (OpenAI's \texttt{text-embedding-3-small}) and measured a median cosine similarity of ${\sim}0.11$, consistent with the paper's reported 0.1--0.2 range (Appendix~\ref{app:null-pair-sampling}).
Against this null, the paper's 0.8 convergence threshold sits at the 99.95th percentile, making convergence appear meaningful.
This null compares, for example, a response to \emph{``Come up a short blurb to introduce a religion called The Next Exodus Society.''} with a response to \emph{``If there were double the amount of oxygen in the air, what would happen? Write in 100 words.''} Such pairs share no topic, vocabulary, or format.
Any reasonable conditional generator, model or human, beats this baseline. It answers in the prompt's language, stays on topic, and uses a chat register. Each of these raises similarity before any question of model convergence arises.
The null shows only that coherent, on-topic text is close to other coherent, on-topic text. It does not test the Hivemind hypothesis.

\findingpar{finding:null-demanding}{: Under a More Demanding Null, the Margin Shrinks.}

\begin{table}[t!]
\centering
\caption{\textbf{The Hivemind's Convergence Threshold Nearly Coincides with the Floor Set by Genuinely Different Ideas.} Median cosine similarity under \texttt{text-embedding-3-small} for the paper's null (responses to different prompts) versus our two constructions of a more demanding null (same-prompt responses expressing genuinely different ideas, determined using either geometric clustering or LLM labeling).}
\label{tab:null-comparison}
\begin{tabular}{lcc}
\toprule
Null distribution & Median & Gap to 0.8 threshold \\
\midrule
Different prompts \citep{jiang2025artificial} & ${\sim}0.11$ & ${\sim}0.69$ (99.95th pctl) \\
\midrule
Same prompt, genuinely different ideas (ours): & & \\
\quad geometric clustering, 50 prompts & $0.723$ & ${\sim}0.08$ \\
\quad LLM-labeled, 15 prompts & $0.703$ & ${\sim}0.10$ \\
\bottomrule
\end{tabular}
\end{table}

The question is whether same-prompt responses are more similar than prompt conditioning alone induces.
A null that answers it is the similarity of same-prompt responses that express \emph{genuinely different ideas}, e.g., a river-of-time metaphor paired with a thief-stealing-hours metaphor, both answering ``Write a metaphor involving time.'' Such pairs satisfy every expectation above (same language, topic, approximate length, chat register); their similarity estimates the floor set by prompt conditioning alone.
We measure this floor in two ways:
\begin{enumerate}
    \item \emph{Geometric clustering (all 50 prompts).} We identified clusters in each prompt's response pool with HDBSCAN \citep{campello2013density} (spectral-clustering fallback for degenerate solutions; Appendix~\ref{app:null-pair-sampling}) and computed across-cluster similarity for all 50 prompts, keeping only pairs of responses in \emph{different} clusters. Median across-cluster cosine: $0.723$. Its clusters come from the same geometry used to measure similarity, so we treat it as an embedding-internal baseline and check it against the LLM labels. The two agree on the similarity level of different-idea pairs, not on how many clusters or ideas there are; counts play no role in this null. The vehicle-label test below uses a separate labeling (Appendix~\ref{app:concept-label-details}). Appendix~\ref{app:null-pair-sampling} audits the fallback partition and label granularity.
    \item \emph{LLM labeling (15 prompts).} As a second labeling method, Claude Fable 5 assigned each response to one core idea. It reads the text alone and uses no embedding geometry. It covers 100 responses per prompt across 15 prompts. Labeler-independence details are in Appendix~\ref{app:null-pair-sampling}. Median across-idea cosine: $0.703$.
\end{enumerate}
Table~\ref{tab:null-comparison} contains the results. Genuinely different ideas answering the same prompt embed at 0.70--0.72 similarity under both constructions. The prompt-specific expectations of Finding~\ref{finding:null-undemanding} (shared language, topic, length, register) thus account for much of the ${\sim}0.7$ floor.
At the pair level, the paper reports that under min-$p$ decoding ``81\% of response pairs still exceed 0.7 similarity and 61.2\% exceed 0.8'' (its default-decoding figures are at least as high).
Under LLM labeling, our different-idea pairs exceed the same thresholds $54\%$ and $20\%$ of the time. Under geometric clustering, our different-cluster pairs exceed them $61\%$ and $32\%$ of the time. A pair above either threshold can simply be a river metaphor and a thief metaphor.
A gap remains. At 0.8, the paper's same-prompt pairs exceed the threshold roughly two to three times as often as our different-idea pairs. The ratio is 1.9 under geometric clustering and 3.1 under LLM labeling.
The convergence the paper measures is real but smaller than its null makes it appear. The large gap reflects the weakness of the cross-prompt null.
Appendix~\ref{app:per-prompt-floors} reports the per-prompt view, which agrees: roughly half of prompts' different-idea floors already exceed 0.7 (30 of 50 under geometric clustering, 8 of 15 under LLM labeling).
Medians alone understate the overlap, so we also summarize how far the same-idea and different-idea cosine distributions overlap (Appendix~\ref{app:pair-cosine-overlap}). Pairwise cosine, including at the 0.8 threshold, is a modest predictor of whether two responses express the same idea.

One caveat: our floor is measured on model responses, so any model-wide writing style (register, length, phrasing habits) that raises similarity between different ideas is part of it. Within-model cross-prompt pairs, which share style but no topic, score $0.111$ versus $0.109$ across models (Appendix~\ref{app:null-pair-sampling}), so style alone adds essentially nothing without a shared topic. Whether it adds something within a topic cannot be resolved without a human baseline.

\paragraph{Vehicle Labels Test the Two-Cluster Claim Directly.}
\citet{jiang2025artificial} reported that time-metaphor responses ``form just two primary clusters'' (``a river'' and ``a weaver'').
This is a claim about \emph{content}. The direct test is therefore the distribution of responses over vehicles, not the number of density modes.
Finding~\ref{finding:vehicle-labels} (Section~\ref{sec:artificial_hivemind_figure1}) reports this test on both pools, with the statistic defined there.

The baseline problem is not confined to the cosine analysis; the paper's remaining measurements
have no null at all.

\paragraph{A Note: No Human-Response Baseline Is Measured, so ``Homogeneous'' Has No Referent.}
The Hivemind's safety framing is a displacement claim: model outputs are narrower than the human expression they will replace, so repeated exposure flattens collective creativity. Testing this requires knowing how diverse \emph{human} responses to the same prompts are. The paper does not measure this.

The missing baseline matters because the flagship example, ``time is a river,'' is among the oldest and most dominant human metaphors for time, from Heraclitus to the \textsc{time is motion} mapping of cognitive linguistics \citep{johnson1980metaphors}.
A model pool that concentrates on the river metaphor is consistent with two different hypotheses: pathological collapse or a faithful fit to human expression.
A matched human-response collection is the most valuable follow-up experiment to \citet{jiang2025artificial}, though beyond our scope. It would also show whether model vehicle distributions are more concentrated than human ones (the top-1 vehicle share is the natural statistic).
Building a genuinely human response pool is hard today: crowdsourced responses may themselves be LLM-generated, so a credible benchmark needs safeguards such as attention checks and provenance verification.
We collected a small three-prompt pilot of crowdsourced human responses but suspected a majority of LLM-generated submissions, so we do not report it as a baseline.
Even the paper's most vivid evidence (two models returning character-identical responses) needs this baseline: identical answers are collisions, and collisions are common when a prompt's answer distribution has low entropy (e.g., humans asked to complete ``as busy as a \_\_\_'' will also collide).

\findingpar{finding:indistinguishability}{: The Model-Indistinguishability Statistic Has No Null, and Plausible Anchors Bracket It on Both Sides.}
Figure~8 of \citet{jiang2025artificial} is the paper's best-designed measurement because it is internally controlled: pool all $25 \times 50 = 1{,}250$ responses to a query, take the top-$N$ most mutually similar, and count the unique models represented. The authors find ${\sim}8$ unique models at $N{=}50$ and read this as strong cross-model convergence.

The paper states only the collapse anchor: perfectly disjoint models would yield 1. It does not state the opposite anchor. If model identity carried no information (maximal mixing), a uniform draw of 50 responses from the pool would contain about $21.9$ unique models on average (derivation and its heuristic-anchor caveat in Appendix~\ref{app:null-analysis-details}). The observed 8 lies between 1 and ${\sim}22$: closer to collapse than to full mixing, but far from either. The statistic needs a proper null: the expected count when 25 genuinely diverse generators answer the same prompt. That remains future work.
Without a null, this statistic does not discriminate a Hivemind from ordinary same-topic clustering.

\section{Are Training-Time Interventions Necessary? An Unsupported Premise, No Positive Evidence; Prompting Raises Measured Diversity}\label{sec:inference-time}

\citet{jiang2025artificial} asked whether the Hivemind can be countered at inference time or requires training-time intervention.
The difference matters: inference-time fixes are cheap, fast, and available to anyone; training-time fixes are slow, expensive, and available only to model developers.
The paper's conclusion is a claim about generalizability and who must act: ``more generalizable solutions are needed at the model training level to robustly preserve output diversity without requiring user intervention.''

This training-level conclusion is unsupported in three ways (Finding~\ref{finding:training-level}). We then show that a cheap inference-time intervention that providers can apply without user intervention, prompting, reliably raises measured response diversity above baseline across twelve topics and fifteen models (Finding~\ref{finding:rewordings}).

\begin{figure}[t!]
    \centering
    \includegraphics[width=\linewidth]{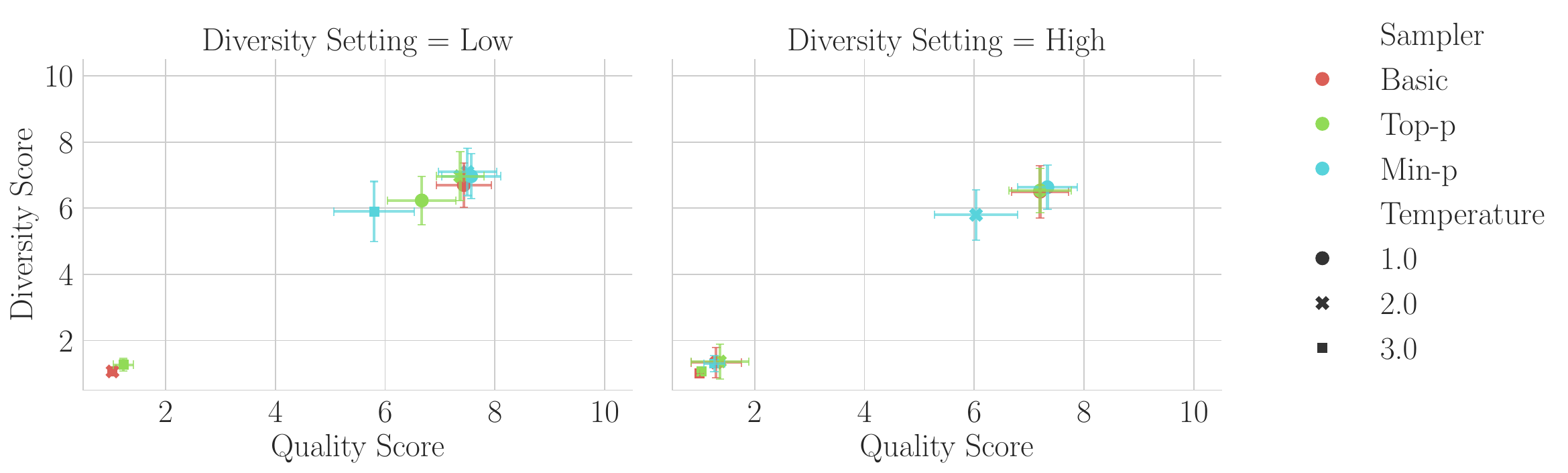}
    \caption{\textbf{An Unsupported Premise: The Min-$p$ Authors' Own Human Evaluation Indicates Min-$p$ Sampling Does Not Increase Diversity.} The visualization is adapted from Fig.~3 of \citet{schaeffer2025minpmaxexaggerationcritical}, but the underlying data originate from the min-$p$ paper itself: the plot shows human-evaluation scores that \citet{minh2025turning} collected and released, from an evaluation whose design, subjects, and judgment pipeline are the min-$p$ authors' own, independent of the present analysis. The plot shows the second of their two human evaluations. Reanalyzed, neither of their two studies shows min-$p$ outperforming basic or top-$p$ sampling on quality, diversity, or their tradeoff. In the second study, min-$p$ leads only where its own quality and diversity are lower than in other conditions \citep{schaeffer2025minpmaxexaggerationcritical}.
    \citet{jiang2025artificial}'s finding is more consistent with the explanation that min-$p$ \citep{minh2025turning} does not increase diversity, rather than the conclusion that training-time interventions are necessary.}
    \label{fig:min-p-human-eval}
\end{figure}

\findingpar{finding:training-level}{: The Conclusion That Mitigation Must Happen at Training Time Does Not Follow From the Evidence.}

\citet{jiang2025artificial} reasoned as follows. They chose min-$p$ \citep{minh2025turning}, a sampling algorithm claimed to increase output diversity. They found that homogeneity largely persisted: ``81\% of response pairs still exceed 0.7 similarity and 61.2\% exceed 0.8, revealing mode collapse even under diversity-oriented decoding''. They then drew the training-level conclusion quoted above.
The conclusion is unsupported in three ways.
\begin{enumerate}
    \item \textbf{The inference is a logical jump.} Even taking the min-$p$ result at face value, one sampler's failure to reduce homogeneity says nothing about other inference-time interventions.

    \item \textbf{The paper offers no positive evidence for training-level solutions.} It never demonstrates that any training-level change reduces homogeneity, so even a genuine failure of every tested inference-time intervention would not identify training as the remedy.

    \item \textbf{The cited premise fails on the min-$p$ authors' own data.} Min-$p$'s failure is informative only if min-$p$ increases diversity. The min-$p$ authors' own evidence does not support that premise. \citet{schaeffer2025minpmaxexaggerationcritical} reanalyzed the human-evaluation data that the min-$p$ authors collected and released, and found three problems. The original human evaluation omitted one third of the collected scores. It relied on statistical tests that did not hold up on reexamination. It mischaracterized annotator feedback that favored a baseline sampler more often than min-$p$. Reanalyzed, the min-$p$ authors' own scores show no advantage for min-$p$ over baseline samplers on quality, diversity, or the tradeoff between the two. The min-$p$ authors added a second human evaluation, with a different implementation, task, and rubric, to their camera-ready paper. They read that study as confirming min-$p$. Reanalyzed, that second study again shows no advantage (replotted in Fig.~\ref{fig:min-p-human-eval}). The reanalysis found that none of the min-$p$ paper's four lines of support survives. A sampler with no demonstrated diversity benefit failing to reduce homogeneity is uninformative about inference-time interventions in general.
\end{enumerate}
The paper admits both a strong reading (inference-time interventions cannot reduce mode collapse) and an accurate, weaker reading (solutions should generalize and not require user intervention); Appendix~\ref{app:training-level-two-readings} works through both and shows the conclusion is unsupported under either.
The prompt-phrasing result below is counterevidence against the strong reading: it shows the measured diversity metrics are readily movable at inference time (though we do not resolve output quality).

\findingpar{finding:rewordings}{: Cheap Rewordings Reliably Increase Model Output Diversity.}

Prior work already shows that prompt-level interventions raise measured diversity \citep{meincke2024prompting, cegin2024effects, kambhatla2025measuring, zhang2026verbalized}. One benchmark found a system prompt that asks for novelty only marginally effective under a distinctness classifier \citep{zhang2025noveltybench}. Our own idea-count probe below agrees with that result. Rewording moves the embedding metrics and the modal-idea share, not the number of countable ideas. We do not claim this finding is new. It is counterevidence to the strong reading. It adds a test on the original paper's own task, embedding model, and similarity statistic. That statistic, Rao's Q, is one of our three metrics. We fixed the phrasings ex ante. The test spans twelve topics and fifteen models and checks the length confound.
Inference-time interventions include at least two independent choices: the sampling algorithm and the prompt.
\citet{jiang2025artificial} tested the sampler but not the prompt.
Prompt-level interventions also need not burden users, e.g., model providers can set system prompts and rewrite prompts in the product layer with no user intervention. 
\paragraph{Experimental Setup.}

Using the same sampling hyperparameters as Section~\ref{sec:artificial_hivemind_figure1}, we test five additional prompt phrasings across the same twelve topics and fifteen models:
\begin{enumerate}
    \item \texttt{baseline}: ``Write a metaphor about \{topic\}'' (\citet{jiang2025artificial}'s main text phrasing).
    \item \texttt{creative\_adjective}: ``Write a creative metaphor about \{topic\}''.
    \item \texttt{quality\_anchor}: ``Write an original, memorable metaphor about \{topic\} that avoids clich\'es.''
    \item \texttt{unexpected\_domain}: ``Write a metaphor about \{topic\}, drawing inspiration from an unexpected domain. Bonus points for creativity and originality.''
    \item \texttt{poet\_persona}: ``You are a poet known for startling, unexpected imagery that avoids clich\'es. Write a metaphor about \{topic\}.''
    \item \texttt{anti\_modal}: ``Write a metaphor about \{topic\}. Before responding, consider what metaphor most language models would produce for this prompt, then deliberately choose a different approach.''
\end{enumerate}
We chose these five phrasings ex ante, expecting them to improve diversity scores.
For each (topic, phrasing, model) tuple, we sample 50 responses.

We measured diversity via the response embeddings (OpenAI \texttt{text-embedding-3-small}; Google \texttt{EmbeddingGemma-300m} results agree qualitatively, Appendix~\ref{app:diversity-robustness}). We use three metrics that probe different facets of the embedding geometry. \emph{Effective rank} \citep{royvetterli2007} and \emph{participation ratio} \citep{gao2017,belldean1970} are eigenspectrum measures of how many dimensions the responses use. \emph{Rao's quadratic entropy} \citep{rao1982} is the mean pairwise dissimilarity among responses.

Each metric is reported at three nested scales from ecology \citep{whittaker1972,ricottaSzeidl2009}: $\alpha$ (within-model diversity, averaged across the 15 models), $\beta$ (diversity among the 15 model centroids; for Rao's Q the centroids are renormalized to the unit sphere), and $\gamma$ (diversity of the full 750-response pool per cell). The decomposition separates each model's own diversity from that arising when models sample different regions of output space (Appendix~\ref{app:alpha-beta-gamma-primer} provides definitions).

\begin{figure}[t!]
\centering
\includegraphics[width=\linewidth]{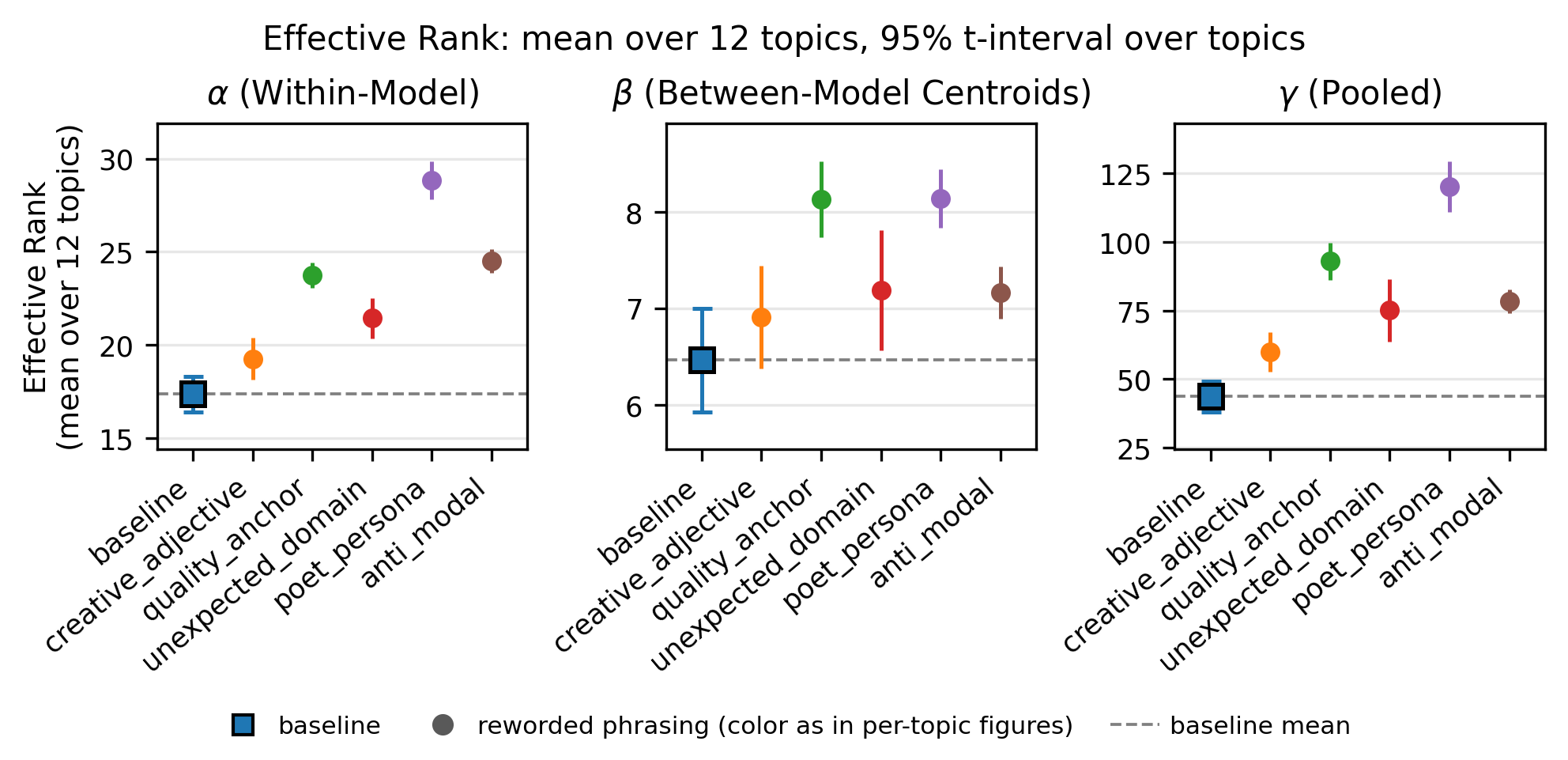}
\caption{\textbf{Effective Rank Rises Under Every Prompt Phrasing at Every Scale.} Effective rank (exponential of the Shannon entropy of the normalized covariance eigenspectrum) at the $\alpha$, $\beta$, and $\gamma$ scales. Each value is the mean over the 12 topics for one of six phrasings. The square marks baseline and the dashed line marks its mean. Error bars are 95\% $t$-intervals over the 12 topics.
All five non-baseline phrasings raise effective rank over \texttt{baseline} at every scale (mean across 12 topics). \texttt{Poet\_persona} is strongest at $\alpha$ ($17.4 \to 28.9$, $+66.1\%$) and $\gamma$ ($43.7 \to 120.2$, $+175.0\%$) and leads at $\beta$ ($6.47 \to 8.14$, $+25.9\%$). Per-topic values with per-topic jackknife intervals are in Fig.~\ref{fig:prompt-phrasing-er-per-topic} (Appendix~\ref{app:prompt-phrasing-secondary-metrics}). Unlike Rao's Q, effective rank rises at $\beta$ because it depends on the eigenspectrum, not the centroid norm.
}
\label{fig:prompt-phrasing-er}
\end{figure}

Each of the five phrasings, chosen ex ante with no tuning against results, outranks baseline in 79.6--91.7\% of the 108 topic-scale-metric cells (12 topics $\times$ 3 scales $\times$ 3 metrics, primary embedding; the secondary embedding agrees, Appendix~\ref{app:diversity-robustness}). The strongest phrasing, \texttt{poet\_persona}, more than triples pooled participation ratio ($13.8 \to 44.8$; Fig.~\ref{fig:prompt-phrasing-pr}; effective-rank view in Fig.~\ref{fig:prompt-phrasing-er}; Rao's Q view in Appendix~\ref{app:prompt-phrasing-secondary-metrics}, Fig.~\ref{fig:prompt-phrasing-rao}). Sign tests paired over the 12 topics put most $\alpha$- and $\gamma$-scale gains at the exact-test floor; at $\beta$ the effect is weaker and absent under Rao's Q (Appendix~\ref{app:beta-rao-flatness}). Deliberate prompt engineering for a specific application would push these gains further. Appendix~\ref{app:prompt-phrasing-additional} reports the full significance statistics, metric-agreement analysis, and robustness analyses; none changes the conclusion.

The reworded prompts also change response length (median $0.74\times$ to $3.01\times$ baseline), a potential confound. Appendix~\ref{app:length-confound} shows length cannot explain the gains: the phrasing with the largest gain produces the \emph{shortest} responses.
Rising effective rank could reflect only more varied vocabulary and sentence structure, with no new ideas.
An LLM idea-labeling probe (Appendix~\ref{app:conceptual-vs-stylistic}) assigns each response to one core idea. The modal idea is the most common idea in a (topic, phrasing) cell.
Under the strongest rewordings, its share of classifiable responses falls from 0.48 to 0.34--0.35.
Its share of all responses falls from 0.39 at baseline to 0.18--0.20. However, the fraction of unclassifiable responses rises from 0.20 at baseline to 0.39--0.45 under the rewordings. A larger unclassifiable fraction lowers the all-response modal share by construction.
The drop in modal-idea concentration holds under either share.
No idea-count measure shifts significantly. The effective number of ideas is 3.34 at baseline versus 3.99 (\texttt{poet\_persona}) and 3.97 (\texttt{anti\_modal}). The smallest two-sided sign-test $p$ on that change is 0.39. The same confound limits the idea count (Appendix~\ref{app:conceptual-vs-stylistic}).
We therefore state Finding~\ref{finding:rewordings}'s gains as measured embedding-space diversity plus reduced modal-idea concentration among classifiable responses, not as more countable ideas.

We do not resolve whether these phrasings preserve or improve output quality, nor whether embedding-space diversity equals human-valued creativity. Our claim is narrower: simple inference-time prompt changes move the measured diversity metrics substantially, so the training-level conclusion is unsupported.

\paragraph{Our Pool Reproduces the Original Paper's Intra-Model Similarity Statistic.}
On the measurement that defines the intra-model side of the Hivemind, our data agree with the original paper's.
\citet{jiang2025artificial} report that the average pairwise similarity among same-prompt responses from one model ``typically exceeds 0.8'' (in 79\% of their query response pools).
Our baseline within-model Rao's Q ($Q_\alpha \approx 0.191$) converts to the off-diagonal mean pairwise cosine $\bar{c}$ (conversion in Appendix~\ref{app:alpha-beta-gamma-primer}).
Applied per (model, topic) cell and averaged over our 15 models and 12 topics in the same OpenAI \texttt{text-embedding-3-small} space, this yields $\bar{c} = 0.805$, consistent with the original report. Different prompt sets and model rosters preclude an exact match. A second embedding model agrees (per-topic and second-embedding detail in Appendices~\ref{app:per-topic-diversity-ranking} and \ref{app:jiang-43model-retest-details}).

We reproduce the measurement but dispute its interpretation. Without an idea-level floor (Finding~\ref{finding:null-demanding}), a mean cosine near 0.8 cannot distinguish conceptual collapse from prompt conditioning plus shared style, because same-prompt different-idea responses already have cosine ${\sim}0.7$.
A mean pairwise cosine of $0.8$ is also consistent with Finding~\ref{finding:vehicle-labels}: a pool of one dominant vehicle plus a minority of different vehicles and a pool of uniform mild rephrasings can produce the same mean. The statistic the paper reports cannot distinguish the two; the vehicle labels can, and they show the former.
The reproduction also answers Section~\ref{sec:artificial_hivemind_figure1}'s model-vintage confound: had the 2025--26 model generation fixed the measured homogeneity, this statistic would have dropped in our pool but did not.

\section{Discussion}\label{sec:discussion}

\paragraph{Summary of Findings.}\label{sec:discussion-summary}

Five lines of evidence undercut the Artificial Hivemind's flagship example: visualization, spectral, clustering, language model labels, and case selection.
Four of them say the two-cluster description is wrong.
Visualization: the original paper's own 2D view, replicated on our data, shows many groupings, not two.
Spectral: PC1+PC2 capture only 25--41\% of the variance. The eigenspectrum rules out a tight two-cluster collapse.
Clustering: three procedures on the full-dimensional embeddings select $K \gg 2$. Mixture modeling agrees on the original authors' own 43-model pool.
Vehicle labels, which use no embedding geometry, reject the two-cluster description on both pools. The structure is one dominant vehicle plus a heavy tail of distinct minority vehicles. The weaving family is a small minority on both pools (Tables~\ref{tab:ours-concept-distribution} and~\ref{tab:jiang-image-distribution}).
The fifth line is a different claim: ``time'' is among our least diverse topics, a favorable rather than representative case. It ranks first of twelve (least diverse) on both within-model statistics and third on pooled Rao's Q (Appendix~\ref{app:per-topic-diversity-ranking}, Table~\ref{tab:per-topic-diversity-ranking}).
The original paper calibrates its cosine-similarity metric against a null that any conditional generator beats. The ``convergence threshold'' sits only ${\sim}0.1$ above a floor that includes demonstrably non-convergent responses.
Much of the measured homogeneity is therefore the shared geometry of answering the same prompt. Only a matched human pool could establish the size of that share. A smaller residual effect survives our stricter null (Finding~\ref{finding:null-demanding}).
The case for training-level solutions does not hold. Min-$p$ has not been shown to increase diversity. Prompt phrasing, a deployable intervention the paper did not test, reliably increases measured diversity.

\paragraph{Related Work.}\label{sec:discussion-related-work}
\citet{gorecki2025monoculture} ask whether deployed models constitute a monoculture or a multiplicity, and \citet{bommasani2022picking} study whether algorithmic monoculture homogenizes outcomes.
On the behavioral side, \citet{padmakumar2024does} find that writing with an instruction-tuned model reduces the resulting text's content diversity, \citet{kirk2024understanding} show that RLHF reduces output diversity relative to supervised fine-tuning, and \citet{santurkar2023whose} document that models over-represent particular opinion distributions.
On the measurement side, contextual embedding spaces are anisotropic, with high baseline cosine similarity between random words \citep{ethayarajh2019contextual}, and cosine similarity of learned embeddings can be an unreliable proxy for semantic similarity \citep{steck2024cosine}; both bear directly on the threshold-based convergence claims we re-examine here.

\citet{schaeffer2025positionmachinelearningconferences} argue for a refutations track at machine-learning venues; this paper is an example.

\paragraph{Reevaluating Implications for Society and AI Safety.}\label{sec:discussion-ai-safety}

\citet{jiang2025artificial} frames language-model homogeneity as posing ``long-term AI safety risks'' and argues the remedy must come at the training level. Our results revise both.
The empirical case for a dangerous Artificial Hivemind is inconclusive.
Concerns about AI-driven homogenization of human thought deserve evidence that survives independent re-examination under strong baselines; the current instruments do not yet provide it.
We do not dispute that the labels concentrate on one vehicle on both pools.
We dispute the two-cluster description.
Section~\ref{sec:artificial_hivemind_figure1} states why that shape bears on the remedy.
The conclusion against inference-time mitigation rests on an unsupported premise: min-$p$ has never been shown to increase diversity, so its failure to reduce homogeneity says nothing about inference-time interventions generally.
Prompting, the cheaper intervention, already moves the metrics: asking for more creative outputs raises measured diversity over baseline.
With no positive evidence for training fixes, the call for expensive training interventions is premature.

\paragraph{Limitations.}\label{sec:discussion-limitations}
First, ours is a \emph{conceptual} replication with current models, not a direct one. The model-vintage difference is a genuine confound. We address it in two ways. We rerun the clustering analyses on the original authors' own 43-model responses, where clustering methods recover more than two clusters at essentially every setting we tested. (This shared pool, 43 open-weight models of their 70+ rather than the 25-model pool behind the flagship figure, is geometrically coarser than ours at matched sample size.) We also show that our pool reproduces the original intra-model similarity statistic. Second, we do not benchmark model diversity against a matched human-response pool. Third, embedding-space diversity is not human-valued creativity or quality: our geometric measures capture dispersion in representation space, a proxy for, not a definition of, the qualities readers care about. Fourth, the prompt rephrasings that raise measured diversity may trade diversity against quality; quality under the rephrased prompts remains unresolved, so our claim is that the diversity metrics move, not that quality is preserved.

\paragraph{Toward a Validated Diversity Instrument.}\label{sec:discussion-metric-validity}
No instrument for measuring semantic homogeneity has been validated at the scale and on the prompt distribution at which homogeneity claims are now made.
Embedding cosine similarity is used as the measurement without having been shown to track what it claims to measure. The anisotropy and semantic-proxy results cited above \citep{ethayarajh2019contextual, steck2024cosine} give specific reasons to doubt that it does. A few rogue dimensions dominate cosine in contextual embedding spaces, and standardizing them away improves agreement with human similarity judgments \citep{timkey2021all}.
We adopt \citet{jiang2025artificial}'s metric deliberately, so that our findings hold on the original paper's own terms.
Our clustering conclusion does not rest on that instrument alone. Vehicle labels induced from the text of our own \OursConceptPoolSize{} time-metaphor responses and of the original authors' \JiangConceptPoolSize{} use no embedding geometry. They reject the same two-cluster description on both pools, finding one dominant vehicle plus a heavy tail of distinct minority vehicles (Section~\ref{sec:artificial_hivemind_figure1}, Tables~\ref{tab:ours-concept-distribution} and~\ref{tab:jiang-image-distribution}).
No validated replacement exists: a language model rating the similarity of language-model outputs is circular, since judges prefer text resembling their own generations \citep{panickssery2024llm, zheng2023judging}, and judge ratings shift with prompt phrasing and choice of judge \citep{stureborg2024large}.
Lexical and $n$-gram measures capture surface form rather than content \citep{tevet2021evaluating}, and human annotation does not scale to millions of responses, though \citet{zhang2025noveltybench} trains a distinctness classifier on a purpose-built pool of human-written responses, reaching 0.81 AUC on held-out human equivalence labels.
The gap is a fact about the state of the field, not a defense of the current instrument.
The constructive step is construct validation \citep{jacobs2021measurement}: collect human similarity judgments on matched sets of responses, then test which candidate tracks them (embedding cosine at various thresholds, LLM-judge ratings, lexical diversity, effective number of human-annotated categories).
\citet{tevet2021evaluating} already ran a version of this experiment, and the result supports our skepticism. Sentence-BERT cosine was the best automatic content-diversity metric but still trailed human raters. Every neural metric degraded on their subset where form diversity is neutralized by construction, so cosine's content sensitivity depends partly on form variation.
What has not been done is that exercise at today's scale: open-ended real-user prompts, current embedding models, many responses per prompt, and a similarity threshold calibrated against human judgments rather than fixed by convention.
Until some instrument passes that test, ``models converge above threshold $X$'' is a claim about geometry, not about ideas.

\paragraph{Future Directions.}\label{sec:discussion-future-directions}
Language model homogeneity deserves rigorous study. The most valuable next steps are clear: a matched human-response pool, a diversity instrument validated against human judgments, a proper null for the model-indistinguishability statistic, and tail-sensitive summaries in place of scalar means. Any of these could strengthen or weaken the Hivemind case.
Our contribution is to clarify what the current instruments can and cannot establish. We hope this reanalysis, which the original authors' data sharing made possible, leads to better measurement and clearer understanding.

\paragraph{Data and Code Availability.}\label{sec:discussion-data-availability}
\ifdefined\arxivversion
Our code and data are available from the authors on request.
These materials include the response-generation code, the ${\sim}9{,}000$ responses of Section~\ref{sec:artificial_hivemind_figure1}, and the prompt-phrasing responses of Section~\ref{sec:inference-time}.
They also include the embedding-generation scripts and the clustering, null-analysis, and figure-generation code.
These materials let a reader recompute every number that rests on our own responses.
\else
An anonymized archive accompanies this submission as supplementary material.
It holds the response-generation code, the ${\sim}9{,}000$ responses of Section~\ref{sec:artificial_hivemind_figure1}, and the prompt-phrasing responses of Section~\ref{sec:inference-time}.
It also holds the embedding-generation scripts and the clustering, null-analysis, and figure-generation code.
The archive lets a reader recompute every number that rests on our own responses.
\fi
Numbers that rest on the original authors' responses need their data.
The original authors privately shared the response data from \citet{jiang2025artificial} used in Section~\ref{sec:artificial_hivemind_figure1} and Section~\ref{sec:wrong-null}.
Requests for that data should go to them.

\clearpage
\bibliographystyle{tmlr}     %
\bibliography{references}

\clearpage
\appendix
\section{The 15-Model Roster}\label{app:model-roster}

Table~\ref{tab:fifteen-model-roster} lists the 15 API models used for our generations in Sections~\ref{sec:artificial_hivemind_figure1} and \ref{sec:inference-time}, with the exact API identifier strings recorded in our data.
None of the identifiers encodes a release date, so we omit release dates rather than guess them.
Two of the model families, GPT-4o and GPT-4o Mini, appear in \citet{jiang2025artificial}'s own evaluations and therefore predate that paper's circulation.
Pretraining and post-training pipelines take months, so a paper's circulation cannot quickly change the models providers deploy.
Sampling used temperature $=1.0$ and top-$p=0.9$ throughout, with one provider-specific exception. The Anthropic API advises against setting top-$p$ and temperature together, so for Anthropic models we set only temperature $=1.0$.

\begin{table}[h]
\centering
\caption{\textbf{Our 15-Model Pool Spans Six Anthropic, Five OpenAI, and Four Google Models with Pinned API Identifiers.} The API identifier is the exact string passed to each provider's API (recorded verbatim in our data). Release dates are omitted because none of the identifiers encodes a date.}
\label{tab:fifteen-model-roster}
\begin{tabular}{lll}
\toprule
Model & Provider & API identifier \\
\midrule
Claude Opus 4.6 & Anthropic & \texttt{claude-opus-4-6} \\
Claude Opus 4.5 & Anthropic & \texttt{claude-opus-4-5} \\
Claude Sonnet 4.6 & Anthropic & \texttt{claude-sonnet-4-6} \\
Claude Sonnet 4.5 & Anthropic & \texttt{claude-sonnet-4-5} \\
Claude Sonnet 4.0 & Anthropic & \texttt{claude-sonnet-4-0} \\
Claude Haiku 4.5 & Anthropic & \texttt{claude-haiku-4-5} \\
\midrule
GPT-4.1 & OpenAI & \texttt{gpt-4.1} \\
GPT-4.1 Mini & OpenAI & \texttt{gpt-4.1-mini} \\
GPT-4.1 Nano & OpenAI & \texttt{gpt-4.1-nano} \\
GPT-4o & OpenAI & \texttt{gpt-4o} \\
GPT-4o Mini & OpenAI & \texttt{gpt-4o-mini} \\
\midrule
Gemini 3.1 Pro (preview) & Google & \texttt{gemini-3.1-pro-preview} \\
Gemini 3 Pro (preview) & Google & \texttt{gemini-3-pro-preview} \\
Gemini 3 Flash (preview) & Google & \texttt{gemini-3-flash-preview} \\
Gemini 2.5 Flash & Google & \texttt{gemini-2.5-flash} \\
\bottomrule
\end{tabular}
\end{table}

\section{Verifying the Original Paper's Exact Prompt Wording}\label{app:prompt-wording-check}

Section~\ref{sec:introduction} describes our prompt as closely matching the original paper's. The original paper uses two wordings of the flagship prompt. Its main text and Figure 1 caption give ``Write a metaphor about time.'' Its released prompt file, the title text in its own Figure 1 (Fig.~\ref{fig:jiang-figure1}), and the \texttt{"prompt"} field of its privately shared response data all use ``Write a metaphor involving time.'' We therefore treat ``involving'' as the prompt the original responses actually answered. This appendix checks whether the wording difference affects our findings.

\paragraph{Methodology.}

We tested only the flagship topic ``time,'' not all 12 topics of Section~\ref{sec:artificial_hivemind_figure1}.
We used the prompt ``Write a metaphor involving time''.
From the same 15 models as Section~\ref{sec:artificial_hivemind_figure1} (roster in Appendix~\ref{app:model-roster}), we collected 50 responses per model with the same sampling settings: temperature $= 1.0$ and top-$p = 0.9$ (provider-specific exception in Appendix~\ref{app:model-roster}).
Between our original data collection and this check, providers retired two of the fifteen models, Claude Sonnet 4.0 and Gemini 3 Pro (preview). Both now return ``model not found'' errors on every request. The results below therefore use the thirteen models available under both prompt wordings (650 responses per wording, rather than 750).

\paragraph{Results.}

Once both wordings are restricted to the same thirteen models, most measures agree closely. The number of PCA components needed for 90\% cumulative variance differs by only 2--4 (73 vs.\ 78 across the two embedding models). HDBSCAN's cluster count is unchanged (K=10 under both wordings). Population-level ($\gamma$-scale) diversity is modestly higher under ``involving'' than ``about'' for both embedding models: effective rank rises by 3.4--4.6 and participation ratio by 0.9--1.1. This shift is larger than the change from dropping two models at fixed wording, so it is not a sample-size artifact. Per-model ($\alpha$-scale) diversity shows no shift of that size. Spectral eigengap clustering also finds fewer clusters under ``involving'' (K=8 vs.\ 13), but this rests on a single (topic, embedding, $k$) cell and is weak evidence. The vMF-BIC cluster count at 128 dimensions is unstable under either wording. Its BIC curve has several near-tied local optima (e.g., $K=32$, $66$, and $67$ within 1{,}200 BIC units of each other for the matched-N ``about'' subset). We therefore do not treat its swings as evidence of a wording effect. This check re-ran two of the five lines of Section~\ref{sec:artificial_hivemind_figure1}, spectral and clustering. Both hold under either wording. It re-ran no others. We did not re-plot the visualization line's 2D projections under ``involving.'' We did not recompute the vehicle labels of Appendix~\ref{app:concept-label-details} on this check's 650-response ``involving'' pool. (They do cover the original authors' own ``involving'' pool, which is a different set of models.) The case-selection line is a ranking across all twelve topics, which this single-topic check cannot test. This check therefore says nothing either way about those three lines. The wording difference produces a modest increase in aggregate (not per-model) diversity, checked only on this single topic.

\section{Per-Component PCA Variance Spectra on a Log-Log Scale}\label{app:pca-variance-loglog}

\begin{figure}[t!]
\centering
\includegraphics[width=\linewidth]{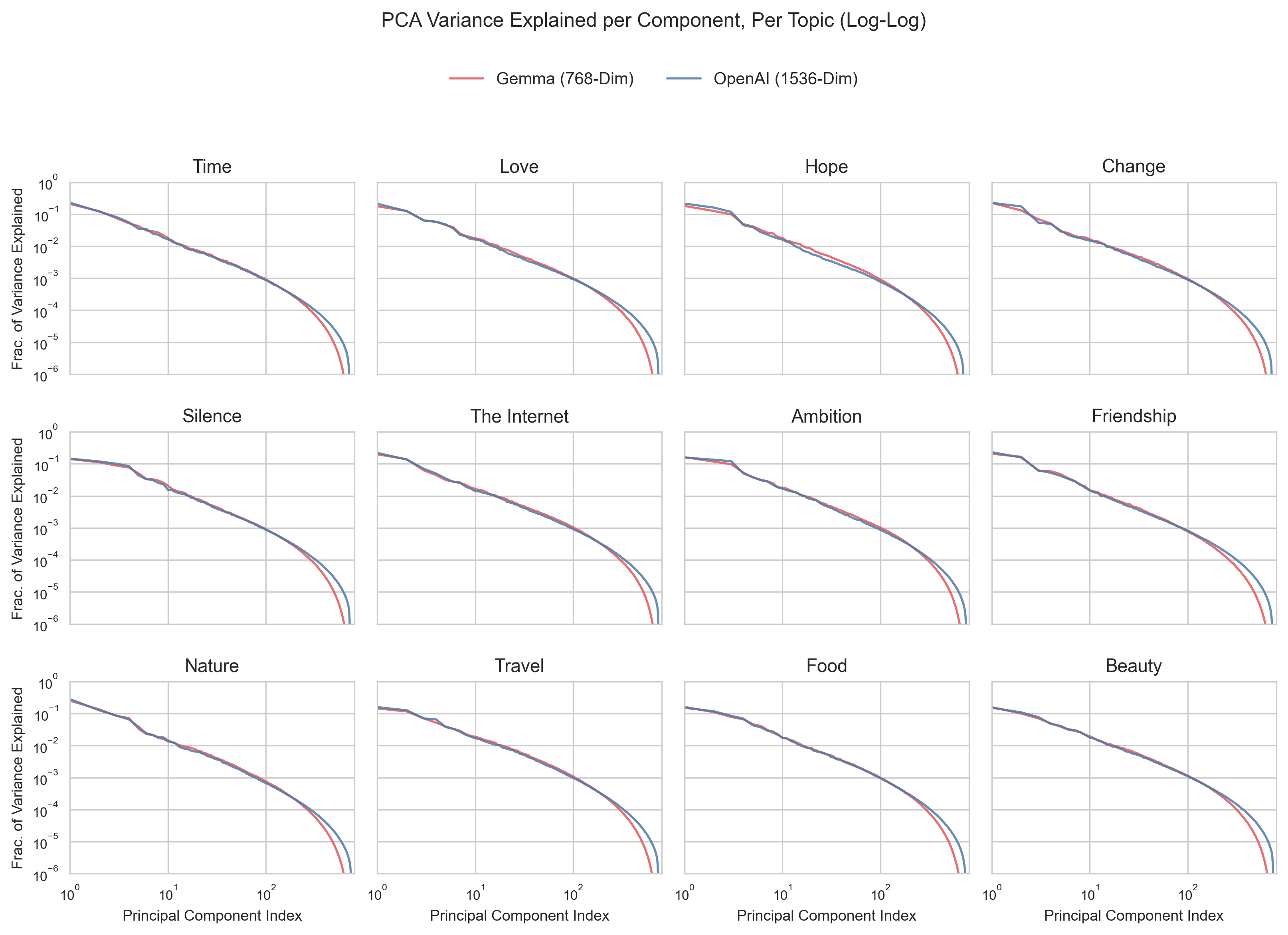}
\caption{\textbf{The Eigenspectrum Decays Smoothly on a Log-Log Scale, with No Sharp Elbow, for Every Topic.} Fraction of variance explained by each principal component (not cumulative), for all 12 baseline topics under both embedding models. Every topic shows a smooth decay across hundreds of components rather than a sharp drop after the first few, corroborating Fig.~\ref{fig:pca-spectrum}'s cumulative-variance view.}
\label{fig:pca-variance-loglog}
\end{figure}

Finding~\ref{finding:spectral} (Section~\ref{sec:artificial_hivemind_figure1}) summarizes the eigenspectra of the baseline response embeddings. The cumulative variance-explained curves appear in the main text (Fig.~\ref{fig:pca-spectrum}).
Fig.~\ref{fig:pca-variance-loglog} shows the per-component view: the fraction of variance explained by each principal component, plotted against component index on log-log axes. A few tight, well-separated clusters would produce a sharp drop after the first few components. On these axes a sharp elbow appears as an abrupt drop rather than a smooth line, so this plot is the standard diagnostic.
Finding~\ref{finding:spectral} reports 60--110 principal components for 90\% cumulative variance. Reaching 95\% requires 113--189 components.
Smooth spectral decay is generic for sentence-embedding clouds of natural text, so smoothness alone is weak evidence of diversity. The spectrum establishes the absence of the variance concentration that a tight two-cluster reading predicts.

\section{Clustering Procedure Details for Section~\ref{sec:artificial_hivemind_figure1}}\label{app:clustering-method-details}

\begin{figure}[t!]
\centering
\includegraphics[width=\linewidth]{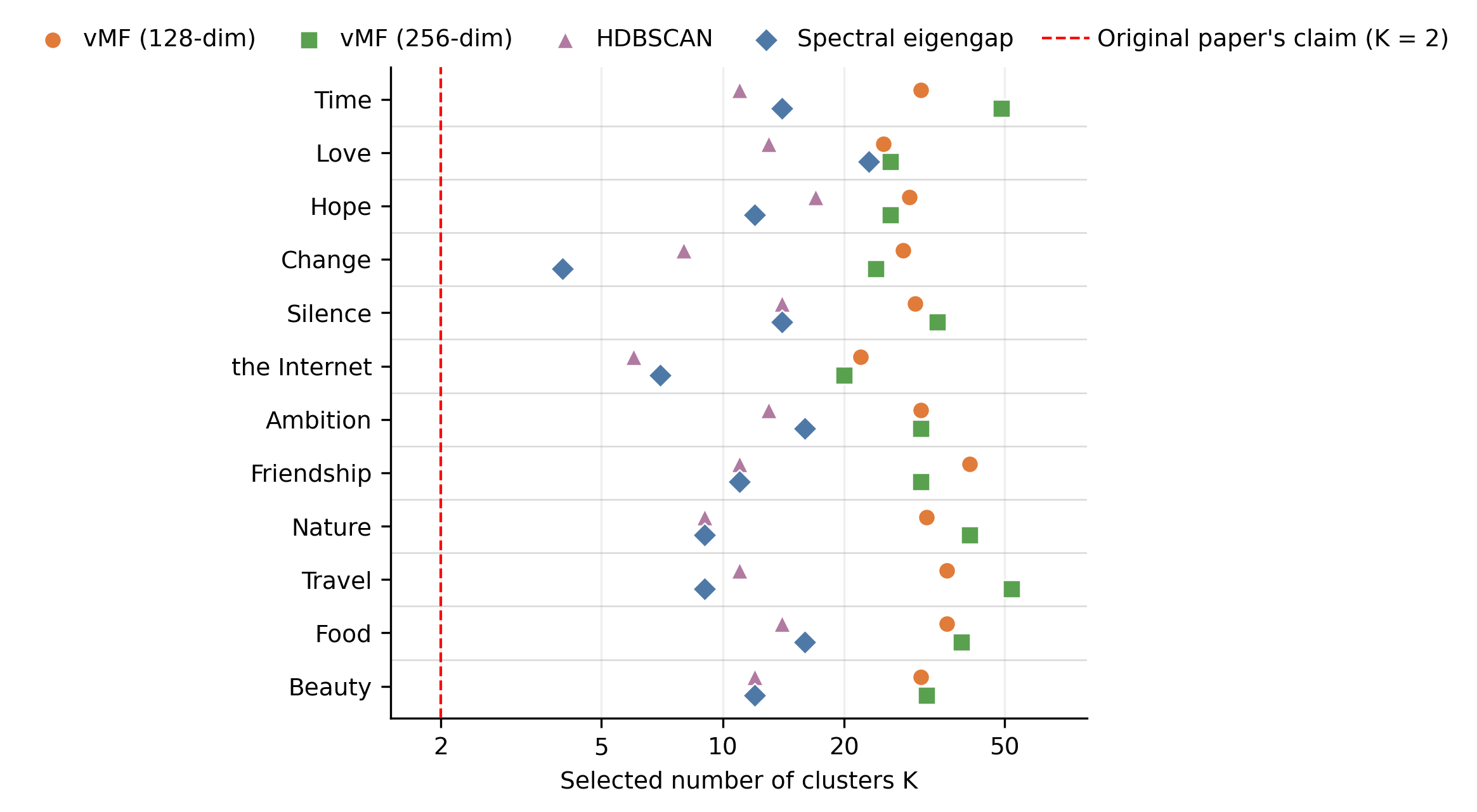}
\caption{\textbf{Selected Cluster Counts per Topic and Method.} Selected $K$ per topic for each of the four methods (vMF--BIC at 128 and 256 dimensions, HDBSCAN, and spectral eigengap) on the 12 baseline topics. The dashed line marks $K{=}2$. The main text's Fig.~\ref{fig:clustering-model-selection} summarizes these per-method distributions.}
\label{fig:clustering-model-selection-per-topic}
\end{figure}

Finding~\ref{finding:clustering} reports the cluster counts selected by three clustering procedures. This appendix records their exact configurations and caveats.
The three procedures are:
(1) von Mises--Fisher (vMF) mixture models \citep{banerjee2005clustering} on the 128- and 256-dimensional Matryoshka slices \citep{kusupati2024matryoshkarepresentationlearning}, renormalized to the unit hypersphere. We sweep the number of clusters $K$ from 1 to 750 and select $K$ by minimizing BIC.
(2) HDBSCAN \citep{campello2013density} with cosine distance at \texttt{min\_cluster\_size}~$= 15$ and \texttt{min\_samples}~$= 1$, on the 128-dimensional embeddings.
We fix \texttt{min\_samples}~$=1$, the setting within a reasonable range that yields the most clusters at fixed \texttt{min\_cluster\_size}; Appendix~\ref{app:hdbscan-sensitivity} sweeps both hyperparameters.
(3) Spectral eigengap analysis \citep{vonluxburg2007tutorial} on a cosine-similarity $k$-NN graph ($k = 20$), also on the 128-dimensional embeddings.
The three methods span probabilistic parametric (vMF), nonparametric density-based (HDBSCAN), and graph-based (spectral) paradigms.
Each carries a caveat. vMF--BIC can over-split in high dimensions. HDBSCAN counts are sensitive to \texttt{min\_cluster\_size}. Spectral eigengaps can be ambiguous. The main text therefore emphasizes the consistent direction of all three procedures ($K \gg 2$) rather than any exact count.
The full per-method ranges across the 12 baseline topics: the vMF mixture models found $K=22$--$41$ clusters per topic at 128 dimensions and $K=20$--$52$ at 256 dimensions; HDBSCAN found $K=6$--$17$ per topic; the spectral eigengap suggests $K=4$--$23$.
Fig.~\ref{fig:clustering-model-selection-per-topic} plots the selected $K$ per topic for all four methods.

\section{HDBSCAN Sensitivity to \texttt{min\_cluster\_size} and \texttt{min\_samples}}\label{app:hdbscan-sensitivity}

HDBSCAN's cluster count depends on both \texttt{min\_cluster\_size} and \texttt{min\_samples}. Our main-text analysis (Section~\ref{sec:artificial_hivemind_figure1}) fixes \texttt{min\_samples}~$=1$. We therefore report a sensitivity sweep on the same embeddings: \texttt{min\_cluster\_size} $\in \{3,\ldots,100\}$ $\times$ \texttt{min\_samples} $\in \{1,\ldots,10\}$, for all 12 baseline topics (11{,}760 fits). Table~\ref{tab:hdbscan-sensitivity} summarizes a coarser sub-grid of plausible settings.

\begin{table}[t!]
\centering
\caption{\textbf{HDBSCAN's Cluster Count Stays Far Above 2 Across a Wide Range of Hyperparameters.} Median [min--max] number of clusters $K$ across the 12 baseline topics, for a coarse grid of \texttt{min\_cluster\_size} (rows) and \texttt{min\_samples} (columns). The main-text operating point (\texttt{min\_cluster\_size}~$=15$, \texttt{min\_samples}~$=1$) is bolded.}
\label{tab:hdbscan-sensitivity}
\begin{tabular}{lcccc}
\toprule
\texttt{min\_cluster\_size} & \texttt{min\_samples}$=1$ & $=3$ & $=5$ & $=10$ \\
\midrule
5  & 32.5 [23--40] & 26.0 [16--34] & 19.0 [12--26] & 10.0 [7--16] \\
10 & 15.5 [11--20] & 14.0 [8--19]  & 13.0 [8--17]  & 9.5 [7--13] \\
\textbf{15} & \textbf{11.5 [6--17]} & 11.0 [2--15]  & 9.0 [2--14]   & 8.0 [4--11] \\
20 & 8.5 [2--12]   & 8.0 [2--11]   & 7.0 [4--10]   & 6.5 [4--10] \\
30 & 6.0 [2--9]    & 6.0 [2--10]   & 5.5 [2--10]   & 5.0 [4--7] \\
50 & 4.0 [2--6]    & 4.5 [2--6]    & 4.0 [2--5]    & 3.5 [2--5] \\
\bottomrule
\end{tabular}
\end{table}

Across this grid, median $K$ across the 12 topics never falls below 3.5. It stays at or above 8 for \texttt{min\_cluster\_size}~$\leq 15$ at every \texttt{min\_samples} tested. Ten of the twelve topics never produce $K \leq 2$ anywhere in the full sweep. Two are more sensitive: ``change'' (first at \texttt{min\_cluster\_size}~$=14$) and ``food'' (only at the single most conservative corner tested). At the operating point used in the main text, every topic yields $K$ between 6 and 17. Within our range, \texttt{min\_samples}~$=1$ yields the most clusters at fixed \texttt{min\_cluster\_size}. This choice affects how many clusters are found, not whether clusters are found. $K \leq 2$ occurs in only 11\% of the 11{,}760 fits, concentrated at \texttt{min\_cluster\_size} above roughly 5\% of the per-topic sample size ($n=750$). $K \gg 2$ is robust to reasonable choices of both hyperparameters. Sufficiently coarse settings can force degenerate solutions, as with any density-based method.

\section{Details of the Re-Test on the Original Paper's Own 43-Model Responses}\label{app:jiang-43model-retest-details}

\begin{figure}[t!]
\centering
\includegraphics[width=0.62\linewidth]{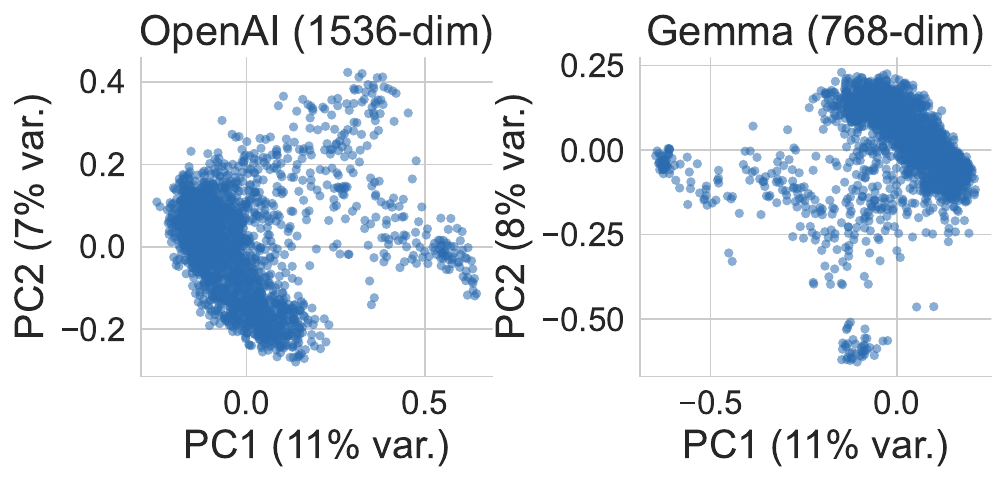}\hfill
\includegraphics[width=0.37\linewidth]{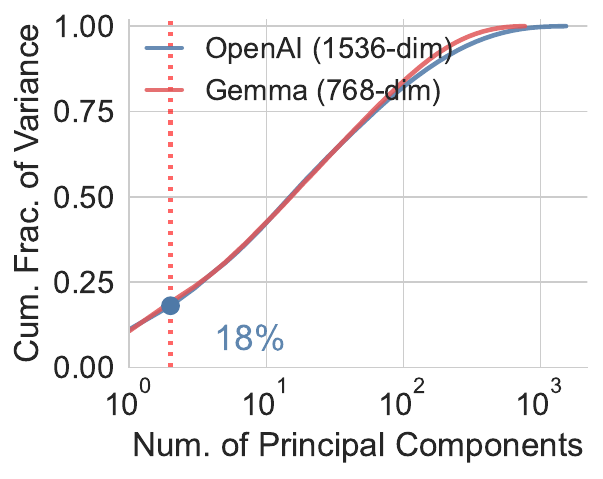}
\caption{\textbf{The 2D View Discards Even More Structure on the Original Paper's Pool Than on Ours.}
PCA of \citet{jiang2025artificial}'s own responses (privately shared with us) to ``Write a metaphor involving time.'' ($n=2{,}150$: $43$ models $\times$ $50$ samples).
\emph{Left two panels:} 2D PCA of the full-dimensional L2-normalized embeddings (OpenAI \texttt{text-embedding-3-small}, left; \texttt{embeddinggemma-300m}, right); PC1+PC2 capture only 18.1\% / 19.1\% of variance, less than on our 15-model data (34.9\% / 33.4\%).
\emph{Right panel:} cumulative variance explained; 90\% of variance requires 186 (OpenAI) / 152 (Gemma) components.
The prompt wording differs from our pool's (``involving'', not ``about''; Appendix~\ref{app:prompt-wording-check}).
The OpenAI-embedding panels use embeddings recomputed with the same embedding model rather than the original run, which can shift per-pair cosine statistics on this prompt by up to ${\sim}0.016$; the Gemma panels are unaffected.}
\label{fig:jiang-43model-pca-variance}
\end{figure}

Section~\ref{sec:artificial_hivemind_figure1} states the headline result of the re-test on the original authors' own 43-model time-prompt responses (Fig.~\ref{fig:jiang-43model-time-reclustering}). This appendix reports the pool and prompt-wording differences from the original, the model-vintage confound the re-test addresses, the PCA views (Fig.~\ref{fig:jiang-43model-pca-variance}), the matched-$n$ subsamples, the coarse-settings analysis, and two caveats. None changes the main-text conclusion.

\paragraph{Pool and Prompt-Wording Differences.}
The original paper made its flagship claim on a 25-model pool. The corpus its authors privately shared with us contains 43 models, all open-weight and disjoint from our 15-model pool. Our pool is mostly newer. It shares GPT-4o and GPT-4o Mini with the original paper's evaluations, but not with the privately shared 43-model corpus (Appendix~\ref{app:model-roster}).
The prompt wording also differs: \citet{jiang2025artificial}'s main text and Figure 1 caption state the prompt as ``Write a metaphor about time,'' but the actual prompt is ``involving time'' (Appendix~\ref{app:prompt-wording-check} examines this difference).

\paragraph{The Model-Vintage Confound.}
Most of our 15 models postdate the original paper's pool (Appendix~\ref{app:model-roster}). This vintage difference is a confound. Training changes in the interval could have altered clustering structure. So ``we find more clusters on newer models'' does not by itself contradict a claim made about older models.
The re-test addresses this. On the original authors' own responses, clustering methods recover more than two clusters at essentially every setting we tested. The two pools do differ geometrically (see the matched-$n$ analysis below).
Our pool also reproduces the original paper's intra-model similarity statistic (Section~\ref{sec:inference-time}). Our baseline within-model mean pairwise cosine is 0.805. This is consistent with the original report that same-prompt average pairwise similarity ``typically exceeds 0.8.''
In the separate EmbeddingGemma-300m space the corresponding value is 0.831. That is a robustness check only and is not on the same scale as the OpenAI-space numbers. On that statistic nothing material changed across vintages. The hypothesis that newer models fixed the measured homogeneity has no support. We do not claim the converse.

\paragraph{Full-Pool Cluster Counts and PCA Views.}
On their pool, every method at the main text's settings points well above two clusters: vMF--BIC selects $K^{*}{=}42$ at 128 dimensions and $K^{*}{=}44$ at 256, HDBSCAN finds $K{=}5$ (10.7\% noise), and the spectral eigengap suggests $K{=}4$ (Fig.~\ref{fig:jiang-43model-time-reclustering}). Two principal components are even less representative than on our data. PC1+PC2 capture 18--19\% of variance, and 90\% requires 186 OpenAI-embedding components or 152 Gemma-embedding components. Fig.~\ref{fig:jiang-43model-pca-variance} reports the exact per-embedding figures and views.

\paragraph{Matched-$n$ Subsamples.}
Cluster counts are sensitive to sample size, so the fair comparison holds $n$ fixed. We drew 15-model subsamples of their pool ($n{=}750$, three seeds, matching our pool's shape; Gemma 128-dimensional slices throughout). On them, vMF--BIC selects $K^{*}{=}16$--$18$ versus our 31, HDBSCAN finds $K{=}2$--$3$ versus our 11, and the spectral eigengap suggests $K{=}1$--$3$ versus our 14.
The matched-$n$ density and spectral counts of 2--3 reflect coarser embedding geometry on their pool, where density-based methods merge neighboring clusters. They are not a collapse to two ideas.
Even there, vMF--BIC never selects $K{=}2$ or anything near it: no near-optimal $K$ falls below 11 in any seed.

\paragraph{Coarse Settings.}
Deliberately coarse settings do produce $K{=}2$ on their full pool. HDBSCAN at \texttt{min\_cluster\_size}~$=50$, the spectral eigengap at $k{=}50$, and HDBSCAN at \texttt{min\_cluster\_size}~$=10$ on the OpenAI embeddings each return exactly two clusters. vMF--BIC selects nothing near $K{=}2$ at any tested setting.
Our own pool also degenerates at such settings (Appendix~\ref{app:hdbscan-sensitivity}). With HDBSCAN \texttt{min\_cluster\_size}~$=50$, our topics fall to a median of 4 clusters (minimum 2). Their matched-$n$ subsamples return no clusters at all (100\% noise).

\paragraph{Caveats.}
This re-test has two caveats.
First, wording. Their pool answers ``Write a metaphor involving time'' while ours answers ``about time.'' The two are not interchangeable (Appendix~\ref{app:prompt-wording-check}).
Second, provenance. The OpenAI-embedding PCA numbers rest on embeddings we recomputed with the same embedding model rather than the original run. Recomputation can shift per-pair cosine statistics on this prompt by up to ${\sim}0.016$. The clustering comparisons use the independent Gemma embeddings and are unaffected.

\section{How the Vehicle Labels of Section~\ref{sec:artificial_hivemind_figure1} Were Induced}\label{app:concept-label-details}

Finding~\ref{finding:vehicle-labels} labels the vehicle of each response, what it compares time to, on two pools: our own $\OursConceptPoolSize$ responses to ``Write a metaphor about time'' ($15$ models $\times$ $50$ samples) and the original authors' privately shared $\JiangConceptPoolSize$ responses to ``Write a metaphor involving time'' ($43$ models $\times$ $50$ samples). The labels come from the response text alone, with no embedding, clustering, or similarity metric. Findings~\ref{finding:visualization}--\ref{finding:clustering} all read the same embeddings, so this line is independent of them. We labeled $15$ of the $\OursConceptPoolSize$ responses in our pool by hand, as worked examples. A language model (Claude Fable 5) labeled every other response in both pools. We call the labels model-induced throughout.

Every reader, relabeler, and audit in the rest of this appendix is a language model instance, not a person. The reader is a Claude Fable 5 instance in an agent harness that reads one batch and writes its labels. The adversarial audit and the merge audits are separate instances prompted to challenge the label set. The blind relabeler is a separate instance that sees only the response text and no label set.

The idea labels of Finding~\ref{finding:null-demanding} are a separate labeling under a different protocol and different prompts. Claude Fable 5 produced both labelings. The idea protocol classifies each cell's 100 responses in one independent call, 87 calls in total. No error carries from one call to the next. The vehicle protocol is a sequential reading with an append-only label set. An early error therefore propagates to every later batch. Neither labeling feeds the other.

Neither task is a similarity rating, the use of language-model judges that Section~\ref{sec:discussion-metric-validity} calls circular. Each asks the model to name what a response compares time to, or which one idea it expresses. A reader can check any label against the response text. Claude Fable 5 is in neither pool. Six of our fifteen pool models share its provider.

\paragraph{Terminology.}
Section~\ref{sec:introduction} and Finding~\ref{finding:vehicle-labels} define tenor, vehicle, and ground \citep{richards1936philosophy}. This appendix adds one term: a \emph{response} is one model output. Only the time responses were labeled, so the tenor is time throughout this appendix. The other eleven topics of Section~\ref{sec:artificial_hivemind_figure1} are other tenors and were not labeled. Here a \emph{vehicle} is what the response compares time to (\texttt{river}). A \emph{ground} is what the response says about time using that vehicle (``no moment can be kept''). One response may carry several vehicles, and several grounds per vehicle.

\paragraph{Procedure.}
The label set is open and append-only, loosely inspired by the Indian Buffet Process. New vehicles and grounds may be created at any point. Nothing is renamed, merged, or removed while reading is under way. The two pools did not start alike. Our pool's label set was seeded with $7$ vehicles and $27$ (vehicle, ground) pairs. We labeled the $15$ worked examples (pool positions $0$--$14$) and transcribed their pairs. That transcription is the first of our $37$ batches, so the model read $36$ of them. The original authors' label set started empty. The reader works through a pool in a fixed order that interleaves models: response 0 of every model, then response 1 of every model, and so on. This keeps any one model's style from dominating the early vocabulary. It read our pool in $37$ sequential batches and the original authors' in $143$. For each response it records every distinct (vehicle, ground) pair the response makes. Each pair is recorded once per response, however many times the response rephrases it. The record is therefore a binary response-by-pair matrix, not a count. Each batch sees the labels created so far plus the boundary rules (conventions) set so far. The reader added $10$ conventions while reading our pool, at positions $5$, $24$, $37$, $54$, $200$, $203$, $355$, $397$, $516$, and $690$. It added $8$ while reading theirs, at positions $9$, $9$, $39$, $47$, $124$, $125$, $591$, and $958$. At the 10\% mark (ours $180$ of $750$, theirs $118$ of $2{,}150$), an adversarial audit reviewed both chains before reading resumed. Two of their pool's conventions, at positions $124$ and $125$, were corrections that the audit mandated. Each convention binds the batches after it, not the batches before it. The consolidation pass below corrects this asymmetry of sequential labeling. Reading our pool created $42$ vehicles, $253$ distinct (vehicle, ground) pairs, and $3{,}109$ pair--response marks; reading theirs created $146$ vehicles, $1{,}041$ pairs, and $11{,}565$ pair--response marks. A response's several pairs under one vehicle count as one vehicle--response mark. Pair--response marks therefore exceed the vehicle--response marks used below.

\paragraph{Consolidation.}
Merging was forbidden during reading, so the label set is over-split by construction. Merging happens afterwards in two audited passes. Every merge is a recorded mapping line with a written reason, reviewed by an independent audit before it is applied, and reversible. The first pass deduplicates vehicles that name the same thing under different names. Our pool had five merges (\texttt{scroll} into \texttt{book}, \texttt{archivist} into \texttt{calligrapher}, \texttt{fuse} into \texttt{candle}, \texttt{conveyor-belt} into \texttt{train}, \texttt{metronome} into \texttt{engine}). They reduce $42$ vehicles to the $\OursNDistinctConcepts$ of Table~\ref{tab:ours-concept-distribution}. Their pool had two (\texttt{whirlwind} into \texttt{breeze}, \texttt{marathon} into \texttt{race}), reducing $146$ to $\JiangNDistinctImages$. Some pairs were considered and kept apart, each with a written reason: \texttt{ocean}/\texttt{tide}, \texttt{glacier}/\texttt{river}, and the maker vehicles (\texttt{sculptor}, \texttt{weaver}, \texttt{architect}, \texttt{gardener}, \texttt{alchemist}, \texttt{calligrapher}). The \texttt{ocean}/\texttt{tide} ruling is within our pool. It is distinct from the across-pool naming in which their \texttt{ocean} is our \texttt{tide}. No ground is merged across different vehicles in this pass, so what time is compared to is preserved exactly. The second pass pools ground wordings judged equivalent. It yields $\OursNClaims$ distinct grounds on our pool and $\JiangNClaims$ on theirs. The vehicle tables do not depend on it; only the grounds-per-response figures below use it.

\paragraph{Reliability Checks.}
Two checks bear on reliability. The first is consistency on repeated inputs. Both pools contain byte-identical duplicate responses. Duplicates land in different batches under different accumulated context. Identical text should still receive identical (vehicle, ground) pairs. Our pool has $18$ duplicate groups and theirs $13$. The duplicates are short (at most $55$ words). In our pool $12$ of $18$ groups carry only \texttt{river}; in theirs $12$ of $13$ do. This check therefore covers the easy cases. Pairwise agreement over duplicate pairs is $98.3\%$ on our pool ($116$ of $118$ pairs) and $96.2\%$ on theirs ($51$ of $53$). Fully agreeing groups number $17$ of $18$ and $11$ of $13$. Each disagreement contests a single ground, never the vehicle. Three groups of one-sentence \texttt{river} responses supply $93$ of our $118$ pairs, and two supply $38$ of their $53$. The second check is a blind relabel. An independent reader that saw only the response text labeled $\BlindRelabelNPerPool$ random responses per pool. After synonym normalization, its primary vehicle was in the pipeline's vehicle set for $\BlindRelabelOursPrimaryInSet$ of $\BlindRelabelNPerPool$ responses on our pool. On theirs it was $\BlindRelabelJiangPrimaryInSet$ of $\BlindRelabelNPerPool$. On raw names the counts are $\BlindRelabelOursPrimaryInSetRaw$ and $\BlindRelabelJiangPrimaryInSetRaw$. They agreed on whether \texttt{river} is present for $\BlindRelabelOursRiverAgree$ of $\BlindRelabelNPerPool$ responses (ours) and $\BlindRelabelJiangRiverAgree$ of $\BlindRelabelNPerPool$ (theirs).

\paragraph{Denominators and the Two Variants.}
$\OursConceptNZeroAssertion$ of the $\OursConceptPoolSize$ responses makes a comparison but asserts nothing the conventions recognize, so it marks nothing. Table~\ref{tab:ours-concept-distribution} keeps it in the denominator and reports it as a separate row. Table~\ref{tab:ours-concept-distribution-with-recovered} repeats the table with its vehicle recovered by hand (it is \texttt{\OursTopConcept}). A response may carry several vehicles, so response counts and marks differ: $\OursConceptPoolSize$ responses produce $\OursConceptMarks$ vehicle--response marks. We compute the effective number of vehicles over the marks, which partition, not over responses, which do not.

\begin{table}[t]
\centering
\caption{\textbf{Vehicle Distribution With the $1$ Zero-Assertion Response's Vehicle Recovered.} As Table~\ref{tab:ours-concept-distribution}, except that the $1$ response that made a comparison while asserting nothing the conventions recognize has its vehicle recovered by hand (it is \texttt{river}). All $750$ responses then carry a vehicle, so the \emph{no vehicle} row of Table~\ref{tab:ours-concept-distribution} is empty and is omitted. The effective number is again computed over the $1{,}230$ vehicle--response marks, not over the response column. Those $1{,}230$ are the $1{,}174$ in the $14$ kept rows plus $56$ in the tail. A response can carry more than one tail vehicle. $11$ of the $45$ tail responses carry two, so the tail holds $56$ vehicle--response marks over $45$ responses. The shape is unchanged: recovering the zero-assertion response moves the top vehicle's share and the effective number of vehicles only slightly.}
\label{tab:ours-concept-distribution-with-recovered}
\begin{tabular}{lrr}
\toprule
Vehicle & $n$ responses & Share of all $750$ \\
\midrule
\texttt{river} & 605 & 80.7\% \\
\texttt{sculptor} & 138 & 18.4\% \\
\texttt{pickpocket} & 123 & 16.4\% \\
\texttt{weaver} & 102 & 13.6\% \\
\texttt{fire} & 45 & 6.0\% \\
\texttt{currency} & 31 & 4.1\% \\
\texttt{train} & 31 & 4.1\% \\
\texttt{glacier} & 22 & 2.9\% \\
\texttt{calligrapher} & 20 & 2.7\% \\
\texttt{tide} & 19 & 2.5\% \\
\texttt{candle} & 17 & 2.3\% \\
\texttt{book} & 7 & 0.9\% \\
\texttt{ocean} & 7 & 0.9\% \\
\texttt{shadow} & 7 & 0.9\% \\
\midrule
\emph{other vehicles} ($n<7$; 23 of them) & 45 & 6.0\% \\
\midrule
\multicolumn{3}{l}{\emph{All vehicles}: $750$ responses, $1{,}230$ vehicle--response marks} \\
\multicolumn{3}{l}{\quad $1{,}230 = 1{,}174 + 56$: the $14$ kept rows plus the vehicle--response marks on the $45$ tail responses} \\
\multicolumn{3}{l}{Effective number of vehicles: $6.23$ (incl.\ remainder), $5.60$ (kept only)} \\
\bottomrule
\end{tabular}
\end{table}

\paragraph{Multi-Label Structure: Many Grounds per Response, One Dominant Vehicle.}
A response is not a single label. Our responses make $\OursMeanClaimsPerResponse$ grounds on average, and \OursShareTwoPlusClaims{} make two or more. The original authors' pool is similar ($\JiangMeanClaimsPerResponse$ and \JiangShareTwoPlusClaims). The dominance of \texttt{river} is one vehicle used for many distinct grounds, not one sentence repeated. $\OursRiverNClaims$ distinct grounds sit under \texttt{river} in our pool (effective number ${\sim}\OursRiverEffNClaims$) and $\JiangRiverNClaims$ in theirs (${\sim}\JiangRiverEffNClaims$). This is also why the tables carry two denominators. \OursTopConceptShareAll{} of our responses carry \texttt{river}, and \texttt{river} is \OursTopImageShareMarks{} of all vehicle--response marks (theirs: \JiangTopImageShareResponses{} and \JiangTopImageShareMarks).

\paragraph{Menu Answering, the Main Caveat.}
Finding~\ref{finding:vehicle-labels} states the menu counts. $\OursNMenuResponses$ responses answer ``Write a metaphor about time'' with a menu of several separate metaphors rather than one. $\OursNMenuCueAmongMenu$ of those announce it with an explicit cue (``here are a few metaphors, depending on the mood,'' a numbered list, and similar), against \OursNMenuCueAmongCommitted{} of the $\OursNCommittedResponses$ single-vehicle responses. The behavior belongs to a few models, not to the topic. It appears in $\OursNModelsWithMenuResponses$ of the $15$ models, reaches \OursTopMenuModelRate{} in one of them, and is absent from all five OpenAI models. Table~\ref{tab:ours-menu-decomposition} decomposes every vehicle's count into responses that committed to it, menu heads, and menu tails. Table~\ref{tab:ours-committed-only-concept-distribution} reports the distribution over committed responses alone. The tail of Table~\ref{tab:ours-concept-distribution} therefore comes largely from a few models that brainstorm: $\OursNConceptsOnlyInMenuResponses$ of the $\OursNDistinctConcepts$ vehicles never occur in a single-metaphor response. Under both readings there is one dominant vehicle and no second one.

\begin{table}[t]
\centering
\caption{\textbf{The Multi-Label Vehicle Tail Is Almost Entirely One Response Behavior, Not Metaphors That Combine Several Vehicles.} Of our $750$ responses, $617$ answered ``Write a metaphor about time'' with exactly one comparison (\emph{committed}). $132$ answered with a menu of several (\emph{menu}), and $1$ made a comparison while asserting nothing. $131$ of the menu responses (99.2\%) carry an explicit cue such as ``here are a few metaphors, depending on the mood,'' against 0 of the $617$ committed responses. Menu answering is concentrated in $5$ of $15$ models and is near-universal in three of them (\texttt{gemini-3.1-pro-preview} 86.0\%, \texttt{gemini-2.5-flash} 82.0\%, \texttt{gemini-3-pro-preview} 82.0\%). The five OpenAI models produce none. ``Menu head'' is the first comparison a menu response offers, ``menu tail'' any later one. $26$ of the $37$ vehicles occur \emph{only} inside menu responses.}
\label{tab:ours-menu-decomposition}
\begin{tabular}{lrrrr}
\toprule
Vehicle & Total & Committed & Menu head & Menu tail \\
\midrule
\texttt{river} & 604 & 525 & 45 & 34 \\
\texttt{sculptor} & 138 & 22 & 42 & 74 \\
\texttt{pickpocket} & 123 & 32 & 24 & 67 \\
\texttt{weaver} & 102 & 24 & 5 & 73 \\
\texttt{fire} & 45 & 0 & 0 & 45 \\
\texttt{currency} & 31 & 1 & 0 & 30 \\
\texttt{train} & 31 & 0 & 0 & 31 \\
\texttt{glacier} & 22 & 2 & 6 & 14 \\
\texttt{calligrapher} & 20 & 2 & 3 & 15 \\
\texttt{tide} & 19 & 2 & 5 & 12 \\
\texttt{candle} & 17 & 2 & 0 & 15 \\
\texttt{book} & 7 & 0 & 0 & 7 \\
\texttt{ocean} & 7 & 3 & 0 & 4 \\
\texttt{shadow} & 7 & 0 & 0 & 7 \\
\midrule
\emph{23 further vehicles} ($n<7$ each) & 56 & 2 & 2 & 52 \\
\midrule
Total (vehicle--response marks) & 1{,}229 & 617 & 132 & 480 \\
\bottomrule
\end{tabular}
\end{table}

\begin{table}[t]
\centering
\caption{\textbf{Restricted to the $617$ Responses That Answered With a Single Metaphor, the Vehicle Distribution Is Far More Concentrated.} Dropping the menu responses of Table~\ref{tab:ours-menu-decomposition} removes the multi-label ambiguity entirely. One vehicle then holds 85.1\% of responses, and the effective number of vehicles falls from the $6.23$ of Table~\ref{tab:ours-concept-distribution} to $1.86$. The two-cluster reading fails under this restriction too, and by a wider margin. There is one dominant vehicle and no second one.}
\label{tab:ours-committed-only-concept-distribution}
\begin{tabular}{lrr}
\toprule
Vehicle & $n$ responses & Share of the $617$ \\
\midrule
\texttt{river} & 525 & 85.1\% \\
\texttt{pickpocket} & 32 & 5.2\% \\
\texttt{weaver} & 24 & 3.9\% \\
\texttt{sculptor} & 22 & 3.6\% \\
\midrule
\emph{other vehicles} ($n<7$; 7 of them) & 14 & 2.3\% \\
\midrule
\multicolumn{3}{l}{Effective number of vehicles: $1.86$ (incl.\ remainder), $1.69$ (kept only)} \\
\bottomrule
\end{tabular}
\end{table}

\paragraph{The Same Reading on the Original Authors' Pool.}
Table~\ref{tab:jiang-image-distribution} reports the vehicle distribution of the original authors' own $\JiangConceptPoolSize$ responses under the same protocol. The re-test paragraph of Section~\ref{sec:artificial_hivemind_figure1} states the result. $\JiangNZeroImage$ of their responses carry no (vehicle, ground) pair; $\JiangNZeroPairWithVehicle$ of those name a vehicle without asserting a ground. $\JiangNKeptImages$ vehicles clear the pool's retention bar. The table's caption states the effect of one brainstorm-list model on the tail. Without it, $\JiangNDistinctImagesNoQwQ$ distinct vehicles remain and the conclusion is unchanged.

\begin{table}[tp]
\centering
\caption{\textbf{The Original Authors' Own $2{,}150$ Time-Metaphor Responses Show One Dominant Vehicle Plus a Heavy Tail.} Distribution of the $2{,}150$ privately shared responses of \citet{jiang2025artificial} ($43$ models $\times$ $50$ samples, ``Write a metaphor involving time'') over the vehicles induced by the same reading protocol as Table~\ref{tab:ours-concept-distribution}. A vehicle is what the response compares time to. The reading is multi-label, so a response counts once per vehicle it carries and the shares do not sum to $1$. Vehicles are kept when ${\geq}20$ responses carry them (0.93\% of the pool, the same relative bar as Table~\ref{tab:ours-concept-distribution}'s $n{\geq}7$); the remaining $115$ are pooled. \texttt{river} is carried by 90.2\% of responses and holds 54.0\% of the $3{,}593$ vehicle--response marks. The effective number of vehicles is $7.11$: $\exp$ of the Shannon entropy over the vehicle--response marks, with the pooled remainder as one category (the rule of Table~\ref{tab:ours-concept-distribution}). It is computed over the $3{,}593$ vehicle--response marks, not over the response column. Those $3{,}593$ are the $3{,}018$ in the $29$ kept rows plus $575$ in the tail. A response that lists several comparisons can carry more than one tail vehicle. $53$ of the $128$ tail responses carry two or more (the maximum is $33$), so the tail holds $575$ vehicle--response marks over $128$ responses. The shape is one dominant vehicle plus a heavy tail, not the two clusters the original paper describes. The second vehicle, \texttt{thief}, is carried by 4.5\% of responses. The weaving family (\texttt{tapestry}), the second cluster the original paper names, is a small minority at 3.3\%. \textbf{Read the tail with this caveat.} One model, \texttt{Qwen\_QwQ-32B-Preview}, answers with brainstorm lists. Its $50$ responses supply 36.8\% of the vehicle--response marks, and $66$ of the $144$ vehicles occur only in them. Without that model the pool has $78$ distinct vehicles and $7$ clear the bar. The effective number is then $2.05$ and the weaving family holds 1.5\%. Restricted to the $1{,}995$ responses that carry exactly one vehicle, \texttt{river} holds 92.0\% and the weaving family $13$ responses. The conclusion, one dominant vehicle and no second dominant one, holds under every reading. Label names differ between pools: their \texttt{tapestry} is our \texttt{weaver} (the weaving family), their \texttt{thief} is our \texttt{pickpocket}, and their \texttt{ocean} is our \texttt{tide}.}
\label{tab:jiang-image-distribution}
\small
\renewcommand{\arraystretch}{0.9}
\begin{tabular}{lrrr}
\toprule
Vehicle & $n$ responses & Share of all $2{,}150$ & Share of $3{,}593$ vehicle--response marks \\
\midrule
\texttt{river} & 1{,}940 & 90.2\% & 54.0\% \\
\texttt{thief} & 97 & 4.5\% & 2.7\% \\
\texttt{tapestry} & 71 & 3.3\% & 2.0\% \\
\texttt{painter} & 62 & 2.9\% & 1.7\% \\
\texttt{ocean} & 57 & 2.7\% & 1.6\% \\
\texttt{journey} & 54 & 2.5\% & 1.5\% \\
\texttt{clock} & 46 & 2.1\% & 1.3\% \\
\texttt{dance-partner} & 46 & 2.1\% & 1.3\% \\
\texttt{mirror} & 44 & 2.0\% & 1.2\% \\
\texttt{garden} & 43 & 2.0\% & 1.2\% \\
\texttt{symphony} & 41 & 1.9\% & 1.1\% \\
\texttt{puzzle} & 38 & 1.8\% & 1.1\% \\
\texttt{currency} & 37 & 1.7\% & 1.0\% \\
\texttt{book} & 35 & 1.6\% & 1.0\% \\
\texttt{library} & 35 & 1.6\% & 1.0\% \\
\texttt{train} & 34 & 1.6\% & 0.9\% \\
\texttt{tree} & 34 & 1.6\% & 0.9\% \\
\texttt{maze} & 33 & 1.5\% & 0.9\% \\
\texttt{theater} & 32 & 1.5\% & 0.9\% \\
\texttt{teacher} & 31 & 1.4\% & 0.9\% \\
\texttt{sculptor} & 27 & 1.3\% & 0.8\% \\
\texttt{dream} & 26 & 1.2\% & 0.7\% \\
\texttt{game} & 24 & 1.1\% & 0.7\% \\
\texttt{spiderweb} & 24 & 1.1\% & 0.7\% \\
\texttt{spiral} & 23 & 1.1\% & 0.6\% \\
\texttt{tape-player} & 22 & 1.0\% & 0.6\% \\
\texttt{breeze} & 21 & 1.0\% & 0.6\% \\
\texttt{mosaic} & 21 & 1.0\% & 0.6\% \\
\texttt{race} & 20 & 0.9\% & 0.6\% \\
\midrule
\emph{other vehicles} ($n<20$; 115 of them) & 128 & 6.0\% & 16.0\% \\
\emph{no (vehicle, ground) pair} & 14 & 0.7\% & --- \\
\midrule
\multicolumn{4}{l}{\emph{All vehicles}: $2{,}150$ responses, $3{,}593$ vehicle--response marks} \\
\multicolumn{4}{l}{Effective number of vehicles: $7.11$ (incl.\ remainder), $6.12$ (kept only)} \\
\bottomrule
\end{tabular}
\end{table}

\paragraph{What This Labeling Does and Does Not Establish.}
This is one labeling at one granularity, and nothing fixes vehicle granularity independently. A coarser reading would merge \texttt{ocean} and \texttt{tide} into \texttt{river}. A finer one would split \texttt{river} further. Either move changes the effective number. The labeling supports a comparative, coarse claim. No partition of these responses at any granularity we examined yields a second vehicle close to the leading one, because a single vehicle holds \OursTopConceptShareAll{} of them. Claims that depend on the exact effective number carry the granularity caveat. The claim of Finding~\ref{finding:vehicle-labels} does not.

\section{Per-Topic Diversity Ranking Under the Baseline Prompt}\label{app:per-topic-diversity-ranking}

Finding~\ref{finding:case-selection} and the summary in Section~\ref{sec:discussion} describe ``time'' as one of the least diverse of our twelve topics: first of twelve within-model and third of twelve pooled. Table~\ref{tab:per-topic-diversity-ranking} reports the full per-topic ranking.
``Time'' ranks first (least diverse) on within-model Rao's Q ($Q_\alpha = 0.144$) and on within-model mean pairwise cosine ($\bar{c} = 0.854$, the highest of the twelve). It ranks third on pooled Rao's Q ($Q_\gamma = 0.301$, behind ``the Internet'' at $0.279$ and ``change'' at $0.284$).
The ranking uses concentration statistics (Rao's Q and mean pairwise cosine). Effective rank and participation ratio, computed on mean-centered embeddings as in Section~\ref{sec:inference-time}, measure how many directions the scatter uses, not how tight it is. On those, ``time'' ranks among the most diverse (within-model $\mathrm{ER}_\alpha$ 11 of 12, $\mathrm{PR}_\alpha$ 12 of 12), because its small residual scatter around the dominant river vehicle points in many directions. We therefore do not use them to rank topics.
The $\bar{c}$ column also underlies Section~\ref{sec:inference-time}'s reproduction of the original paper's intra-model similarity statistic (per-topic range 0.741--0.854; 6 of 12 topics above 0.8).

\begin{table}[tbp]
\centering
\caption{\textbf{``Time'' Is the Least Diverse Topic on the Within-Model Statistics and Third-Least on Pooled Diversity.} Per-topic diversity under the baseline prompt (\texttt{text-embedding-3-small}). $Q_\alpha$/$Q_\gamma$: Rao's quadratic entropy at the within-model/pooled scale; $\bar{c}$: within-model mean pairwise cosine (off-diagonal). Parenthesized ranks run from 1 (least diverse) to 12 (most diverse); rows sorted by the $Q_\gamma$ rank.}
\label{tab:per-topic-diversity-ranking}
\begin{tabular}{lccc}
\toprule
Topic & $Q_\alpha$ (rank) & $Q_\gamma$ (rank) & $\bar{c}$ (rank) \\
\midrule
the Internet & 0.152 (3) & 0.279 (1) & 0.845 (3) \\
change & 0.147 (2) & 0.284 (2) & 0.850 (2) \\
time & 0.144 (1) & 0.301 (3) & 0.854 (1) \\
ambition & 0.166 (5) & 0.309 (4) & 0.830 (5) \\
hope & 0.158 (4) & 0.322 (5) & 0.839 (4) \\
friendship & 0.185 (6) & 0.332 (6) & 0.811 (6) \\
silence & 0.200 (7) & 0.337 (7) & 0.796 (7) \\
travel & 0.224 (9) & 0.351 (8) & 0.772 (9) \\
beauty & 0.228 (11) & 0.372 (9) & 0.768 (11) \\
love & 0.226 (10) & 0.378 (10) & 0.769 (10) \\
nature & 0.207 (8) & 0.400 (11) & 0.789 (8) \\
food & 0.254 (12) & 0.431 (12) & 0.741 (12) \\
\bottomrule
\end{tabular}
\end{table}

The embedding ranking above uses concentration statistics, which measure tightness only (Appendix~\ref{app:alpha-beta-gamma-primer}). As a second instrument, the idea labels of Appendix~\ref{app:conceptual-vs-stylistic} give a content-based ranking. Those labels cover 100 responses per topic under the baseline prompt, one idea per response. Claude Fable 5 produced them under two sampling seeds.

Table~\ref{tab:per-topic-idea-label-ranking} reports that ranking. ``Time'' ranks second of twelve on both modal-idea share and the effective number of ideas. The two instruments therefore agree on time's rank. Caveats: 100 of 750 responses per topic and one label per response.

\begin{table}[tbp]
\centering
\caption{\textbf{Idea Labels Rank ``Time'' Second of Twelve on Both Concentration Measures.} Per-topic idea concentration under the baseline prompt, from the idea-labeling probe of Appendix~\ref{app:conceptual-vs-stylistic}. Protocol: 100 responses sampled per (topic, phrasing) cell, stratified over the 15 models; one idea label per response, assigned by Claude Fable 5; ideas kept at $\geq 10$ members; two sampling seeds. Modal-idea share: fraction of the 100 responses on the most common idea, per seed and averaged over the two seeds. Effective ideas: $\exp$ of the Shannon entropy of the kept-idea shares, averaged over the two seeds. Parenthesized ranks run from 1 (most concentrated, least diverse) to 12. They are computed on the two-seed means. Ties take the minimum rank. The last two columns repeat the embedding-based ranks of Table~\ref{tab:per-topic-diversity-ranking} ($Q_\alpha$: within-model; $Q_\gamma$: pooled). Rows sorted by the modal-idea share rank. Spearman $\rho$ across the 12 topics: modal-idea share rank vs.\ $Q_\alpha$ rank 0.61, vs.\ $Q_\gamma$ rank 0.68; effective-ideas rank vs.\ $Q_\alpha$ rank 0.57, vs.\ $Q_\gamma$ rank 0.69.}
\label{tab:per-topic-idea-label-ranking}
\begin{tabular}{lcccc}
\toprule
Topic & Modal-idea share (rank) & Effective ideas (rank) & $Q_\alpha$ rank & $Q_\gamma$ rank \\
 & seed 1 / seed 2 / mean & mean & & \\
\midrule
change & 0.760 / 0.670 / 0.715 (1) & 1.57 (1) & 2 & 2 \\
time & 0.660 / 0.700 / 0.680 (2) & 1.63 (2) & 1 & 3 \\
the Internet & 0.730 / 0.520 / 0.625 (3) & 2.40 (3) & 3 & 1 \\
travel & 0.380 / 0.460 / 0.420 (4) & 2.95 (5) & 9 & 8 \\
love & 0.380 / 0.440 / 0.410 (5) & 3.31 (6) & 10 & 10 \\
friendship & 0.350 / 0.370 / 0.360 (6) & 2.84 (4) & 6 & 6 \\
ambition & 0.340 / 0.250 / 0.295 (7) & 3.83 (7) & 5 & 4 \\
silence & 0.220 / 0.260 / 0.240 (8) & 3.95 (9) & 7 & 7 \\
food & 0.230 / 0.250 / 0.240 (8) & 4.33 (10) & 12 & 12 \\
hope & 0.240 / 0.240 / 0.240 (8) & 4.42 (11) & 4 & 5 \\
beauty & 0.220 / 0.250 / 0.235 (11) & 3.88 (8) & 11 & 9 \\
nature & 0.210 / 0.250 / 0.230 (12) & 4.90 (12) & 8 & 11 \\
\bottomrule
\end{tabular}
\end{table}

\section{Additional Details for the Null-Distribution Analyses}\label{app:null-analysis-details}

This appendix gives methods and secondary views for Section~\ref{sec:wrong-null}. It states which statistics use sampled pairs and which use all pairs, gives labeling details for Finding~\ref{finding:null-demanding}, reports the per-prompt distribution of the different-idea floor, and summarizes the overlap between same-idea and different-idea pair cosines.

\subsection{Pair Sampling, Enumeration, and Labeling Details}\label{app:null-pair-sampling}

For each analysis we sampled 500 cross-prompt pairs (different-prompt null) and up to 500 same-prompt pairs per prompt (same-prompt conditions). The medians in Table~\ref{tab:null-comparison} come from these sampled pairs.
The pair-level fractions in Finding~\ref{finding:null-demanding} (the $61\%$/$32\%$ and $54\%$/$20\%$ of pairs exceeding 0.7/0.8) use all pairs. So does the per-prompt view in Table~\ref{tab:null-distribution}. We computed them over the full set of persisted embeddings and labels (cluster labels or idea labels).
Finding~\ref{finding:null-undemanding}'s replicated cross-prompt null has median cosine similarity $0.111$ within models and $0.109$ across models.

We recomputed embeddings with the paper's embedding model. Per-pair cosine values on the time prompt can shift by up to ${\sim}0.016$. Aggregate fractions and medians reproduce at reported precision.

Two labeling details for Finding~\ref{finding:null-demanding} follow.
In the geometric-clustering construction, HDBSCAN sometimes returned a degenerate solution (fewer than 3 clusters, more than 10, or over half the responses labeled noise). In those cases the pipeline fell back to spectral clustering with $K{=}5$ on the same embeddings. 16 of the 50 prompts use HDBSCAN clusters and 34 use the spectral fallback.
The 16 HDBSCAN-native prompts have a median across-cluster cosine of $0.737$, slightly \emph{above} the full-set median of $0.723$ (Table~\ref{tab:null-distribution}). The spectral fallback therefore does not inflate the floor; if anything, it lowers it slightly.
A sweep over the fallback $K$ moves the pooled median only from $0.715$ to $0.728$ (Table~\ref{tab:null-distribution}). The low end is $K{=}3$, where one prompt yields no across-cluster pairs and is excluded, leaving 49. The high end is $K{=}8$. The floor is therefore not an artifact of fixing $K{=}5$.
The floor also tends to rise under finer label granularity, because over-splitting relabels same-idea pairs as different-idea pairs. Neither labeling fixes granularity on its own. We therefore report both constructions and compare them (Table~\ref{tab:null-distribution}).
In the LLM-labeling construction, the labeler (Claude Fable 5) is not among the models that generated this pool. Labeler identity is therefore independent of the responses it labels. The labels only partition responses into ideas. The geometric-clustering construction reaches the same different-idea similarity level with no LLM labeler.

\subsection{Per-Prompt Distribution of the Different-Idea Floor}\label{app:per-prompt-floors}

Finding~\ref{finding:null-demanding} reports pair-level fractions. This subsection reports the per-prompt view (Table~\ref{tab:null-distribution}). It agrees, but less strongly. A threshold of 0.7 falls near the \emph{median} of the different-idea distribution. 30 of 50 prompts (60\%) exceed it under geometric clustering. 8 of 15 (53\%) exceed it under LLM labeling. Fewer prompts' floors exceed 0.8 (11/50, 22\%, and 1/15, 7\%).

\begin{table}[tbp]
\centering
\caption{\textbf{Roughly Half of Prompts' Different-Idea Floors Already Exceed the Paper's 0.7 Threshold.} Distribution of the per-prompt different-idea cosine similarity (each prompt contributes one weighted estimate over its different-idea pairs) for the two constructions of Table~\ref{tab:null-comparison}. Fractions are of \emph{prompts} whose floor exceeds the stated threshold; Finding~\ref{finding:null-demanding} reports the corresponding pair-level fractions.}
\label{tab:null-distribution}
\resizebox{\linewidth}{!}{%
\begin{tabular}{lccccc}
\toprule
Null construction & Median & Per-prompt range & IQR (25th--75th) & Prompts ${>}0.7$ & Prompts ${>}0.8$ \\
\midrule
Geometric clustering, 50 prompts & $0.723$ & $0.26$--$0.90$ & $0.61$--$0.80$ & $60\%$ (30/50) & $22\%$ (11/50) \\
\quad fallback $K{=}3$, 49 prompts & $0.715$ & $0.25$--$0.89$ & $0.61$--$0.79$ & $55\%$ (27/49) & $20\%$ (10/49) \\
\quad fallback $K{=}8$, 50 prompts & $0.728$ & $0.27$--$0.90$ & $0.61$--$0.80$ & $64\%$ (32/50) & $26\%$ (13/50) \\
\quad HDBSCAN-native subset, 16 prompts & $0.737$ & $0.39$--$0.87$ & $0.60$--$0.79$ & $62.5\%$ (10/16) & $25\%$ (4/16) \\
LLM-labeled, 15 prompts & $0.703$ & $0.33$--$0.81$ & $0.64$--$0.75$ & $53\%$ (8/15) & $7\%$ (1/15) \\
\bottomrule
\end{tabular}}
\end{table}

\subsection{Overlap Between Same-Idea and Different-Idea Pair Cosines}\label{app:pair-cosine-overlap}

Finding~\ref{finding:null-demanding} notes that medians understate the overlap between the same-idea and different-idea pair-cosine distributions. We summarize that overlap by the AUC for predicting, from cosine alone, whether a pair shares an idea.
Under the LLM idea labels the median AUC is 0.727 across 15 prompts. Cosine is thus a modest predictor of whether two responses share an idea.
The geometric cluster labels give 0.777. That construction is partially circular, because its clusters come from the same geometry being scored. We therefore weight the text-derived idea labels more.

\subsection{Derivation of the \texorpdfstring{${\sim}21.9$}{~21.9} Maximal-Mixing Anchor}

Finding~\ref{finding:indistinguishability} (Section~\ref{sec:wrong-null}) reports that a uniform draw of 50 responses from the 1,250-response pool would contain ${\sim}21.9$ unique models on average if model identity carried no information (maximal mixing). This subsection derives that anchor.
Each of the $M{=}25$ models is absent from a uniform draw of $N{=}50$ out of $Mn{=}1{,}250$ with probability $\binom{1200}{50}/\binom{1250}{50} \approx 0.125$, so the expected count is $25 \times 0.875 \approx 21.9$.
A top-$N$ similarity set is not a uniform draw, so this is a heuristic anchor, not a proper null.

\section{Two Readings of the Original Paper's Training-Level Conclusion}\label{app:training-level-two-readings}

Finding~\ref{finding:training-level} (Section~\ref{sec:inference-time}) states that the original paper's training-level conclusion is unsupported. That conclusion admits two readings. This appendix works through both.
The sentence ``more generalizable solutions are needed at the model training level to robustly preserve output diversity without requiring user intervention'' admits a strong reading: inference-time interventions cannot reduce mode collapse. It also admits an accurate, weaker reading about generalizability and burden: solutions should generalize and should not require user intervention.
The conclusion is unsupported under either reading.
The evidence is that one sampler with no demonstrated diversity benefit failed to reduce homogeneity. This cannot support the strong reading. One sampler's failure says nothing about other inference-time interventions, such as other samplers and prompts. Nor does it support the accurate reading. It gives no evidence about what generalizes or where the burden should fall.
Under either reading the inference to training-level solutions also lacks positive evidence, since the paper never demonstrates that training-level changes reduce homogeneity.

\section{A Primer on \texorpdfstring{$\alpha$, $\beta$, $\gamma$}{alpha, beta, gamma} Diversity for Machine Learning Readers}\label{app:alpha-beta-gamma-primer}

The $\alpha$/$\beta$/$\gamma$ decomposition comes from ecology \citep{whittaker1972,ricottaSzeidl2009}. It splits a population's total diversity into diversity within each sub-population and diversity across sub-populations. This appendix introduces the decomposition, defines each scale for our three instruments (Rao's quadratic entropy, effective rank, participation ratio), and states the implications for Section~\ref{sec:inference-time}.

\subsection{Why Three Scales}

A single diversity number for a collection conflates two sources of diversity. Consider two forests with the same number of trees and the same 50 species:
\begin{itemize}
    \item \textbf{Forest A:} every $10\times 10$ plot contains all 50 species, distributed roughly uniformly.
    \item \textbf{Forest B:} every $10\times 10$ plot contains just one species, but different plots contain different species.
\end{itemize}
Any metric computed on the pooled forest (``$\gamma$-diversity'') gives the same answer for both. But the forests differ. A is varied inside every plot. B is uniform inside every plot but varies across plots. Telling them apart requires a metric for within-plot diversity (``$\alpha$-diversity'') and one for how much the plots differ from each other (``$\beta$-diversity''). Forest~A has high $\alpha$ and low $\beta$; Forest~B has low $\alpha$ and high $\beta$.

The same distinction applies to language-model responses. ``Do models produce homogeneous outputs?'' is two questions:
\begin{itemize}
    \item Does each model produce narrow, repetitive responses? ($\alpha$)
    \item Do different models occupy different regions of response space, or do they all cluster in the same region? ($\beta$)
\end{itemize}
A single pooled metric ($\gamma$) cannot separate these. The Artificial Hivemind thesis implies that $\alpha$ is low (each model is repetitive) and $\beta$ is low (different models cluster together). Reporting all three scales lets us test the two claims separately.

\subsection{Setup and Notation}

Fix a (topic, prompt phrasing) cell. We have $M = 15$ models, each contributing $N = 50$ response embeddings $x_{m,j} \in \mathbb{R}^d$, all L2-normalized so $\|x_{m,j}\| = 1$. Let
\begin{align*}
    X_m &= \{x_{m,1}, \dots, x_{m,N}\} & \text{(one model's responses)}, \\
    X_{\mathrm{pool}} &= \textstyle\bigcup_{m=1}^{M} X_m & \text{(all $MN = 750$ responses pooled)}, \\
    \mu_m &= \tfrac{1}{N} \textstyle\sum_{j=1}^{N} x_{m,j} & \text{(model $m$'s centroid)}, \\
    \bar{\mu} &= \tfrac{1}{M} \textstyle\sum_{m=1}^{M} \mu_m & \text{(mean centroid)}.
\end{align*}
Each diversity instrument $D(\cdot)$ takes a set of vectors and returns a non-negative scalar. The $\alpha$/$\beta$/$\gamma$ scales are defined the same way for every $D$:
\begin{align*}
    D_\alpha &= \tfrac{1}{M} \textstyle\sum_{m=1}^{M} D(X_m) && \text{(mean within-model diversity)}, \\
    D_\beta  &= D(\{\mu_1, \dots, \mu_M\}) && \text{(diversity among the $M$ centroids)}, \\
    D_\gamma &= D(X_{\mathrm{pool}}) && \text{(diversity of the pooled 750-response cloud)}.
\end{align*}
In words: $\alpha$ is how varied one model is, on average; $\beta$ is how much the models' typical responses differ from each other; $\gamma$ is how varied the union of all responses is.

\subsection{Rao's Quadratic Entropy}

Rao's Q \citep{rao1982} is the mean pairwise dissimilarity of a set $S = \{y_1, \dots, y_K\}$,
\[
Q(S) \;=\; \frac{1}{K^2} \sum_{i=1}^{K} \sum_{j=1}^{K} d(y_i, y_j),
\]
for a symmetric dissimilarity $d(\cdot,\cdot)$. We use cosine dissimilarity, $d(x,y) = 1 - x^\top y = \tfrac{1}{2}\|x - y\|^2$ on L2-normalized vectors, which yields two useful rewrites:
\[
Q(S) \;=\; 1 - \|\bar{y}\|^2 \;=\; \tfrac{1}{K} \sum_{i=1}^{K} \|y_i - \bar{y}\|^2 \;=\; \mathrm{tr}(\Sigma_S),
\]
where $\bar{y} = \tfrac{1}{K}\sum_i y_i$ and $\Sigma_S$ is the (biased) sample covariance of $S$. That is, Rao's Q under cosine dissimilarity is the total variance, and on L2-normalized data it equals $1 - \|\text{centroid}\|^2$.
Section~\ref{sec:inference-time} uses one conversion. On one model's $N{=}50$ responses, removing the $N{=}50$ self-pairs turns Rao's Q into the off-diagonal mean pairwise cosine $\bar{c} = 1 - \tfrac{50}{49} Q$.

Applying the generic $\alpha/\beta/\gamma$ templates:
\begin{align*}
    Q_\alpha &= \tfrac{1}{M}\textstyle\sum_m \mathrm{tr}(\Sigma_m), \\
    Q_\beta  &= \tfrac{1}{M}\textstyle\sum_m \|\mu_m - \bar{\mu}\|^2 \;=\; \mathrm{tr}(\Sigma_{\mathrm{between}}), \\
    Q_\gamma &= \tfrac{1}{MN}\textstyle\sum_{m,j} \|x_{m,j} - \bar{\mu}\|^2 \;=\; \mathrm{tr}(\Sigma_{\mathrm{pool}}),
\end{align*}
where $\Sigma_m$, $\Sigma_{\mathrm{between}}$, and $\Sigma_{\mathrm{pool}}$ are the within-model, between-centroid, and pooled covariances.

\paragraph{Rao's Q decomposes additively.} The law of total variance applied to the covariance itself gives $\Sigma_{\mathrm{pool}} = \tfrac{1}{M}\sum_m \Sigma_m + \Sigma_{\mathrm{between}}$. Taking the trace,
\[
\boxed{\;Q_\gamma \;=\; Q_\alpha + Q_\beta\;}
\]
\emph{exactly}, whenever each model contributes the same number of responses \emph{and} $Q_\beta$ is computed on the raw centroids as $\mathrm{tr}(\Sigma_{\mathrm{between}})$ above (the ``variance convention''). If all models collapse to a single (different) point each, $Q_\alpha = 0$ and $Q_\gamma = Q_\beta$. If all models share the same distribution, $Q_\beta = 0$ and $Q_\gamma = Q_\alpha$.

\paragraph{A second, angular $Q_\beta$: renormalized centroids.} Section~\ref{sec:inference-time}'s figures do not report the variance-convention $Q_\beta = \mathrm{tr}(\Sigma_{\mathrm{between}})$ directly. Instead, each raw centroid $\mu_m$ is projected back onto the unit sphere, $c_m = \mu_m / \|\mu_m\|$, and $Q_\beta$ is computed as Rao's Q of the renormalized centroid set $\{c_1, \dots, c_M\}$. Writing $\bar{c} = \tfrac{1}{M}\sum_m c_m$ for the mean of these unit-norm centroids (the same symbol as the mean pairwise cosine in the conversion above, but a different quantity), and using $Q(S) = 1-\|\bar y\|^2$ on any L2-normalized input set,
\[
Q_\beta \;=\; 1 - \|\bar{c}\|^2,
\]
an angular summary of the \emph{renormalized} centroid cloud. It depends only on the norm of the mean direction $\bar c$, not on how the individual $c_m$ are arranged around it. It does not depend on the dimensionality the centroids span. Effective rank and participation ratio, by contrast, read the shape of $\Sigma_{\mathrm{between}}$'s spectrum rather than its trace. Appendix~\ref{app:beta-rao-flatness} uses this point to explain why the three instruments disagree at $\beta$.

Renormalization discards each centroid's original norm. So $\bar c$, the mean of unit vectors, differs from $\bar\mu$, the mean of the raw, sub-unit-norm centroids used in the variance convention above. In particular, $1-\|\bar\mu\|^2$ equals $Q_\gamma$, not $Q_\beta$, so the raw-centroid mean norm is not a $\beta$ quantity. The boxed additivity identity therefore holds \emph{exactly} for the variance-convention $Q_\beta = \mathrm{tr}(\Sigma_{\mathrm{between}})$, but only \emph{approximately} for the renormalized-centroid $Q_\beta$ that the figures plot. At baseline (Section~\ref{sec:inference-time}), $Q_\alpha \approx 0.191$ and the plotted $Q_\beta \approx 0.184$ sum to $\approx 0.375$, versus $Q_\gamma \approx 0.341$. The gap of $\approx 0.033$ is the norm information that renormalization discards. The two $Q_\beta$'s answer different questions. The variance convention asks how much of the pooled variance is between-model. The renormalized-centroid convention asks how tightly the centroids' \emph{directions} agree, independent of how concentrated each model's responses are.

\subsection{Effective Rank}

Effective rank \citep{royvetterli2007} measures how many dimensions a cloud uses, via the entropy of the normalized eigenvalue spectrum. For a set $S$ with sample covariance $\Sigma$ of eigenvalues $\lambda_1 \geq \dots \geq \lambda_d \geq 0$, let $p_i = \lambda_i / \textstyle\sum_k \lambda_k$ (treat $0 \log 0 \equiv 0$). Then
\[
\mathrm{ER}(S) \;=\; \exp\!\Bigl(-\textstyle\sum_{i=1}^{d} p_i \log p_i\Bigr).
\]
$\mathrm{ER} = 1$ when one eigenvalue carries all the variance (a rank-$1$ cloud), $\mathrm{ER} = d$ when all eigenvalues are equal (a fully isotropic cloud), and it interpolates continuously in between as an ``effective number of directions''.

The $\alpha/\beta/\gamma$ scales:
\begin{align*}
    \mathrm{ER}_\alpha &= \tfrac{1}{M}\textstyle\sum_m \mathrm{ER}(\Sigma_m), \quad
    \mathrm{ER}_\beta = \mathrm{ER}(\Sigma_{\mathrm{between}}), \quad
    \mathrm{ER}_\gamma = \mathrm{ER}(\Sigma_{\mathrm{pool}}).
\end{align*}

\paragraph{Effective rank does not decompose additively.} The covariance matrices do decompose: $\Sigma_{\mathrm{pool}} = \bar{\Sigma}_{\mathrm{within}} + \Sigma_{\mathrm{between}}$ where $\bar{\Sigma}_{\mathrm{within}} = \tfrac{1}{M}\sum_m \Sigma_m$. But $\mathrm{ER}$ is a nonlinear functional of the eigenvalue spectrum, so in general
\[
\mathrm{ER}_\gamma \;\neq\; \mathrm{ER}_\alpha + \mathrm{ER}_\beta.
\]
When within-model and between-centroid variation lie in \emph{different} subspaces, $\Sigma_{\mathrm{pool}}$ has support on more directions than either piece alone. Then $\mathrm{ER}_\gamma$ can substantially exceed $\mathrm{ER}_\alpha + \mathrm{ER}_\beta$. When they lie in overlapping subspaces, pooling redistributes variance across existing directions and the inequality can reverse. In Section~\ref{sec:inference-time}, $\mathrm{ER}_\gamma \gg \mathrm{ER}_\alpha$ and $\mathrm{ER}_\gamma \gg \mathrm{ER}_\beta$. This is consistent with pooling adding directions of variation that no single model, and no centroid set, contains alone.

\subsection{Participation Ratio}

Participation ratio \citep{belldean1970,gao2017} is a second effective-dimensionality measure, with the same goal as effective rank but a different functional form. For covariance $\Sigma$ with eigenvalues $\lambda_1, \dots, \lambda_d \geq 0$,
\[
\mathrm{PR}(\Sigma) \;=\; \frac{\bigl(\textstyle\sum_i \lambda_i\bigr)^2}{\textstyle\sum_i \lambda_i^2} \;=\; \frac{\mathrm{tr}(\Sigma)^2}{\mathrm{tr}(\Sigma^2)}.
\]
Like $\mathrm{ER}$, $\mathrm{PR} = 1$ for a rank-$1$ covariance and $\mathrm{PR} = d$ for an isotropic one, with continuous interpolation in between. The $\alpha/\beta/\gamma$ instantiations mirror effective rank's:
\begin{align*}
    \mathrm{PR}_\alpha &= \tfrac{1}{M}\textstyle\sum_m \mathrm{PR}(\Sigma_m), \quad
    \mathrm{PR}_\beta = \mathrm{PR}(\Sigma_{\mathrm{between}}), \quad
    \mathrm{PR}_\gamma = \mathrm{PR}(\Sigma_{\mathrm{pool}}).
\end{align*}
Like $\mathrm{ER}$, $\mathrm{PR}$ is nonlinear in the eigenspectrum, so $\mathrm{PR}_\gamma \neq \mathrm{PR}_\alpha + \mathrm{PR}_\beta$ in general. The two metrics often agree qualitatively, since both increase as the covariance spectrum spreads more evenly. They can disagree quantitatively when the eigenvalue tail is heavy, because $\mathrm{ER}$'s entropy is more tail-sensitive than $\mathrm{PR}$'s second-moment ratio.

We compute effective rank and participation ratio on mean-centered embeddings. They therefore measure the shape of the spread, not the shared mean direction. The $\alpha$- and $\gamma$-scale gains in Figs.~\ref{fig:prompt-phrasing-er} and \ref{fig:prompt-phrasing-pr} also hold without centering. Every phrasing still beats \texttt{baseline}, with smaller absolute values, because the mean direction dominates the uncentered spectrum.

\subsection{Four Regimes of Response Populations}

The three-scale decomposition distinguishes four regimes that a single pooled number cannot:

\begin{itemize}
    \item \textbf{Low $\alpha$, low $\beta$ (``hivemind''):} each model is internally repetitive and different models cluster in the same region. Pooled $\gamma$ is also small. This is the Artificial Hivemind thesis.
    \item \textbf{High $\alpha$, low $\beta$ (``shared, individually varied''):} each model is internally diverse, but all models cover roughly the same region. Pooled diversity is only marginally larger than $\alpha$.
    \item \textbf{Low $\alpha$, high $\beta$ (``specialized, collectively varied''):} each model is individually repetitive but different models specialize in different regions. Pooled diversity is dominated by between-model spread.
    \item \textbf{High $\alpha$, high $\beta$ (``diverse and differentiated''):} each model is internally diverse \emph{and} different models go to different regions. Pooled $\gamma$ is far larger than either $\alpha$ or $\beta$ alone.
\end{itemize}
Finding~\ref{finding:rewordings} shows that reworded prompts raise $\alpha$ and $\gamma$ under every metric as a mean across the twelve topics (Figs.~\ref{fig:prompt-phrasing-er}, \ref{fig:prompt-phrasing-rao}, and \ref{fig:prompt-phrasing-pr}), and $\beta$ under the eigenvalue-based metrics. This moves the population toward ``diverse and differentiated'' and away from the ``hivemind'' regime.

\subsection{Implications for This Paper}

Three implications follow from the decomposition:

\paragraph{Each metric--scale pair answers a different question; they are not substitutes.} $Q_\gamma$ sums $Q_\alpha$ and $Q_\beta$ exactly under the variance convention, but $\mathrm{ER}_\gamma$ and $\mathrm{PR}_\gamma$ do not. The pooled eigenvalue-based scores therefore carry information that their $\alpha$ and $\beta$ scores do not. Conversely, a $\gamma$-only report would miss the $\alpha$-vs-$\beta$ distinction entirely.

\paragraph{The three instruments at $\beta$ measure related but non-identical geometry.} As plotted in Section~\ref{sec:inference-time}'s figures, Rao's Q at $\beta$ depends only on $\|\bar{c}\|$, the norm of the mean of the L2-\emph{renormalized} centroids $c_m = \mu_m/\|\mu_m\|$, not on the raw centroid mean $\bar\mu$ of the variance convention above. Effective rank and participation ratio at $\beta$ instead read the \emph{shape} of the raw centroid covariance spectrum, $\Sigma_{\mathrm{between}}$. Both are valid summaries of how much the models differ, but they can move differently. Appendix~\ref{app:beta-rao-flatness} shows this: prompt phrasings leave $\|\bar{c}\|$ unchanged ($Q_\beta$ is flat) but increase the dimensionality the centroids span ($\mathrm{ER}_\beta$ and $\mathrm{PR}_\beta$ rise).

\paragraph{``Total diversity'' alone is not diagnostic.} The Artificial Hivemind framing is equally consistent with (i) every model being internally repetitive, or (ii) every model being diverse but all covering the same region. The two call for different remedies (e.g., ``diversify each model'' vs.\ ``combine more models''). Only reporting $\alpha$ and $\beta$ separately, rather than only $\gamma$, can distinguish them.

\section{Additional Analyses of the Prompt-Phrasing Result}\label{app:prompt-phrasing-additional}

This appendix collects the secondary views and robustness analyses behind Finding~\ref{finding:rewordings}. It reports the significance statistics and metric agreement. It reports the Rao's-Q and participation-ratio versions of the main text's effective-rank figure. It explains why Rao's Q is flat at the $\beta$ scale. It reports the response-length confound and the LLM idea-labeling probe. None changes Finding~\ref{finding:rewordings}'s conclusion.

\subsection{Significance Tests and Metric Agreement for Finding~\ref{finding:rewordings}}\label{app:prompt-phrasing-significance}

The 108 cells reuse the same 12 topics across scales and metrics, so we test significance at the topic level. For each (phrasing, scale, metric) combination we run a one-sided sign test and a one-sided Wilcoxon signed-rank test, paired over the 12 topics. The smallest attainable sign-test $p$-value is $2.4 \times 10^{-4}$. At the $\alpha$ and $\gamma$ scales, 25 of the 30 sign tests reach that floor, meaning the phrasing beats baseline in all 12 topics (13 of 15 tests at $\alpha$, 12 of 15 at $\gamma$). At $\beta$ the effect is weaker. Under Rao's Q it is absent: no phrasing beats baseline there, and \texttt{poet\_persona} wins only 1 of 12 topics. Appendix~\ref{app:beta-rao-flatness} explains this $\beta$-scale degeneracy of Rao's Q. The baseline prompt ranks worst of the six phrasings in 72\% of cells and best in only 5 (all at $\beta$ under Rao's Q). The three metrics rank the six phrasings almost identically at $\alpha$ and $\gamma$ (phrasing-mean Spearman $\rho \geq +0.83$ for all three pairs at $\alpha$; $\rho \geq +0.94$ at $\gamma$). The ordering therefore does not depend on the choice of metric.
In 95\% of the 108 cells, at least one of the five phrasings outranks baseline. The main text reports per-phrasing win rates instead, because this at-least-one statistic is easy to misread as a per-phrasing average.

\subsection{Rao's Q and Participation Ratio Across Prompt Phrasings}\label{app:prompt-phrasing-secondary-metrics}

\begin{figure}[t!]
\centering
\includegraphics[width=0.8\linewidth]{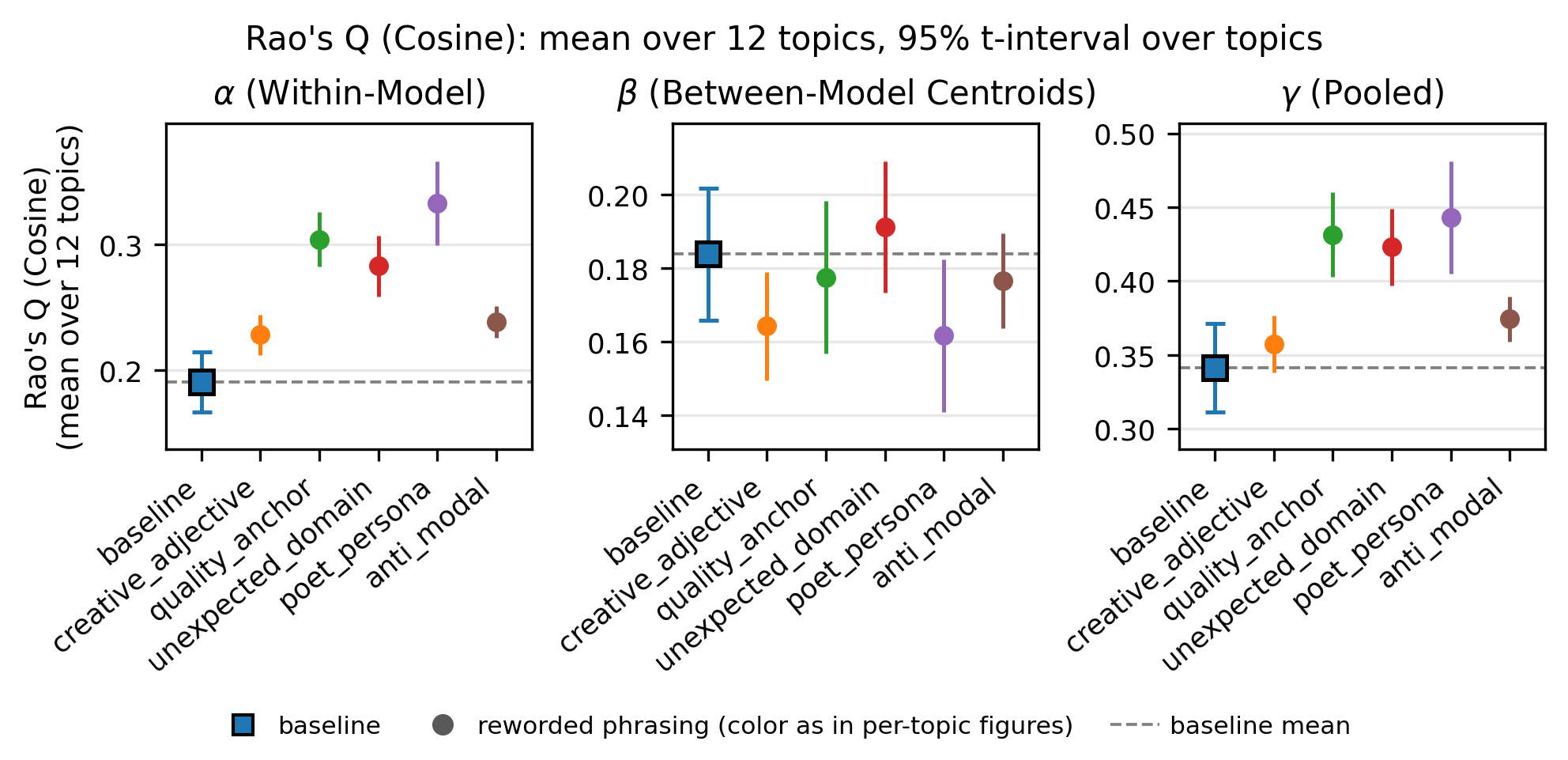}
\caption{\textbf{Rao's Q Rises Under Every Prompt Phrasing at the Within-Model and Pooled Scales.} Rao's quadratic entropy \citep{rao1982} at the $\alpha$, $\beta$, and $\gamma$ scales. At $\beta$ the model centroids are renormalized to the unit sphere before computing $Q$ (Appendix~\ref{app:alpha-beta-gamma-primer}). Each value is the mean over the 12 topics for one of six phrasings. The square marks baseline and the dashed line marks its mean. Error bars are 95\% $t$-intervals over the 12 topics.
All five non-baseline phrasings raise $\alpha$- and $\gamma$-scale Rao's Q over \texttt{baseline}. At $\gamma$, \texttt{poet\_persona} is the strongest phrasing ($0.341 \to 0.443$, $+29.8\%$) and \texttt{anti\_modal} the weakest of the five ($+9.7\%$). At $\beta$, Rao's Q is essentially flat: only \texttt{unexpected\_domain} exceeds baseline ($+4.0\%$). Per-topic values with per-topic jackknife intervals are in Fig.~\ref{fig:prompt-phrasing-rao-per-topic}.
}
\label{fig:prompt-phrasing-rao}
\end{figure}

\begin{figure}[t!]
\centering
\includegraphics[width=0.8\linewidth]{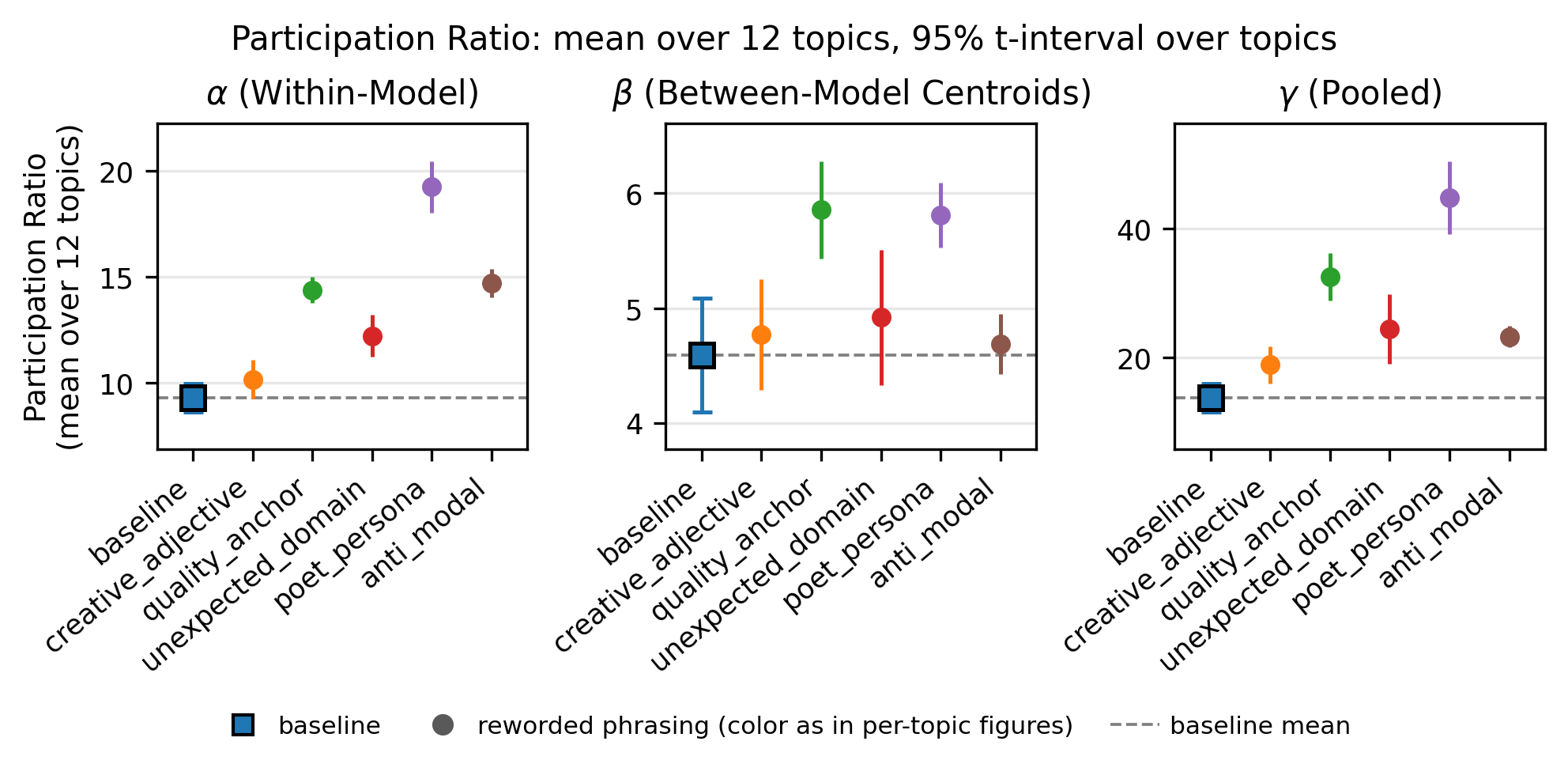}
\caption{\textbf{Participation Ratio Rises Under Every Prompt Phrasing at Every Scale.} Participation ratio \citep{gao2017,belldean1970}, $(\mathrm{tr}\,\Sigma)^2 / \mathrm{tr}(\Sigma^2)$ on the response-covariance eigenvalues, at the $\alpha$, $\beta$, and $\gamma$ scales. Each value is the mean over the 12 topics for one of six phrasings. The square marks baseline and the dashed line marks its mean. Error bars are 95\% $t$-intervals over the 12 topics.
All five non-baseline phrasings raise participation ratio over \texttt{baseline} at every scale. \texttt{Poet\_persona} is the strongest at $\alpha$ ($9.3 \to 19.2$, $+107.1\%$) and at $\gamma$ ($13.8 \to 44.8$, $+224.0\%$). At $\beta$, \texttt{quality\_anchor} leads at $+27.5\%$ ($4.59 \to 5.86$), with \texttt{poet\_persona} close behind ($+26.4\%$). Per-topic values with per-topic jackknife intervals are in Fig.~\ref{fig:prompt-phrasing-pr-per-topic}.
}
\label{fig:prompt-phrasing-pr}
\end{figure}

\begin{figure}[t!]
\centering
\includegraphics[width=\linewidth]{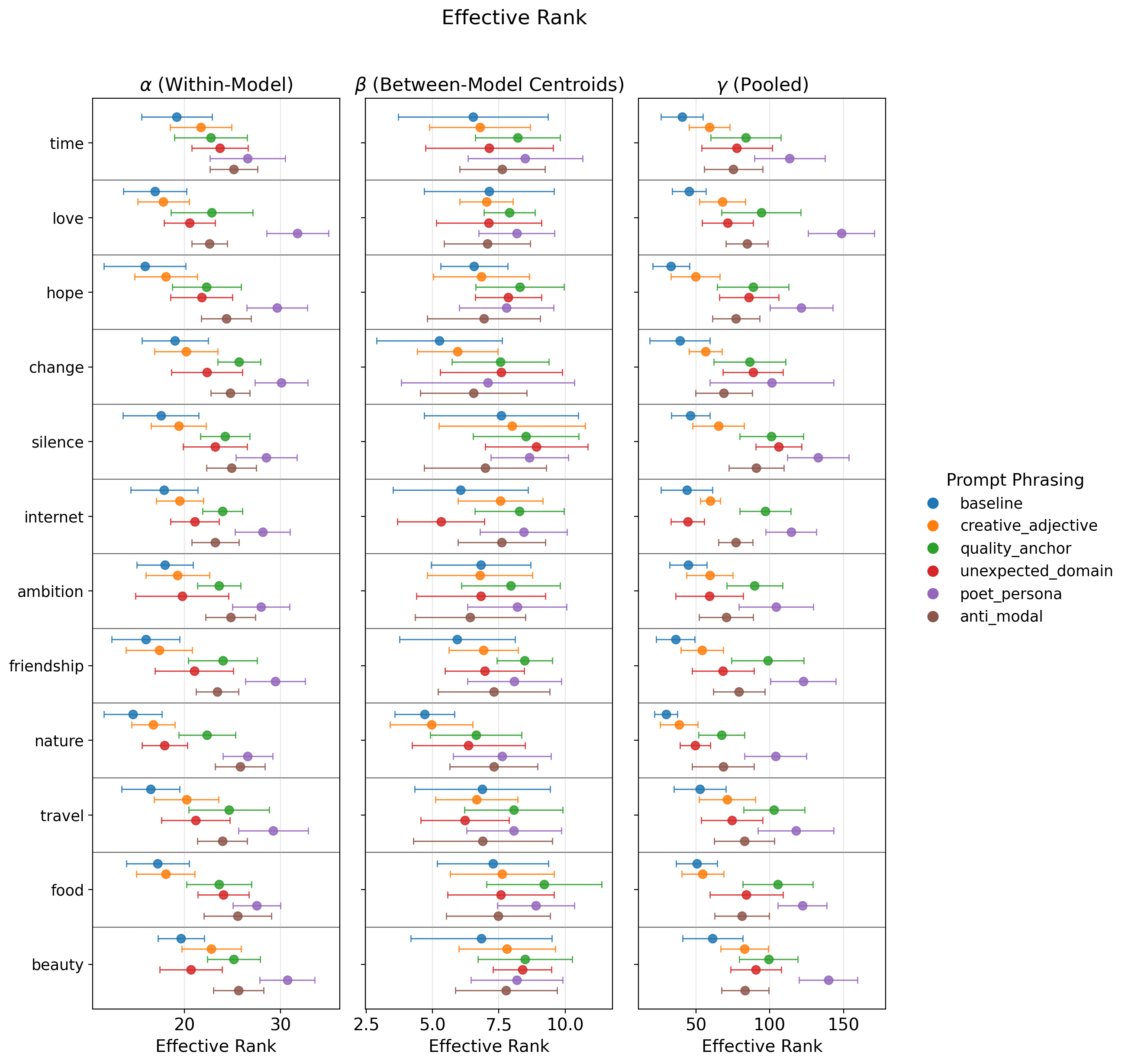}
\caption{\textbf{Effective Rank per Topic Across the Six Prompt Phrasings.} Effective rank per topic (12 rows) at the $\alpha$, $\beta$, and $\gamma$ scales for the six phrasings. Error bars are 95\% leave-one-model-out jackknife intervals over the fifteen models. The row label \texttt{internet} abbreviates ``the Internet'' here and in the per-topic figures that follow. The main text's Fig.~\ref{fig:prompt-phrasing-er} shows the mean over topics.}
\label{fig:prompt-phrasing-er-per-topic}
\end{figure}

\begin{figure}[t!]
\centering
\includegraphics[width=\linewidth]{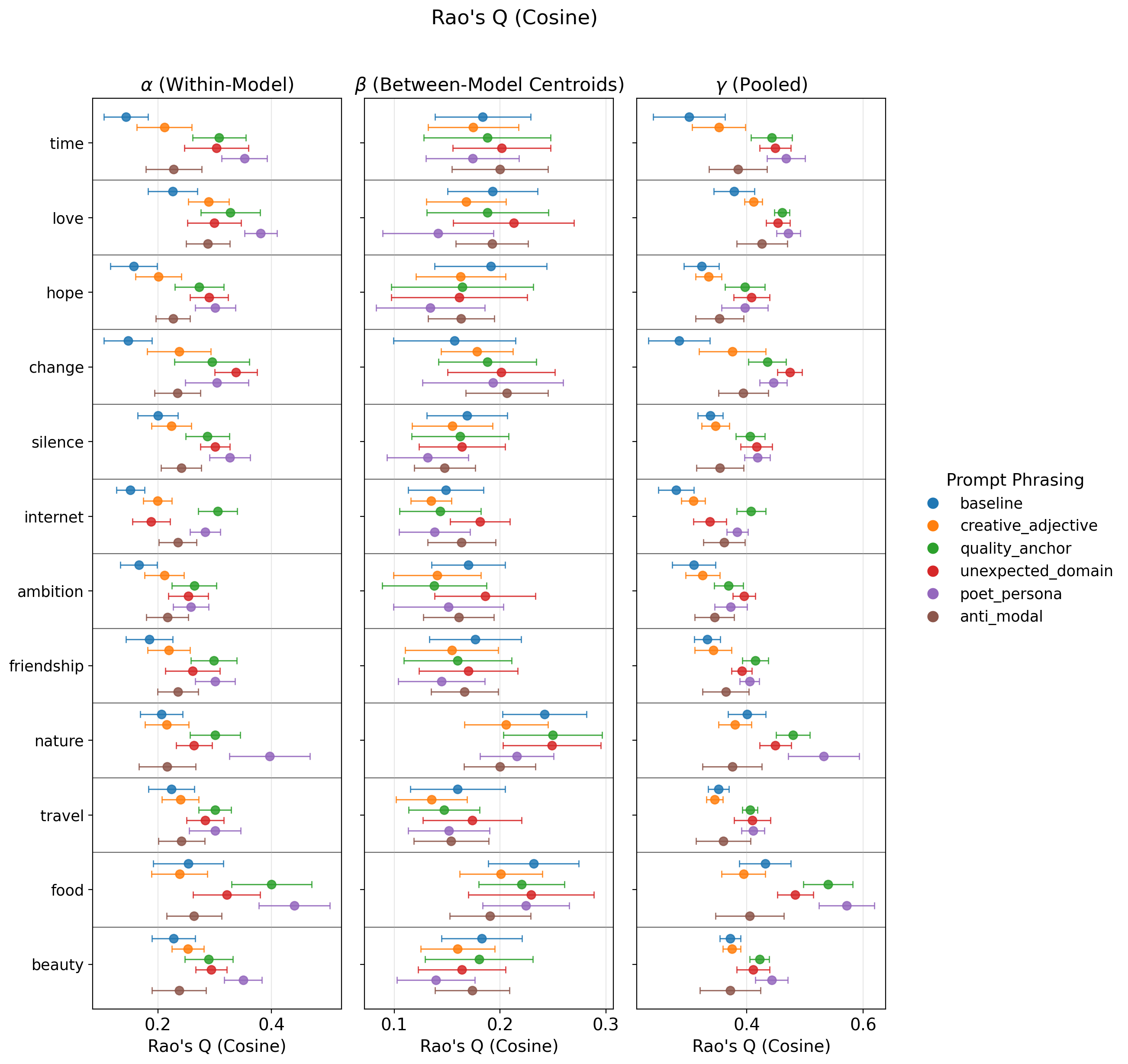}
\caption{\textbf{Rao's Q per Topic Across the Six Prompt Phrasings.} Rao's quadratic entropy per topic (12 rows) at the $\alpha$, $\beta$, and $\gamma$ scales for the six phrasings. Each (phrasing, topic, model) cell has 50 responses. Error bars are 95\% leave-one-model-out jackknife intervals over the fifteen models. Fig.~\ref{fig:prompt-phrasing-rao} shows the mean over topics.}
\label{fig:prompt-phrasing-rao-per-topic}
\end{figure}

\begin{figure}[t!]
\centering
\includegraphics[width=\linewidth]{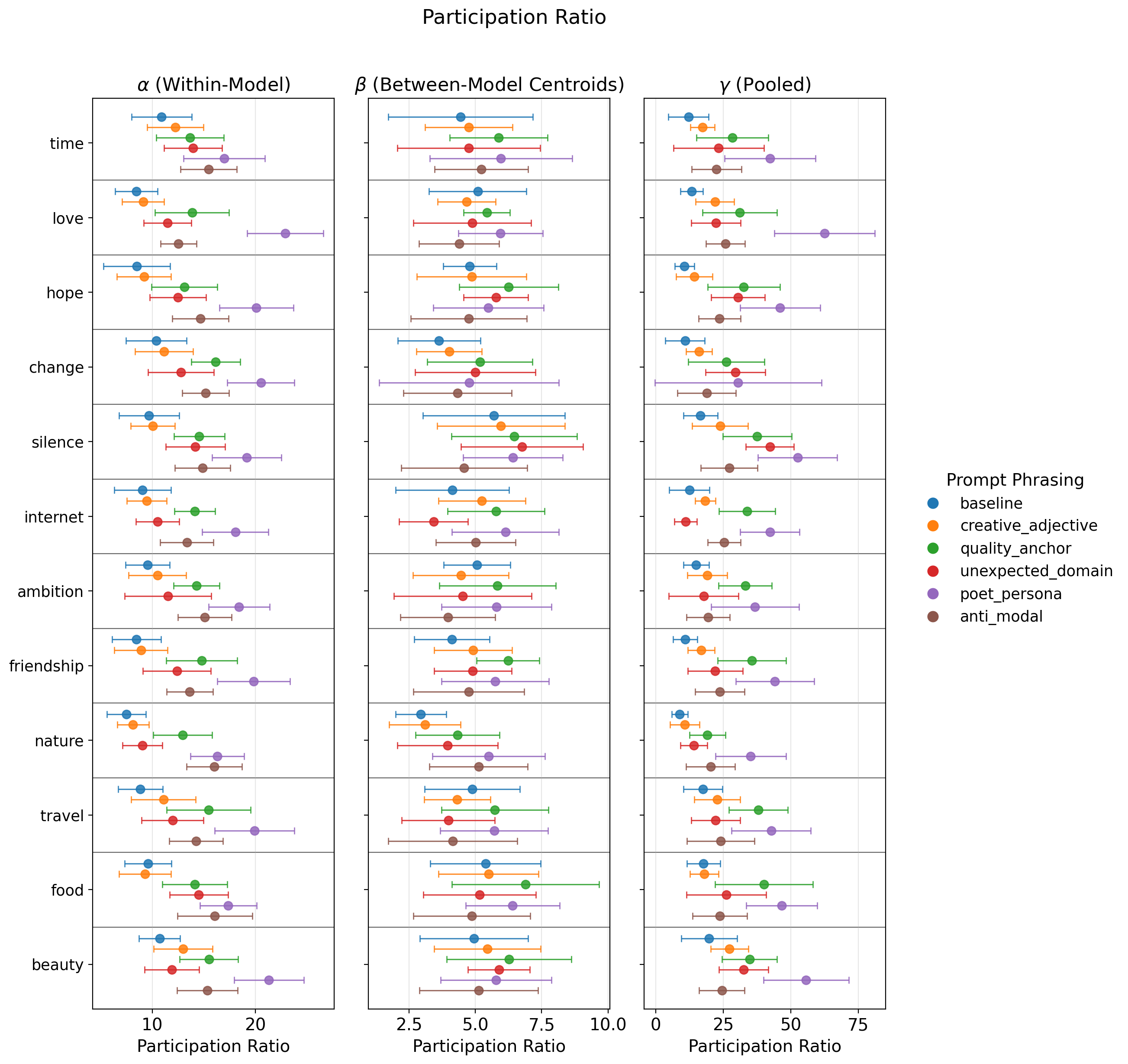}
\caption{\textbf{Participation Ratio per Topic Across the Six Prompt Phrasings.} Participation ratio per topic (12 rows) at the $\alpha$, $\beta$, and $\gamma$ scales for the six phrasings. Error bars are 95\% leave-one-model-out jackknife intervals over the fifteen models. Fig.~\ref{fig:prompt-phrasing-pr} shows the mean over topics.}
\label{fig:prompt-phrasing-pr-per-topic}
\end{figure}

Fig.~\ref{fig:prompt-phrasing-er} in the main text reports effective rank across the six phrasings, averaged over the twelve topics. Figs.~\ref{fig:prompt-phrasing-rao} and \ref{fig:prompt-phrasing-pr} report the same means under Rao's quadratic entropy and participation ratio. Figs.~\ref{fig:prompt-phrasing-er-per-topic}, \ref{fig:prompt-phrasing-rao-per-topic}, and \ref{fig:prompt-phrasing-pr-per-topic} report the per-topic values.

\subsection{Why Rao's Q Is Flat at the \texorpdfstring{$\beta$}{beta} Scale}\label{app:beta-rao-flatness}

The three metrics disagree at $\beta$. Rao's Q is essentially flat: only \texttt{unexpected\_domain} exceeds baseline ($+4.0\%$), and the other four non-baseline phrasings fall $3$--$12\%$ below it. Effective rank and participation ratio rise modestly under every non-baseline phrasing ($\beta$ effective rank $+6.9$ to $+25.9\%$; $\beta$ participation ratio $+2.1$ to $+27.5\%$).
At $\beta$ effective rank, \texttt{poet\_persona} leads at $+25.9\%$, with \texttt{quality\_anchor} essentially tied ($+25.7\%$). The disagreement has a geometric cause. On the renormalized model centroids, $\beta$ Rao's Q reduces to $Q = 1 - \|\bar{c}\|^2$ (derivation in Appendix~\ref{app:alpha-beta-gamma-primer}). This scalar measures how concentrated the centroids are around their mean direction. It ignores how the centroids are arranged around that direction. At baseline, $Q_\beta \approx 0.18$ implies the 15 centroids lie within a narrow solid angle, and no phrasing widens it. Within that solid angle, the eigenvalue-based metrics still detect that the centroid cloud becomes higher-dimensional. At $\alpha$ and $\gamma$, all three metrics agree strongly. At $\beta$, the disagreement is a property of Rao's angular summary, not a failure of any metric.

\subsection{The Response-Length Confound}\label{app:length-confound}

The reworded prompts change response length, a potential confound. Median response length ranges from $0.74\times$ baseline under \texttt{poet\_persona} to $3.01\times$ under \texttt{unexpected\_domain}. Length cannot explain the diversity gains. The phrasing with the largest gain, \texttt{poet\_persona}, produces responses about 26\% \emph{shorter} than baseline. Across the 60 (phrasing, topic) cells, the length ratio correlates negatively with the Rao's-Q gain over baseline (Spearman $\rho = -0.43$ at $\alpha$, $-0.34$ at $\gamma$).

\subsection{An Idea-Labeling Probe of Whether the Gains Add Ideas or Only Wording}\label{app:conceptual-vs-stylistic}

Finding~\ref{finding:rewordings}'s gains could reflect only more varied vocabulary and sentence structure, with no new ideas. To test this, we applied Section~\ref{sec:wrong-null}'s LLM idea-labeling protocol: Claude Fable 5 assigns each response to one core idea. We labeled 100 stratified-sampled responses per (topic, phrasing) cell and kept ideas with ${\geq}10$ members. We ran this on \texttt{baseline}, \texttt{poet\_persona}, and \texttt{anti\_modal} across all twelve topics, twice with independent sampling seeds.
The labeler is not one of the fifteen models in the pool (Appendix~\ref{app:concept-label-details}).
Among responses placed in a kept idea, rewording reduces modal-idea concentration. The modal share falls from 0.48 at baseline to 0.34 under \texttt{poet\_persona} and 0.35 under \texttt{anti\_modal} (pooled over both seeds).
The all-response share drops more, partly because the unclassifiable fraction rises. That share is the fraction of responses expressing a cell's modal idea. Pooled over seeds, it falls from 0.39 at baseline to 0.18 under \texttt{poet\_persona}. That drop holds in 23 of 24 topic-seed cells, with one tie. The two-sided sign test gives $p = 0.0005$ and $p = 0.001$ in the two seeds. This probe's tests are two-sided, unlike the one-sided tests earlier in this appendix. Under \texttt{anti\_modal} the share falls to 0.20. The seed-averaged per-topic deltas give $p = 0.039$ (10 of 12 topics); per-seed $p = 0.039$ and $0.006$. The labeler could not place some responses in any kept idea. That fraction rises from 0.20 at baseline to 0.39--0.45 pooled over seeds. Per seed it is 0.38--0.46 (seed 1) and 0.39--0.43 (seed 2). A larger unclassifiable fraction lowers the all-response modal share by construction.
No idea-count measure shifts significantly. The effective number of ideas is 3.34 at baseline versus 3.99 (\texttt{poet\_persona}) and 3.97 (\texttt{anti\_modal}) pooled. The kept-idea count is 3.62 at baseline versus 4.08 under both rewordings. Neither seed shows a significant per-topic shift. On the seed-averaged per-topic deltas, the smallest of the four idea-count sign-test $p$ values is 0.39.
The rise in the unclassifiable fraction also bears on the idea count. The protocol cannot distinguish many small new ideas from unclassifiable one-off variation. This one confound therefore limits both the share drop and the idea count.

\section{Robustness of the Prompt-Phrasing Result to the Embedding Model}\label{app:diversity-robustness}

\begin{figure}[t!]
\centering
\includegraphics[width=\linewidth]{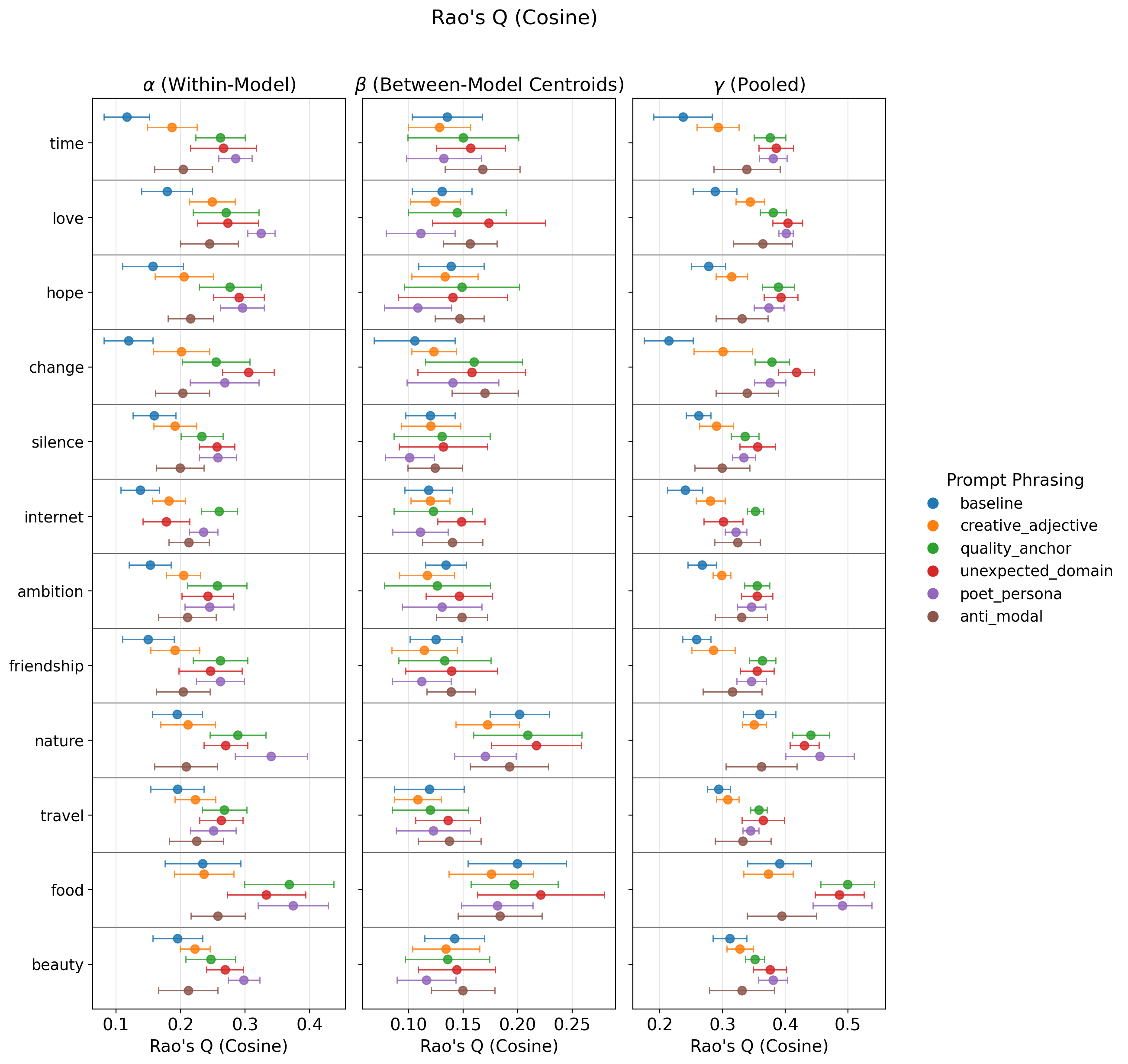}
\caption{\textbf{Rao's Q Across Prompt Phrasings Under EmbeddingGemma-300m; Results Match the Primary OpenAI-Embedding Analysis Qualitatively.} Same layout as Fig.~\ref{fig:prompt-phrasing-rao-per-topic} but with responses embedded by Google DeepMind's \texttt{embeddinggemma-300m} (768 dimensions) instead of OpenAI \texttt{text-embedding-3-small}. Error bars are 95\% leave-one-model-out jackknife confidence intervals over the fifteen models.}
\label{fig:prompt-phrasing-rao-gemma}
\end{figure}

\begin{figure}[t!]
\centering
\includegraphics[width=\linewidth]{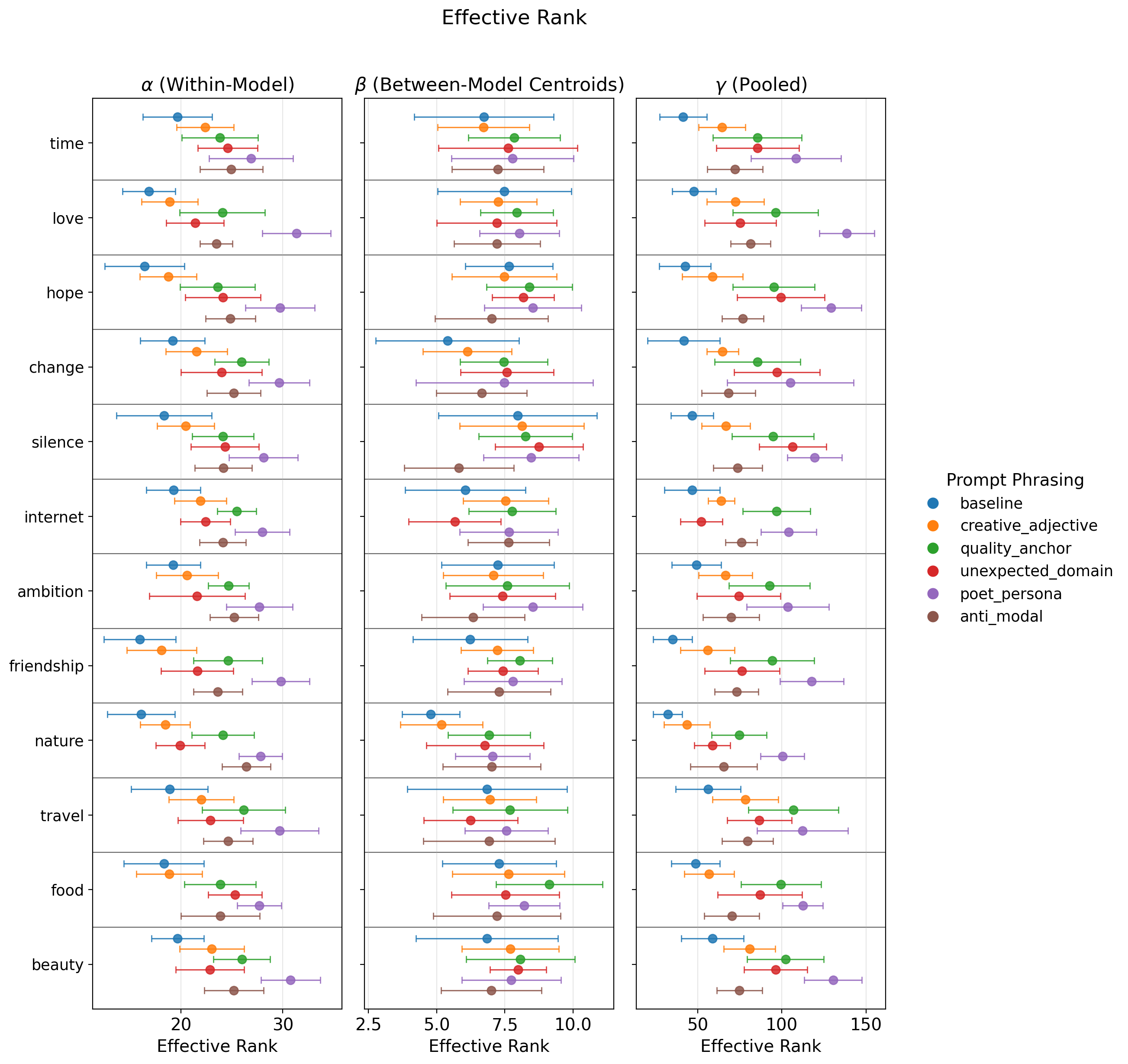}
\caption{\textbf{Effective Rank Across Prompt Phrasings Under EmbeddingGemma-300m; Results Match the Primary OpenAI-Embedding Analysis Qualitatively.} Same layout as Fig.~\ref{fig:prompt-phrasing-er-per-topic} but with responses embedded by \texttt{embeddinggemma-300m}. Error bars are 95\% leave-one-model-out jackknife confidence intervals over the fifteen models.}
\label{fig:prompt-phrasing-er-gemma}
\end{figure}

\begin{figure}[t!]
\centering
\includegraphics[width=\linewidth]{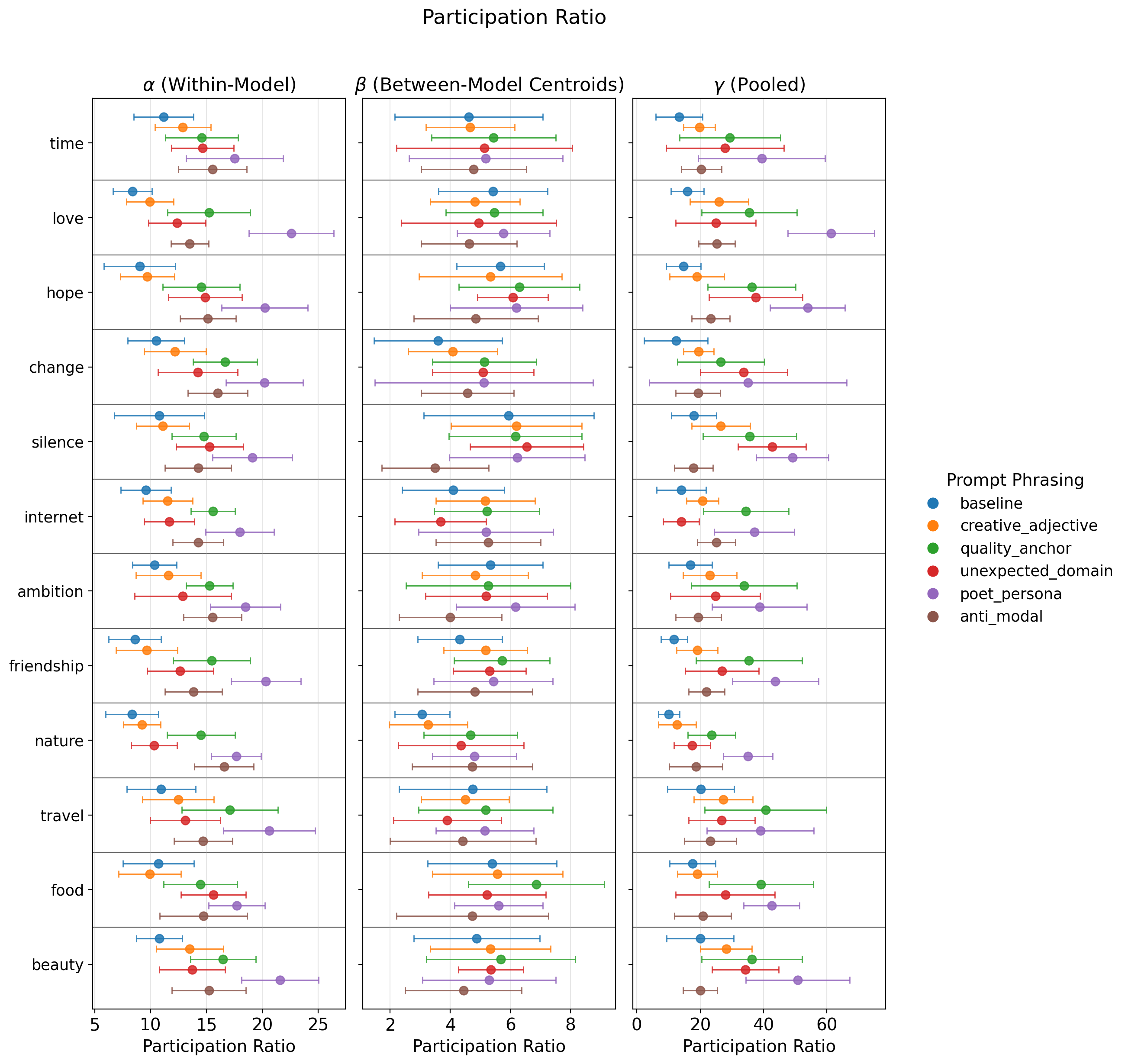}
\caption{\textbf{Participation Ratio Across Prompt Phrasings Under EmbeddingGemma-300m; Results Match the Primary OpenAI-Embedding Analysis Qualitatively.} Same layout as Fig.~\ref{fig:prompt-phrasing-pr-per-topic} but with responses embedded by \texttt{embeddinggemma-300m}. Error bars are 95\% leave-one-model-out jackknife confidence intervals over the fifteen models.}
\label{fig:prompt-phrasing-pr-gemma}
\end{figure}

Section~\ref{sec:inference-time} reports all three diversity metrics (Rao's quadratic entropy, effective rank, participation ratio) on the OpenAI \texttt{text-embedding-3-small} embeddings. We recomputed every metric at every scale on a second embedding model, Google DeepMind's \texttt{embeddinggemma-300m} (768 dimensions). The phrasing ordering and the qualitative conclusion match the primary analysis (Figs.~\ref{fig:prompt-phrasing-rao-gemma}, \ref{fig:prompt-phrasing-er-gemma}, and \ref{fig:prompt-phrasing-pr-gemma}). Each of the five phrasings outranks baseline in 81.5--96.3\% of the 108 topic-scale-metric cells under this embedding (versus 79.6--91.7\% under the primary embedding). All five non-baseline phrasings raise $\alpha$- and $\gamma$-scale diversity above \texttt{baseline}. \texttt{Poet\_persona} is the strongest phrasing at $\alpha$ and $\gamma$ under all three metrics. Rao's Q at $\beta$ is again flatter than the eigenvalue-based metrics, for the reason given in Appendix~\ref{app:beta-rao-flatness}.

\end{document}